\documentclass{article}

\usepackage[%
journal=MSL,    
lang=british,   
]{ems-journal}

\usepackage[utf8]{inputenc}
\RequirePackage[OT1]{fontenc}
	\RequirePackage[numbers]{natbib}
\RequirePackage[shortlabels]{enumitem}

\newcommand{\pa}[1]{\left(#1\right)}
\newcommand{\cro}[1]{\left[#1\right]}
\newcommand{\ac}[1]{\left\{#1\right\}}

\def\cB{\mathcal{B}}
\def\cE{\mathcal{E}}
\def\cF{\mathcal{F}}
\newcommand{\E}{\operatorname{\mathbb{E}}}
\renewcommand{\P}{\operatorname{\mathbb{P}}}
\newcommand{\R}{\mathbb{R}}
\newcommand{\Nbb}{\mathbb{N}}

\newcommand{\Pbb}{\P}

\newcommand{\Acal}{\mathcal{A}}
\newcommand{\Ecal}{\mathcal{E}}
\newcommand{\Fcal}{\mathcal{F}}
\newcommand{\Ncal}{\mathcal{N}}
\newcommand{\Ocal}{\mathcal{O}}

\newcommand{\Vcal}{\mathcal{V}}
\newcommand{\Zcal}{\mathcal{Z}}

\newcommand{\misclas}{\operatorname{misclas}}
\newcommand{\regret}{\operatorname{bad}}

\newcommand{\one}{\mathbf{1}}

\newcommand{\thresh}{{\text{thresh}}}

\newcommand{\within}{{\text{within}}}
\newcommand{\out}{{\text{out}}}

\newcommand{\seuil}{\frac{kI}{16}}
\newcommand{\cfail}{{C_\text{fail}}}
\newcommand{\cmisc}{c_{\text{misclas}}}
\newcommand{\cmiscpost}{c_{\text{misclas}}^\text{after}}

\numberwithin{equation}{section}
\newtheorem{thm}{Theorem}

\newtheorem{lem}[thm]{Lemma}

\newtheorem{conj}{Conjecture}

\begin{document}

\title{Pair-Matching: Link Prediction with Adaptive Queries}



\emsauthor{1}{
	\givenname{Christophe}
	\surname{Giraud}
	\mrid{}
	\orcid{0009-0004-1836-5742}}{C.~Giraud}
\emsauthor{2}{
	\givenname{Yann}
	\surname{Issartel}
	\mrid{}
	\orcid{}}{Y.~Issartel}

\emsauthor{3}{
	\givenname{Luc}
	\surname{Lehericy}
	\mrid{}
	\orcid{}}{L.~Lehericy}

\emsauthor{4}{
	\givenname{Matthieu}
	\surname{Lerasle}
	\mrid{}
	\orcid{}}{M.~Lerasle}

\Emsaffil{1}{
	\department{Laboratoire de Math\'ematiques d'Orsay}
	\organisation{Universit\'e Paris-Saclay, CNRS}
	\rorid{}
	\address{Bâtiment 307, rue Michel Magat}
	\zip{F-91405}
	\city{Orsay}
	\country{France}
	\affemail{christophe.giraud@universite-paris-saclay.fr}}

%
\Emsaffil{2}{
        \department{1}{Laboratoire de Math\'ematiques d'Orsay}
	\organisation{1}{Universit\'e Paris-Saclay, CNRS}
	\rorid{}
	\address{1}{Bâtiment 307, rue Michel Magat}
	\zip{1}{F-91405}
	\city{1}{Orsay}
	\country{1}{France}
	\department{2}{Laboratoire Traitement et Communication de l'Information (LTCI)} 
	\organisation{2}{Institut Polytechnique de Paris, Télécom Paris}%
	\rorid{2}{}
	\address{2}{19 Pl. Marguerite Perey}%
	\zip{2}{91120}
	\city{2}{Palaiseau}
	\country{2}{France} 
	\affemail{2}{yann.issartel@telecom-paris.fr}}

 \Emsaffil{3}{
	\department{1}{Laboratoire de Math\'ematiques d'Orsay}
	\organisation{1}{Universit\'e Paris-Saclay, CNRS}
	\rorid{}
	\address{1}{Bâtiment 307, rue Michel Magat}
	\zip{1}{F-91405}
	\city{1}{Orsay}
	\country{1}{France}
	\department{2}{Laboratoire J.A. Dieudonn\'e} 
	\organisation{2}{Universit\'e C\^ote d’Azur}%
	\rorid{2}{}
	\address{2}{Parc Valrose}%
	\zip{2}{06108}
	\city{2}{Nice}
	\country{2}{France} 
	\affemail{2}{luc.lehericy@univ-cotedazur.fr}}
	
\Emsaffil{4}{
	\department{1}{Centre de Recherche en Économie et de Statistiques (CREST)}
	\organisation{1}{CNRS, École polytechnique, GENES, ENSAE Paris, Institut Polytechnique de Paris}
	\rorid{1}{}
	\address{1}{5 avenue Henri Le Chatelier}
	\zip{1}{91120}
	\city{1}{Palaiseau}
	\country{1}{France}
	\affemail{matthieu.lerasle@ensae.fr}}


\classification[68T05]{62H30}


\begin{abstract}%
The pair-matching problem appears in many applications where one wants to discover matches between pairs of entities or individuals.
Formally, the set of individuals is represented by the nodes of a graph where the edges, unobserved at first, represent the matches.
The algorithm queries pairs of nodes and observes the presence/absence of edges.
Its goal is to discover as many edges as possible with a fixed budget of queries. 
Pair-matching is a particular instance of multi-armed bandit problem in which the arms are pairs of individuals and the rewards are edges linking these pairs. 
This bandit problem is non-standard though, as each arm can only be played once. 

Given this last constraint, sublinear regret can be expected only if the graph presents some underlying structure. 
This paper shows that sublinear regret is achievable in the case where the graph is generated according to a Stochastic Block Model (SBM) with two communities. 
Optimal regret bounds are computed for this pair-matching problem. They exhibit a phase transition related to the Kesten-Stigum threshold for community detection in SBM. 
The pair-matching problem is considered in the case where each node is constrained to be sampled less than a given amount of times, for example for ensuring individual fairness.
We show how optimal regret rates depend on this constraint.
The paper is concluded by a conjecture regarding the optimal regret when the number of communities is larger than 2.
Contrary to the two communities case, we argue that a statistical-computational gap would appear in this problem.
\end{abstract}

\maketitle

\section{Introduction}

\subsection{Motivation}
Many real world data can be represented as a graph of pairwise relationships. Examples include social networks connections, metabolic networks, protein-protein interaction networks, citations network, recommendations and so on. 
Matchmaking algorithms and link prediction algorithms are routinely used in many practical situations  to discover biochemical interactions, new contacts, hidden connections between criminals, or to match players in online multiplayers video games and sport tournaments. 
As testing a link in biological networks, or discovering connections between criminals can be expansive, link prediction algorithms are useful to focus on the  most relevant links. 
In social networks or online video games, they can help in finding relevant partners.

\subsection{Problem}
These applications raise the following mathematical problem that this paper intends to study.
Suppose that there exists a graph whose nodes represent a set of entities or individuals and whose edges represent matches between entities or individuals.
The nodes are known to the statistician while the edges are typically hidden at first.
Matchmaking algorithms make queries on pairs of individuals, trying to discover as many edges as possible.
For biological networks like protein-protein interaction networks, the individuals are proteins, an edge is an interaction between the two proteins and a query is an experiment to test whether the interaction exists.
The goal of matchmaking algorithms is to discover as many edges of the graph as possible while minimizing the number of mismatches. 
To stress that the focus lies on discovering graph structures, the problem at hand is called hereafter pair-matching rather than matchmaking.

In this paper, pair-matching algorithms are constrained to explore the graph as they cannot make queries on edges that have already been observed. 
To learn interesting features on unobserved edges from previous observations, it is necessary to make assumptions on the structure of the hidden graph.
This paper considers the arguably simplest situation where the graph has been generated according to an assortative conditional stochastic block model (SBM) \citep{Holland83} with two balanced communities, see 
Section~\ref{Sec:Setting} for a formal presentation.
In this model, individuals are grouped into two (unobserved) communities and the probability of a match (edge) between two individuals is larger if they belong to the same community than to different ones.
In this context, the set of pairs is partitioned into good and bad ones, good pairs contain two individuals from the same community and bad pairs two individuals from different communities.
A pair-matching algorithm samples pairs and should sample as many good pairs as possible.
Of course, the partition into good and bad pairs is unknown.

When the graph is fully observed, communities are recovered using clustering algorithms, which have been extensively studied over the past few years, see for example the recent overviews in \citep{AbbeReview2017,MooreReview2017,CanLevinaVershyninICM2018}.
A key parameter in the analysis of clustering algorithms, called here \emph{scaling parameter} $s$, is the ratio
\[
s=\frac{(p-q)^2}{p+q},
\]
where $p$ is the probability of connection within a community and $q$ the probability of connection between communities.
This parameter measures the difficulty of clustering, see Section~\ref{Sec:Setting} for details.
The quality of a pair-matching algorithm is evaluated by the expected number of discovered edges after $T$ queries.
Equivalently, the performance can be measured by the expected number of pairs sampled that do not contain edges, which should be as small as possible, see Section~\ref{Sec:Obj} for details.
This last quantity is proportional to the  expected number of bad pairs sampled, which is called \emph{sampling regret} in this paper.
Besides constraining the algorithms to sample each pair only once, we also force algorithms to sample each individual less than a certain amount of times $B_T$ before $T$ queries have been made. 
This constraint corresponds to practical situations where each individual may not be solicited too many times.  
For example, for fairness policy, the algorithm may be required to sample a similar amount of time a large fraction of the $n$ individuals. 
Such fairness constraint is then implemented by forcing to have at most $B_{T}=cT/n$ queries per individual, where $c\geq 1$.

\subsection{Contribution}\label{section:contribution}
Our objective is to understand precisely the order of magnitude of the optimal regret in this online learning problem.
This task is not trivial due to the originality of our setting, where pairs and nodes cannot be sampled as much as desired, but where the hidden structure of the graph should help to learn useful information.
Our  main contribution is that the sampling regret of any strategy that cannot sample pairs more than once, that is invariant to nodes labelling and which satisfies the above constraint (see Assumptions {\bf (NR)}, {\bf (IL)} and {\bf (SpS)} in Section \ref{Sec:SeqMatch} for details) is larger than
\begin{equation}\label{intro:optimal-regret}
T \wedge \frac{\sqrt{T}\vee (T/B_{T})}{s},
\end{equation}
up to multiplicative constants.
Moreover, a polynomial-time algorithm with sampling regret bounded from above by a constant times $T \wedge \frac{\sqrt{T}\vee (T/B_{T})}{s}$ is described and analyzed, see Theorem~\ref{thm:contraint}.
It follows that $T \wedge r(T,s)$ , where  $r(T,s)= \frac{\sqrt{T}\vee (T/B_{T})}{s}$ is the order of magnitude of the optimal regret we were looking for and that this rate can be achieved in polynomial time. 
As a consequence, when $T=O(1/s^2)$  pairs have been sampled,  the linear sampling regret is unavoidable since  $r(T,s)\geqslant \sqrt{T}/s\geqslant T$. 
Likewise, $r(T,s)\geqslant T/(sB_T)\gtrsim T$ when $B_T=O(1/s)$ and linear sampling regret is also unavoidable when the constraint is too strong.
On the other hand, when $B_T\gg 1/s$, and $T\gg 1/s^2$, our algorithm reaches the optimal sub-linear sampling regret $r(T,s)$.
A particular interesting arrises when $T=n^{\alpha}$ for some $\alpha\in(1,2)$ and when, for fairness requirement, the algorithm is required to sample at most $cT/n$ time each individual.
In this situation actually, the optimal sampling regret becomes of order
\begin{equation*}
T \wedge \frac{T^{1/\alpha}}{s}\;.
\end{equation*}
This result illustrates the price to pay for the fairness constraint: The optimal unconstrained rate $\sqrt{T}/s$ is replaced by the larger rate $T^{1/\alpha}/s$.

\subsection{Related literature}
The following problem, related to matchmaking, has recently attracted attention, in particular in Bradley-Terry models, see \citep{bradley:terry:1952,zemerlo:1929}. 
The task is to infer, from the observation of pairs, a vector of parameters characterizing the strength of players.
Most results considered the case where all the graph is observed, see \citep{hunter:2004,caron:doucet:2012}.
Recent contributions dealing in particular with ranking issues also consider the case of partially observed graphs, see \citep{MR3827087,MR3504618,Jang2016TopKRF} and the references therein.
In all cases, the list of observed pairs is given as input to the algorithm evaluating the strength of all players. 
The choice of a relevant list of successive observed pairs, independent of the observation of the edges is sometimes called a scheduling problem, see \citep{lecorff:lerasle:vernet:2018}.
Scheduling problems are different  from matchmaking problems considered here where the algorithm should choose the observed pairs and can use preliminary observations to make its choice.
For online video games, classical algorithms used to evaluate strength of players are ELO or TRUESKILLS, see \citep{Trueskills2007,Trueskills2018}.
Matchmaking algorithms such as EOMM, see \citep{Chen:2017:EEO:3038912.3052559}, which is used with TRUESKILLS, see \citep{Trueskills2018}, are then used to pair players, taking as inputs these estimated strengths.
In this approach, the number of mismatches during the learning phase is not controlled. 
It is an important conceptual difference with this paper where the matchmaking problem is considered together with the problem of discovering the strength (communities here). 
Here, pair-matching algorithms have to simultaneously explore the graph to evaluate the  strength and sample as many good pairs as possible to optimize the number of matches. 
Closer to our setting is the active ranking literature, where the goal is to discover adaptively the rank or strength of players with a minimal amount of queries, see \citep{NIPS2011_4427, NIPS2015_5903, ActiveRanking2016}. Contrary to our problem, only the exploration matters in adaptive ranking and no notion of regret is investigated.

Pair-matching algorithms take sequential decisions to explore new pairs exploiting previous observations.
This kind of exploration and exploitation dilemma is typical in multi-armed bandit problems, see \citep{Thompson33,robbins1952,LaiRobbins85,BURNETAS96}. 
In stochastic multi-armed bandit problems, a set of actions, called \emph{arms} is proposed to a player who chooses one of these actions at each time step and receives a payoff. 
The payoffs are independent random variables with unknown distribution.
For any arm, payoffs are identically distributed.
The player wants to maximize its total payoff after $T$ queries.
The pair-matching problem introduced above can be seen as a non-standard instance of stochastic multi-armed bandit problems. 
In this interpretation, each pair of nodes is an arm and the associated payoff is $1$ if an edge links these nodes and $0$ otherwise.
The payoffs hence follow a Bernoulli distribution with parameter $p$ for good pairs and parameter $q$ for bad pairs. 
The unusual feature is that each arm can only be played once, so the pair-matcher must choose a new arm at each time step. 
For this reason, optimal strategies differ in spirit from classical strategies in bandit problems, see Section~\ref{sec:bandits} for more details. 
While the proofs of lower bounds involve useful inequalities borrowed from the classical bandit literature, in particular the data processing inequality from \citep{KCG16,GMS18}%
, they require a non-trivial adaptation to work in our setting. The lemmas involved in this adaptation may be of use in a wider class of problems with pair-wise observations.

Forgetting the constraint that a node cannot be sampled more than $B_{T}$ times, the pair-matching bandit problem could be seen as an extreme version of mortal or rotting bandit problems see \citep{MortalNIPS2008, RottingNIPS2017,RottingPMLR2019}, where every arm would systematically die or have zero pay-off after the first sampling. 
Without additional assumptions, the regret would be inexorably linear in the querying budget $T$. 
Here, an important difference with classical mortal or rotting multi-armed bandits is that payoffs are structured by the underlying stochastic block model (SBM). 
While pair-matching can easily be formalized as a bandit problem, and while we explore the parallel with some specific bandit problems like $k$ out of $m$ bandit, it turns out that neither the algorithms nor the results or their proof can be used in our framework.
In particular, our strategy mixes clustering, iterative rounds of screening and exploitation steps in order to identify a sufficiently large set of nodes in a single community, thus focusing on eliminating sub--optimal arms (at the price of also eliminating optimal ones) rather than identifying good ones as bandit algorithms usually do.
Notice though that the paper \citep{de2021bandits}, which is posterior to our work, considered related questions in a different setting.

Stochastic block models  have attracted a lot of attention in the recent years, with a focus on the determination of optimal strategies for clustering and for parameter estimation,  see \citep{AbbeReview2017,MooreReview2017}. 
In this prolific literature, the graph is fully observed and the question is to identify precisely the weakest separation between the probabilities of connection necessary to perfectly or partially recover the communities, or to estimate the parameters of the SBM. 
Closer to our setting, the paper \citep{YP14a} investigates the question of recovering communities from a minimal number of observed pairs, sampled sequentially. 
In this problem, the question is to assign a community to all nodes after a minimal number $T$ of time steps and try to minimize the number of misclassified nodes.
This is quite different from the minimization of the sampling regret considered here, where we seek to find on a budget as many good pairs as possible and not to classify all nodes. 
As discussed in Section~\ref{sec:discuss:step2}, applying the algorithm of \citep{YP14a} would lead to a suboptimal regret in our problem.

The formalization of the pair-matching problem considered in this paper may be restrictive in some applications.
Section~\ref{Sec:KclassesSBM} presents some conjectures that seem reasonable for $K$ classes SBMs.
Other graph structures would also be interesting, such as Bradley-Terry models of \citep{bradley:terry:1952,zemerlo:1929}, which have been used for sport tournaments in \citep{Sir_Red:2009}, chess ranking in \citep{joe:1990} and predictions of animal behaviors in \citep{whiting:stuartfox:oconnor:firth:bennett:bloomberg:2006}. 
Various constraints dealing with first discoveries for example may be interesting depending on the applications: the first match of a node is the most important in some situations,
and, for the search of a life partner, discovering a match with a node already connected in the observed graph is (for most nodes at least) less interesting than a match with an isolated node.
These constraints naturally induce different versions of the pair-matching problem and raise mathematical questions of interest.
Multiplayer video games suggest the extension to hypergraphs of the pair-matching problem. 
Indeed, the value of a player could be evaluated as part of a team and with respect to a possible team of opponents rather than simply as part of a pair.
Finally, in many practical situations, additional information on individuals is available and could be used to improve pair-matching algorithms.
It is clear from our first results that this information is necessary to avoid linear regret in applications such as life partner research. 
These extensions are postponed to follow-up works.
This paper should be seen as a first step to formalize and study the important sequential pair-matching problem.
It focuses on a toy example but opens several interesting questions that arise  when dealing with natural constraints in practical applications of interest.

\subsection{Organization and notation}

The remainder of the paper is decomposed as follows.
Section~\ref{Sec:Set} introduces the formal setting and objectives.
As a warm-up, Section~\ref{sec:Unconstrained} focuses on the case where the algorithms are not constrained to sample nodes more than a certain amount of times.
Section~\ref{sec:contraint} presents the main results where the algorithm are constrained.
Section~\ref{Sec:KclassesSBM} gives conjectures for $K$-classes SBMs.
{Finally, in Section~\ref{sec_simulations}, we assess the behaviour of the algorithms on synthetic data, and we explore the estimation of the scaling parameter as well as the robustness of our result to slight model misspecification.}
The proofs of the main results are postponed to the appendix. 

%

Notation:
we write $x_n\lesssim y_n$ and $x_n=O(y_n)$, if there exist numerical constants such that $x_n\leqslant Cy_n$ for all $n\geq n_0$; and we write
$x_{n} \asymp y_{n}$ and $x_n=\Theta(y_n)$, if $x_n=O(y_n)$ and $y_n=O(x_n)$ that is, if there exist numerical constants $c,c'>0$ and $n_0$ such that $c x_{n}\leq y_{n} \leq c' x_{n}$ for all $n\geq n_0$. We denote by $\lceil x\rceil$ (respectively $\lfloor x \rfloor$)  the upper (resp. lower) integer part of $x$; by $|A|$ the cardinal of a set $A$; and by $A\Delta B$ the symmetric difference between two sets $A$ and $B$.

\section{Setting and Problem Formalization}\label{Sec:Set}

\subsection{A Special Bandit Problem}
\label{sec:bandits}

In the pair-matching problem described above, the data-scientist uncovers sequentially a random graph, whose nodes are clustered into two communities. The probability to have an edge between two nodes within a community is $p$, while this probability is $q$ for two nodes belonging to different communities.   The problem to discover sequentially as many edges as possible in $T$ steps, without querying more than $B_{T}$ times a node, can be interpreted as a non-standard multi-armed bandit problem.
Actually, each pair $\{a,b\}$ of nodes can be seen as an arm and the discovery of match as a payoff.
The payoff of the arm $\{a,b\}$ follows a Bernoulli distribution with parameter $p$ if $a$ and $b$ belongs to the same communities, and with parameter $q$ if they are in different communities.
This bandit problem is non-standard, as 
\begin{enumerate}
\item the arms are sampled at most once,
\item at most $B_T$ arms involving a given node can be sampled up to time $T$,
\item the distribution of the payoffs have a hidden structure inherited from the SBM setup.
\end{enumerate}
Compared with the standard multi-armed bandit problem, points $1$ and $2$ make this problem harder, while point $3$ is a strong structural property that gives hope to find regimes with sub-linear regret.

These special features make this problem quite different from classical bandit problems. 
In classical bandit problems, optimal strategies have to identify the best arm (or some of the best arms) and each arm is played many times to reach this goal. 
Here, half the arms are ``optimal" but one cannot play an arm more than once.
Therefore, instead of identifying one of these, optimal strategies should avoid bad arms,
possibly disregarding a non-negligible proportion of good arms in the process. 

The constraint 2.\ also induces a specific exploration / exploitation trade-off. 
When the community of a node is identified, we wish to query it with a maximum of nodes of the same community in order to maximise the rewards (exploitation).
Yet, we also need to pair this node to some new nodes in order identify the community of new nodes (exploration). 
Since a node can be queried at most $B_{T}$ times, we need  to trade-off between these two strategies.
 
Due to these unusual features of the problem, the classical bandit literature is of little help in order to design some optimal sampling algorithm. It is yet useful to establish our lower bounds, which involve inequalities from \citep{GMS18,KCG16}. 

In the remaining of this section, the problem, the assumptions and the objectives are described more formally.

\subsection{Two-Classes SBM}\label{Sec:Setting}
The $n$ individuals are represented by the set $V=\ac{1,\ldots,n}$. Matches  are represented by a set of edges $E$ between nodes in $V$: there is a match between $a$ and $b$ in $V$ if and only if the pair $\ac{a,b}$ belongs to $E$. 
Hereafter, a set of two distinct elements in $V$ is called a \emph{pair} and an element of $E$ is called an \emph{edge}. 
The graph $(V,E)$ is conveniently represented by its adjacency matrix $A\in\R^{n\times n}$, with entries $A_{ab}=1$ if $\ac{a,b}\in E$ and $A_{ab}=0$ otherwise. 
In the following, any graph $(V,E)$ is identified with its adjacency matrix $(A_{ab})_{a,b \in V}$.
For any pair $e = \{a,b\}$, the notations $A_e$ and $A_{ab}$ are used indifferently. 
Since the graph is undirected, the adjacency matrix $A$ is symmetric, and since there is no self-matching (no self-loop in the graph), the diagonal of $A$ is equal to zero.

Individuals are grouped into two (unknown) communities according to their affinity.
To model this situation, the graph $(V,E)$ is random and distributed as a two-classes conditional stochastic block model. 
Let $0<q,p<1$, and let $n_1$ denote an integer $n_{1}\geq n-n_{1}\geq 1$.
The collection cSBM$(n_{1},n-n_{1},p,q)$ of two-classes conditional stochastic block model distributions on graphs is defined as follows. 
Let $G=\ac{G_{1},G_{2}}$ be a partition of $\ac{1,\ldots,n}$ into two groups, with $|G_{1}|=n_{1}$ and $|G_{2}|=n-n_{1}$. The partition $G$ represents the communities of individuals. Let $\mu_{G}$ denotes the distribution  on graphs with nodes $\ac{1,\ldots,n}$, such that the adjacency matrix is symmetric, null on the diagonal and with lower diagonal entries $(A_{ab})_{a < b}$ sampled as independent Bernoulli random variables with $\mu_{G}(A_{ab}=1)=p$ when $a$ and $b$ belong to the same group $G_{i}$, and $\mu_{G}(A_{ab}=1)=q$ when $a$ and $b$ belong to different groups. In other words, two individuals are matched with probability $p$ if they belong to the same community, and with probability $q$ otherwise. 
The class cSBM$(n_{1},n-n_{1},p,q)$ is defined as the set of all distributions $\mu_G$ defined above, where $G=\{G_1,G_2\}$ describes the set of partitions of $\{1,\ldots,n\}$ satisfying $|G_1|=n_1$ and $|G_2|=n-n_1$:
\begin{multline*}
 \textrm{cSBM}(n_{1},n-n_{1},p,q)\\
 =\ac{\mu_{G}: G=\ac{G_{1},G_{2}} \ \textrm{partition satisfying}\ |G_{1}|=n_{1},\ |G_{2}|=n-n_{1}}.
\end{multline*}
In the following, the communities are balanced and matches happen with higher probability if individuals belong to the same community. 
Formally, $n$ is even and the graph $(V,E)$ has been generated according to a distribution $\mu$ in cSBM$(n/2,n/2,p,q)$, for some unknown parameters $p$ and $q$ such that $0<q<p\leq 1/2$.
As $q<p$, the distribution of $(V,E)$ is called an \emph{assortative} cSBM$(n/2,n/2,p,q)$.
All along the paper, the ratio $p/q$ is also assumed bounded from above. 
To sum up, $p$ and $q$ are smaller than $1/2$ and satisfy 
\begin{equation}\label{eq:pq}
1<p/q \leq \rho^*.
\end{equation}
{The rationale for the upper bound in ~\eqref{eq:pq} is provided in the Remark below Theorem~\ref{thm:non-contraint}.} 
Given $p$ and $q$, the following scaling parameter plays a central role
\begin{equation}\label{eq:def:ScalingParam}
 s= \frac{(p-q)^2}{p+q}.
\end{equation}
This parameter appears in various results in the literature on SBM. 
The following property, proved for example in \citep{YP14b,CRV15, AS15,LZ16,GMZZ17,FC17,GV18}, will be used repeatedly in the paper.
When the graph $(V,E)\sim\text{cSBM}(n_{1},n-n_{1},p,q)$, there exist polynomial-time clustering algorithms that return a partition of $\{1,\ldots,n\}$ such that, with large probability, the proportion of misclassified nodes decreases exponentially:
\begin{equation*}
\textrm{Proportion of misclassified nodes} \leq \exp(-c n s), \quad \textrm{when}\ ns\geq c',
\end{equation*}
where $c,c'>0$ are numerical constants.
The  rate $ns$ of exponential decay in this result is optimal (up to a constant) when (\ref{eq:pq}) is met.
Hence, the scaling parameter $s$ drives the difficulty of clustering. 
To stress the importance of $s$, the following parametrization will be used henceforth
\begin{equation*}
p= s  \pa{\alpha+\sqrt{\alpha}}/2,\quad q=s \pa{\alpha-\sqrt{\alpha}}/2,
\end{equation*}
with $\alpha=(p+q)^2/(p-q)^2$. 
In this parametrization, Assumption (\ref{eq:pq}) is met if and only if $\alpha$ is bounded from below by $(\rho^*+1)^2/ (\rho^*-1)^2$. 
Another useful property is that there exist numerical constants $c_{1},c_{2}>0$ such that non-trivial community recovery is possible as soon as $s\geq c_{1}/n$, see \citep{Decelle2011,Mas14,mossel2018proof,CRV15,AS15,BLM18,FC17,GV18}  and perfect community recovery is possible as soon as $s\geq c_{2}\log(n)/n$, see \citep{AS15, CX16, MNS16}.

The reader familiar with SBM literature may be more comfortable with the parametrization $p=a_{n}/n$ and $q=b_{n}/n$ for a SBM distribution with two communities. For a comfortable translation of the results, the following relations between $s$, $\alpha$ and $a_{n}$, $b_{n}$ are provided: 
\begin{gather*}
 s=\frac{(a_{n}-b_{n})^2}{n(a_{n}+b_{n})}, \qquad \alpha=\frac{(a_{n}+b_{n})^2}{(a_{n}-b_{n})^2},\\
\frac{a_{n}}{b_{n}}=\frac{\alpha+\sqrt{\alpha}}{\alpha-\sqrt{\alpha}}\quad \text{and}  \quad a_{n}+b_{n}=n\alpha s. 
\end{gather*}
{With these notations, the optimal sampling regret \eqref{intro:optimal-regret}  can be rewritten as
\begin{equation*}
T \wedge \frac{\sqrt{T}\vee (T/B_{T})}{s}=T \wedge \pa{\pa{\sqrt{T}\vee (T/B_{T})} \frac{n(a_{n}+b_{n})}{(a_{n}-b_{n})^2}},
\end{equation*} which is smaller than $T$ as soon as
$$B_{T}\wedge \sqrt{T} \geq \frac{n(a_{n}+b_{n})}{(a_{n}-b_{n})^2}.$$
In particular, in the sparse regime where $a_n$ and $b_n$ are constants,  $\sqrt{T}$  has to be of the order of $n$, while $s$ has to be of the order of  $1/n$. The regret, which is lower bounded by $T \wedge \sqrt{T}/s$, is then at least  of the order of $T \wedge n^2$, and thus linear.
}

\subsection{Sequential Matching strategies}\label{Sec:SeqMatch}
Denote by $\Ecal$ the set of all pairs of nodes, that is the set of all subsets of $V$ containing two distinct elements.
Heuristically, a sequential matching strategy samples at each time $t$ a new pair $\widehat e_{t}\in \Ecal$, using only past observations $(\widehat e_{1},\ldots,\widehat e_{t-1},A_{\widehat e_{1}},\ldots,A_{\widehat e_{t-1}})$ and an internal randomness of the algorithm. 

Formally, let $U_{0}, U_{1},\ldots$ be i.i.d uniform random variables in $[0,1]$, independent of $A$ and representing the sequence of internal randomness for the algorithm. 
A sequential matching strategy $\psi$ on $\Ecal$ (shortened \emph{strategy} in the following) is a sequence $\psi = (\psi_{t})_{0\leq t\leq  \binom{n}{2}-1}$ of measurable functions $\psi_{t}:\cE^{t}\times \ac{0,1}^{t}\times [0,1]^{t+1}\to \cE$.
Any sequential matching strategy $\psi$ defines a matching algorithm as follows.
The first pair is sampled as $\widehat e_{1}=\psi_{0}(U_{0})$. Then, at each time $t \geq 0$, the pair $\widehat e_{t+1}$ is defined by 
\begin{equation*}
\widehat e_{t+1}=\psi_{t}(\widehat\cE_{t}, (A_{e})_{e\in \widehat\cE_{t}},U_{0},\ldots,U_t)\quad \textrm{with}\quad \widehat\cE_{t}=\ac{\widehat e_{1},\ldots,\widehat e_{t}}.
\end{equation*}
The strategy takes as input the observed graph $(A_{e})_{e\in \widehat\cE_{t}}$ and possibly an internal independent randomness $U_t$ to output the new observed pair $\widehat e_{t+1}$.

In the following, strategies are assumed to satisfy the following constraints: a pair can only be sampled once and strategies are invariant to labelling of the nodes. 
These constraints can be formalized as follows.
\medskip

\noindent{\bf Non-redundancy (NR).} {\it The strategy $\psi$ samples any pair at most once, that is, for any $0\leq t\leq  \binom{n}{2}-1$ and $e_{1},\ldots,e_{t} \in \cE$, the map $\psi_{t}$ fulfils $\psi_{t}(\{e_{1},\ldots,e_{t}\},\ldots)\notin \ac{e_{1},\ldots,e_{t}}$. }
\medskip

\noindent
Invariance to labelling requires some notation. 
For any pair $e\in\cE$ and any strategy $\psi$, let 
\begin{equation}\label{def:Ne}
N_{e}(\psi,t):={\bf 1}_{e\in \widehat \cE_{t}}
\end{equation}
indicate if the pair $e$ has been sampled or not before time $t$ by the strategy $\psi$.
For any non-redundant strategy $\psi$ (i.e. satisfying {\bf (NR)}), pairs are sampled at most once and the observation of $\{N_{e}(\psi,t): e\in\cE\}$ is equivalent to that of
$\widehat\cE_{t}$.

Let $\mu$ be a distribution in cSBM$(n/2,n/2,p,q)$ and $\sigma$ be a permutation of $V$.  
For any pair $\ac{a,b}\in \cE$, let $\sigma(\ac{a,b}):=\ac{\sigma(a),\sigma(b)}$. 
Let $\mu^\sigma$ denote the distribution of $(A_{\sigma(e)})_{e\in\cE}$, where  $(A_{e})_{e \in \cE}$ is distributed according to $\mu$. \label{def:mu_sigma}
\medskip

\noindent{\bf Invariance to labelling (IL).} {\it The distribution of the outcomes of the strategy $\psi$ is invariant by permutations of the nodes labels: For any $\mu\in \textrm{cSBM}(n/2,n/2,p,q)$ and any permutation $\sigma$ on $V$,
the distribution of  $\big(N_{e}(\psi,t): e\in\cE, 1\leq t \leq  \binom{n}{2}\big)$ under $\mu^{\sigma}$ is the same as the distribution of  $\big(N_{\sigma(e)}(\psi,t): e\in\cE, 1\leq t \leq  \binom{n}{2}\big)$ under $\mu$.}
\medskip

\noindent{\bf Remark.} {Any algorithm can be made invariant to labeling by simply relabeling  all the vertices  at random before applying the algorithm. }\smallskip 

Besides {\bf (NR)} and {\bf (IL)}, we consider strategies that do not sample a node more than $B$ times before time $T$.
This constraint appears naturally in practical situations. 
For example, if the algorithm matches biological entities or individuals, one may not want to query too many times each individual for logistic or acceptability reasons.
To stress that the constraint $B$ typically grows with the time horizon $T$, it is denoted $B_T$.
Formally, for any $a\in V$, let
\begin{equation}
\label{def:Na}
N_{a}(\psi,t)=\sum_{b\in V: b\neq a} N_{\ac{a,b}}(\psi,t)
\end{equation}
denote the number of times the node $a$ has been sampled in a pair $\ac{a,b}$ after $t$ queries. 
\medskip

\noindent{\bf Sparse sampling (SpS).} {\it Let $T$ and $B_T$ denote two integers. The strategy $\psi$ is called $B_T$-sparse up to time $T$ if it satisfies} 
\begin{equation}
\label{eq:sparse}
\forall a \in V, \quad N_{a}(\psi,T) \leq B_T.
\end{equation}

Since $N_{a}(\psi,T)\leq (n-1)\wedge T$ for all nodes $a$, choosing $B_T\geq (n-1)\wedge T$  corresponds to the unconstrained case.

\subsection{Objectives of the Pair-matcher}\label{Sec:Obj}

Let $\mu \in \text{cSBM}(n/2,n/2,p,q)$ be the distribution of an assortative conditional stochastic block model with associated partition $G=\ac{G_{1},G_{2}}$.
Pairs within a community have a larger probability to lead to a match, than pairs between two different communities. Accordingly, pairs $\{a,b\}$ with $a$ and $b$ from the same community are called \emph{good pairs}, and 
 $\cE^{good}(\mu)$ (or simply $\cE^{good}$) denotes  the set of such pairs. Similarly, $\cE^{bad}(\mu)$ (or simply $\cE^{bad}$) denotes the set of pairs $\{a,b\}$ with $a$ and $b$ from two different communities.

The objective of the pair-matcher is to discover as many edges (i.e. matches between individuals) as possible with $T$ queries, in expectation or with high-probability. For simplicity, we focus henceforth on the maximization of the expected number of edges discovered. The strategy $\psi$ of the pair-matcher
should then maximize the number of discovered edges, in expectation with respect to the randomness of the SBM and the strategy. 
Optimal strategies should therefore sample as many pairs in $\cE^{\text{good}}$ as possible.
Formally, consider a time horizon $T$ smaller than $|\cE^{\text{good}}|=2\binom{n/2}{2}\sim n^2/4$, and denote by $N^{bad}(\psi,T)=\sum_{e\in \cE^{bad}}N_{e}(\psi,T)$ (respectively $N^{good}(\psi,T)=\sum_{e\in \cE^{good}}N_{e}(\psi,T)$)  the number of pairs in $\cE^{bad}$ (resp. $\cE^{good}$) sampled up to time $T$.
Applying Wald lemma at the second line, we can compute the expected number of discoveries for any
 strategy $\psi$ 
\begin{align*}
\E_\mu \cro{\sum_{t=1}^T A_{\widehat e_{t}}} &= \sum_{e\in\cE^{good}(\mu)} \E_{\mu}\cro{{\bf 1}_{e\in\widehat\cE_{t}}A_{e}}+\sum_{e\in\cE^{bad}(\mu)} \E_{\mu}\cro{{\bf 1}_{e\in\widehat\cE_{t}}A_{e}} \\
&= \sum_{e\in\cE^{good}(\mu)} p \E_{\mu}\cro{{\bf 1}_{e\in\widehat\cE_{t}}}+\sum_{e\in\cE^{bad}(\mu)} q \E_{\mu}\cro{{\bf 1}_{e\in\widehat\cE_{t}}} \\
& = p \E_\mu \cro{N^{good}(\psi,T)}+q \E_\mu \cro{N^{bad}(\psi,T)}\\
&= pT -(p-q) \E_\mu \cro{N^{bad}(\psi,T)},
\end{align*}
where the last line follows from $N^{good}(\psi,T)+N^{bad}(\psi,T)=T$.
Since $p>q$, the maximal expected value of discoveries is achieved by any oracle strategy $\psi^*$ sampling only edges in $\cE^{\text{good}}$.
In that case, $N^{bad}(\psi^*,T)=0$ and the maximal expected number of discoveries is equal to $pT$.
The regret of the strategy $\psi$ is defined as the difference between $pT$ and its expected number of discoveries:
\begin{equation*}
R_{T}(\psi) = pT-\E_\mu \cro{\sum_{t=1}^T A_{\widehat e_{t}}} = (p-q) \E_\mu \cro{N^{bad}(\psi,T)}.
\end{equation*}
As long as $T\leqslant |\cE_{\text{good}}|$, the regret is proportional to the expected number of sampled between-group pairs $\E_\mu \cro{N^{bad}(\psi,T)}$.
Therefore, the main results analyse this last quantity rather than the regret. 
The expected number of bad sampled pairs $\E_\mu \cro{N^{bad}(\psi,T)}$ is called hereafter \emph{sampling-regret}.
\medskip

\noindent{\bf Remark.} \label{page:inv-sampling} Without assumption on $\psi$, the distribution of $N^{bad}(\psi,T)$ may depend on the distribution $\mu$ of the cSBM. 
On the other hand, when the strategy $\psi$ fulfils {\bf (IL)}, the distribution of $N^{bad}(\psi,T)$ does not depend on the distribution $\mu$ in cSBM$(n/2,n/2,p,q)$. 
Indeed, let $\mu,\mu'$ be two distributions  in cSBM$(n/2,n/2,p,q)$.  
By definition, there exists a permutation $\sigma$ on $\ac{1,\ldots,n}$ such that $\mu'=\mu^{\sigma}$, where $\mu^{\sigma}$ has been defined page~\pageref{def:mu_sigma}. Since $\cE^{bad}(\mu^{\sigma})=\sigma^{-1}(\cE^{bad}(\mu))$, it follows from {\bf (IL)} that the distribution under $\mu^{\sigma}$ of $\sum_{e \in \cE^{bad}(\mu^{\sigma})} N_{e}(\psi,T)$  is the same as the distribution under $\mu$ of $\sum_{e \in \cE^{bad}(\mu)} N_{e}(\psi,T)$.

\section{Warm-up: Unconstrained Optimal Pair-Matching}\label{sec:Unconstrained}
\subsection{Optimal Rates for Unconstrained Pair-Matching}
As a warm-up, we focus first on the simplest case, where $B_{T}=+\infty$, which amounts to remove the constraint {\bf (SpS)}. 
Let $\Psi_{\infty}$ denote the set of strategies $\psi$ fulfilling {\bf (NR)} and {\bf (IL)}. The first main result describes the best sampling-regret that can be achieved by a strategy in $\Psi_{\infty}$, as a function of $s$ and $T$.

\begin{thm}
\label{thm:non-contraint}
Let $T$ and $n$ be positive integers with $T \leq |\cE^{\text{good}}|=2\binom{n/2}{2}$. Let $p,q \in [0,1/2]$ be two  parameters fulfilling (\ref{eq:pq}) and such that
\begin{equation}\label{eq:lower-condition:s}
s \leq \frac{1}{32(1+\rho^*)},
\end{equation}
where the scaling parameter $s$ is defined in \eqref{eq:def:ScalingParam}. 
Then, for any $\mu \in \text{cSBM}(n/2,n/2,p,q)$, 
\begin{equation}
\label{eq:non-contraint}
\inf_{\psi \in \Psi_{\infty}}
	\E_\mu \cro{N^{bad}(\psi,T)} \geq \frac{1}{32} \cro{\frac{\sqrt{T}}{32 (1+\rho^*) s}\wedge T}.
\end{equation}

Moreover, there exist two numerical constants $c_{1},c_{2}>0$, and a strategy $\psi \in \Psi_\infty$ corresponding to a polynomial-time algorithm described in Section~\ref{sec:algo}, 
taking  $s$ as input, 
such that,  for any $p,q$ satisfying \eqref{eq:pq}, any $\mu \in \text{cSBM}(n/2,n/2,p,q)$ and any time horizon $1\leq T\leq c_{2}n^2$
\begin{equation*}
\E_\mu \cro{N^{bad}(\psi,T)} \leq c_{1}\cro{\frac{\sqrt{T}}{s}\wedge T}.
\end{equation*}
\end{thm}

The proof of Theorem \ref{thm:non-contraint} is provided in the appendix. The lower bound is proved in Section \ref{sec:lower} and the upper bound in Section \ref{sec:upper}. 
The upper bound derives from a stronger result showing that similar bounds hold with high probability, see Theorem \ref{thm:upper-appendix} for a precise statement.
Theorem~\ref{thm:non-contraint} provides only the upper bound in expectation for clarity. \medskip

\noindent{\bf Remark.} {The parameter $\rho^*$ only appears in the lower bound. 
In fact, the SNR showing up in the proof of the lower bound is \, $\tilde s:=kl(p,q)\vee kl(q,p) \leq 1/16$, where \, $kl(p,q)=p\log(p/q)+(1-p)\log((1-p)/(1-q))$ is the Kullback-Leibler divergence between two Bernoulli distributions with parameters $p$ and $q$. Under the condition $p/q \leq \rho^*$, we have $s \leq \tilde s\leq 2(1+\rho^*) s$, so $s$ and $\tilde s$ are equivalent; see Lemma~\ref{lem:kullback}. Thus, the quantity $\rho^*$ appears when writing the condition with $s$ instead of $\tilde s$ as SNR. \\
We have chosen to use $s$ instead of $\tilde s$ for convenience, as it is a classical SNR  in the SBM literature, and it allows  us to use existing results and clustering routines straightforwardly.
We stress that $s$ can strongly differ from $\tilde s$ when $\rho^*$ is large, but this difference is large only in the `easiest' setting where $p$ and $q$ are markedly different.
On the other hand, the regime where $\rho^*$ is small is more challenging. In the upper bound,  the constants $c_{1}$ and $c_{2}$ are numerical constants that do not depend on $\rho^*$. }\medskip

Theorem~\ref{thm:non-contraint} states that, when (\ref{eq:pq}) holds, for any $\mu \in \text{cSBM}(n/2,n/2,p,q)$ and any time horizon $1\leq T \leq c_{2} n^2$, the optimal sampling-regret 
\begin{equation*}
\inf_{\psi \in \Psi_{\infty}} \E_\mu \cro{N^{bad}(\psi,T)} \asymp {\frac{\sqrt{T}}{s}\wedge T\,},
\end{equation*}
grows linearly with $T$ as long as $T \lesssim 1/s^2$ and becomes sub-linear, of order $\sqrt{T}/s$, when $T \gtrsim 1/s^2$.\medskip

\noindent{\bf Remark.} {For the convenience of the reader familiar with the SBM literature,  the conclusion of Theorem~\ref{thm:non-contraint} in terms of the parametrization $p=a_{n}/n$ and $q=b_{n}/n$ (as in Section~\ref{Sec:Setting}) is
\begin{equation*}
\inf_{\psi \in \Psi_{\infty}} \E_\mu \cro{N^{bad}(\psi,T)} \ \, \asymp \ \,  \frac{n(a_{n}+b_{n})}{(a_{n}-b_{n})^2} \sqrt{T} \ \,  \wedge \ \, T,
\end{equation*}
since 
 $s=\frac{(a_{n}-b_{n})^2}{n(a_{n}+b_{n})}$. 
}

This result can be understood intuitively. 
As long as communities cannot be recovered better than random, there is no hope of getting better sampling-regret than with purely random sampling of the pairs. 
In this regime, the sampling-regret grows linearly with $T$. 
To identify when this occurs, consider the situation where pairs are sampled at random among $N$ nodes and $T=\beta N^2/2$ (with $\beta\leqslant 1$).
Then the \emph{observed} edges at time $T$ are approximately distributed as in a SBM with $N$ nodes,
within-group connection probability $p_{\beta}=\beta p$, and between-group connection probability $q_{\beta}=\beta q$. 
It follows from \citep{Decelle2011,Mas14,MNS15,BLM18} that weak recovery of the communities is possible if and only if $N(p_{\beta}-q_{\beta})^2\geq 2 (p_{\beta}+q_{\beta})$, which is equivalent to
$\sqrt{\beta T} s \geq \sqrt{2}$ or $T\geq 2/(\beta s^2)$. 
Since $\beta\leq 1$ by definition, no information about the communities can be recovered when $T\leq 2/s^2$.
Hence, the sampling-regret is expected to grow linearly with $T$ for $T=O(1/s^2)$. This intuition is confirmed by Eq.~\eqref{eq:non-contraint}.

When $T \gg 1/s^2$, the situation is different. 
Classical results, such as those in \citep{YP14b,CRV15,AS15,LZ16,MNS16,FC17,GV18} among others, ensure that the communities of $N$ nodes can be recovered almost perfectly if $N\gg 1/s$ and all edges between these nodes are observed. 
Therefore, when $1/s \ll N =\big(\sqrt{T}/s\big)^{1/2} \ll \sqrt{T}$, one can sample all the edges between $N$ nodes and recover almost perfectly their community with a sampling regret smaller than $N^2=\sqrt{T}/s$.

A recipe in order to get a sublinear regret is the following. If we are able to find a community of  $\Theta(\sqrt{T})$ nodes, then we can spend a budget of $T$ queries without further regret by sampling pairs among these  $\Theta(\sqrt{T})$ nodes. To do so, we need 
 to identify a community of  $\Theta(\sqrt{T})$ nodes from the $N$ clustered nodes, with  a regret smaller than  $\sqrt{T}/s$. 
Given the $N$ clustered nodes, a carefully designed screening strategy (see Step 2 of algorithm Section~\ref{sec:algo}), can identify the community of a new node with a sampling regret of order $O(1/s)$.
Proceeding recursively, $\Theta(\sqrt{T})$ new nodes can be identified with a sampling-regret of order $O(\sqrt{T}/s)$. 
The remaining budget of $T$ queries can then be spent by sampling pairs among these $\Theta(\sqrt{T})$ nodes without further regret if there were no errors in the community assignment. 
This informal reasoning suggests that the optimal sampling-regret grows like $\sqrt{T}/s$ when $T\gg 1/s^2$. 
Again, this intuition is confirmed by Eq.~(\ref{eq:non-contraint}). 
An algorithm achieving the optimal upper bound in Theorem~\ref{thm:non-contraint} and taking as input $s$ and the time horizon $T$ is provided in Section \ref{sec:algo}.
It essentially proceeds as in the informal strategy outlined above, even if some steps have to be refined. In particular, the identification of a community of $\Theta(\sqrt{T})$ nodes has to be conducted with care in order to balance the regret and the community assignment errors.
The dependency of the algorithm of Section \ref{sec:algo} on the time horizon $T$, can be easily dropped out  with a classical doubling trick, see Section \ref{sec:upper} in the appendix.




 To sum up the discussion: in the early stage where $T=O(1/s^2)$, one cannot do better than random guessing, up to multiplicative constant factors. 
 In the second stage where $T\geq 1/s^2$, the rate $\sqrt{T}/s$ can be interpreted as follows. 
A total of $\Theta(\sqrt{T})$  nodes are involved at time $T$ and, for each of them, $\Theta(1/s)$  observations are necessary to obtain an educated guess of their community .

Finally, Theorem~\ref{thm:non-contraint} can be equivalently stated in terms of the regret $R_{T}(\psi)$:
for any time horizon $1\leq T \leq c_{2} n^2$, the minimal regret satisfies
\begin{equation*}
\inf_{\psi \in \Psi_{\infty}} R_{T}(\psi) \asymp \sqrt{\alpha}\pa{\sqrt{T}\wedge (sT)\,},
\end{equation*}
when the assumptions of Theorem~\ref{thm:non-contraint} are met.

\subsection{Algorithm with Specified Horizon \texorpdfstring{$T$}{T}}
\label{sec:algo}

This section presents an algorithm achieving the upper bound in Theorem~\ref{thm:non-contraint}.
This algorithm takes as input the scaling parameter $s$ \emph{and} the time horizon $T$. 
This dependency on the time-horizon can be avoided with the classical doubling trick, see Section \ref{sec:upper} in the appendix.
We discuss in Section \ref{sec:SNR}  a heuristic for the preliminary estimation of $s$ involving less than $O(1/s^2)$ edges.

When the horizon $T$ is $O(1/s^2)$, any strategy achieves a regret of order $O(T)$.
Hence, without loss of generality, it is assumed in the remaining of the section that $T\geq c_{th}/s^2$ for some numerical constant $c_{th}$.
Moreover, as Theorem~\ref{thm:non-contraint} holds for $T\leq c_2n^2$, it is also assumed that this condition is fulfilled 
for a sufficiently small constant $c_2$.

The algorithm proceeds in three steps. 
In the first step, a core-set $\Ncal$ of $|\Ncal|=\Theta(\sqrt{T}/\log(s\sqrt{T}))$ vertices is chosen uniformly at random and 
each pair within this core-set is sampled with probability $\Theta\big( (\log(s\sqrt{T}))^2/(s\sqrt{T})\big)$. 
Hence, an average of $\Theta (\sqrt{T}/s)$ pairs are sampled  within this core-set.
A community recovery algorithm is run on this observed graph that outputs two estimated communities with a fraction of misclassified nodes vanishing as $O(\log(s\sqrt{T})/(s\sqrt{T}))$  with high probability.

The second step identifies with high probability $\Theta(\sqrt{T})$ vertices from the same community, say community 1. 
To do so, it picks uniformly at random a set $\mathcal{A}_{0}$ of $8 \sqrt{2T}$ vertices outside of the core-set $\Ncal$ (this is possible thanks to the condition $T\leq c_2n^2$) and samples pairs between this set and the estimated community 1 of the core-set. 
This set of edges is used to estimate the connectivity between these vertices and community 1.
Vertices with low connectivity, that seem to belong to community 2, are removed online to keep the sampling regret under control. The goal of this screening is 
not to classify perfectly the $8 \sqrt{2T}$ picked vertices, but instead to sift out vertices of community 2 with a low sampling regret. In particular, a price to pay to achieve this goal is  to
possibly remove a non-negligible proportion of vertices of community 1 from the $8 \sqrt{2T}$ picked vertices.
This second step of the algorithm is crucial for getting the optimal regret rate $O(\sqrt{T}/s)$. 
A simplified version of this second step  can be connected to a particular $k$ out of $m$ best arms identification  problem. This connection is discussed in Section \ref{sec:discuss:step2} below.

The third step samples all pairs $\{a,b\}$ such that $a$ and $b$ belong to the $\Theta(\sqrt{T})$ vertices isolated in the second step of the algorithm, until the remaining budget of $T$ queries is expended.

The pair-matching algorithm calls an external clustering algorithm (generically denoted by {\tt GOODCLUST} in the following).
{\tt GOODCLUST} takes as input a graph $(V,E)$ and outputs a partition $\widehat{G}=(\widehat{G}_1,\widehat{G}_2)$.
We require that {\tt GOODCLUST} fulfils the following recovery property:
There exist numerical constants $c^\text{GC}, c^\text{GC}_{1}>0$ such that, for all $N = N_1+N_2$ and all $\tilde{p}, \tilde{q} \in [0,1]$, if $(V,E)\sim \text{cSBM}(N_1,N_2,\tilde p,\tilde q)$, the proportion of misclassified nodes 
\[
\varepsilon_N=\frac{|\widehat{G_1}\Delta G_{1}|+|\widehat{G_2}\Delta G_{2}|}{2N},
\]
with $\Delta$ the symmetric difference, satisfies
\begin{equation}\label{eq:GoodClust}
\varepsilon_N \leq \exp \left( - c^\text{GC}_{1} N  \frac{(\widetilde{p}-\widetilde{q})^2}{\widetilde{p}}\right),
\end{equation}
with probability at least $1 - c^\text{GC}/N^3$. 
Algorithms achieving this proportion of misclassification can be found e.g. in \citep{GV18}, see also \citep{YP14b,CRV15, AS15,LZ16,GMZZ17,FC17} for similar results.

\noindent { \def\arraystretch{1.3}
\begin{tabular}{|l|}
\hline
 \textbf{Unconstrained Algorithm}\label{alg:unconstrained} \\
\hline
\begin{minipage}{0.99\textwidth} \centering
\begin{minipage}{0.97\textwidth}

\medskip
{\bf Inputs:}  $s$ scaling parameter, $T$ time horizon, $V$ {set of} nodes.\medskip

{\bf Internal constants:} $c_{\Ocal_{0}}=2\vee (1/c^\text{GC}_{1})$, $C_{k}=2200$ and $C_{I}=4$. 

\medskip
\noindent \textbf{Step 1: finding communities in a core-set}
\begin{itemize}[topsep=0pt]
\item[1.] Sample uniformly at random a set $\Ncal\subset V$ of $N = \left\lceil \sqrt{T}/\log(s \sqrt{T}) \right\rceil$ nodes.
\item[2.] Sample each pair of $\Ncal$ with probability $\displaystyle c_{\Ocal_0} \frac{\sqrt{T}}{s\binom{N}{2}}$, call $\Ocal_0\subset \cE$ the output.
\item[3.] Estimate global connectivity $\tau=(p+q)/2$ by $\displaystyle \hat{\tau} = \frac{1}{|\Ocal_0|} \sum_{e \in \Ocal_0} A_{e}$. 
 \item[4.] Run {\tt GOODCLUST} on  the graph with nodes set $\Ncal$ and edges present in $\Ocal_0$. Output, for any $x \in \Ncal$, $\widehat Z_{x}$ the estimated community of $x$. Choose the label $\hat Z=1$ for the largest estimated community.
\end{itemize}

\medskip
\noindent \textbf{Step 2: expanding the communities}
\begin{itemize}[topsep=0pt]
\item[5.] Sample uniformly at random a set $\Acal_{0}$ of $|\Acal_{0}|=\left\lceil 8\sqrt{2T} \right\rceil$ nodes in $V\setminus\Ncal$.
\item[6.] Set  $k = \left\lceil C_k/s \right\rceil $ and  $I = \left\lceil C_I \log (s\sqrt{T}) \right\rceil$
\item[7.] {\bf For} $i=1,\ldots,I$, {\bf do}
\begin{itemize}
\item[(a)] {\bf For} $x\in \Acal_{i-1}$, sample $k$ nodes $(y_{k(i-1)+a}^x)_{a=1,\ldots,k}$ uniformly at random in $\Ncal \cap \ac{\hat Z=1} \setminus \{y^x_a\}_{a = 1, \dots, k(i-1)}$. 
\item[(b)] Sample the pairs   $(\{x,y_{k(i-1)+a}^x\})_{a=1,\ldots,k}$  and let $\hat p_{x,i}=\frac{1}{ki} \sum_{a=1}^{ki} A_{xy_{a}^x}$.
\item[(c)] Select $\Acal_{i}=\ac{x\in \Acal_{i-1}: \hat p_{x,i} \geq \hat \tau}$.  
\item[(d)] {\bf In case}\footnote{with high probability, this undesirable case does not happen} where $\Acal_i=\emptyset$, {\bf then} set $\Acal_I=\emptyset$ and {\bf BREAK}.
\end{itemize}
\end{itemize}

\medskip
\noindent \textbf{Step 3: sampling pairs within estimated communities}
\begin{itemize}[topsep=0pt]
\item[8.] 
Sample uniformly at random pairs within the set $\Acal_{I}$ until $T$ pairs have been sampled overall.
If the number of sampled pairs is smaller than $T$ after all pairs in $\Acal_{I}$ have been sampled$^a$, then sample the remaining pairs at random.
\end{itemize}
\medskip

\noindent \textbf{Output:} $T$ pairs sampled at steps 2.,  7.(b) and 8. of the algorithm. 

\end{minipage}%
\end{minipage} \\
\hline
\end{tabular} }

\subsection{Community Expansion versus \texorpdfstring{$k$}{k} out of \texorpdfstring{$m$}{m} Best Arm Identification}\label{sec:discuss:step2}
As proved in Lemma~\ref{th_step1} in the Appendix~\ref{sec:upper}, after Step 1, with high probability, we end up with a set of $N$ classified nodes, where at most $O(1/s)$ of them are misclassified, and the empirical connectivity $\hat \tau$ does not deviate from the population one $\tau=(p+q)/2$ by more than $(p-q)/4$. The goal of Step 2 is then to identify $\sqrt{2T}$ new nodes of community 1, with at most $O(1/s)$ misclassified nodes and a regret at most $O(\sqrt{T}/s)$. Let us connect this problem to a $k$ out of $m$ best arms identification problem.

Let us consider a simplified version of the problem of Step 2. Assume that we have identified $N_{1}=N/2$ nodes of community 1 with no error, that we have access to the population connectivity $\tau$ and that among the $M=8\sqrt{2T}$ nodes in $\mathcal{A}_{0}$, half of them are of community 1.
Then, each node $a\in \mathcal{A}_{0}$ can be seen as an arm, and pulling the arm $a$ amounts to query a pair $\{a,b\}$ with $b$ one of the $N_{1}$ nodes of community 1 identified at Step 1.
The mean reward of the arm $a$ is $p$ if it belongs to  community 1, and $q$ otherwise.
Hence, a simplified version of the problem in Step 2 amounts to identify $k=\sqrt{2T}$ out of $m=M/2=4\sqrt{2T}$ best arms, with at most $O(1/s)$ errors, and a cumulated regret $O(k/s)$. 
We have the additional constraint that an arm can be pulled at most $N_{1}$ times, but we will forget  this additional feature in this discussion, for simplicity of the comparison.

The problem of identifying $k$ out of $m$ best arms with a tolerance $\epsilon$ has been investigated in \citep{pmlr-v24-goschin12a, koutofm2019}. 
The focus on these papers is on the minimal sample size needed to identify $k$ arms whose expected reward is larger than the $m^{\rm th}$ largest expected reward minus $\epsilon$.
The main results of \citep{koutofm2019} states that, with probability at least $1-k^{-2}$, the algorithm AL-Q-FK can recover with a sample size
\begin{equation*}
O\pa{\frac{1}{(p-q)^2}\pa{M\log\pa{ \frac{m+1}{m+1-k}}+k\log(k)}}
\end{equation*}
$k$ out of the $m$ best arms with a tolerance $\epsilon=(p-q)/2$.
The sampling regret is not considered and it can be as large as the sample size.
In the same setting, the screening algorithm of Step 2 achieves the following performance.  
For $m\geq ck\geq c'/s$, with probability at least $1-c''k^{-2}$ a budget of at most $O(ks^{-1}\log(sk))$ queries, and a sampling regret at most $O(k/s)$, the algorithm identifies a set of arms with at least $k$ out of $m$ best arms and at most $O(1/s)$ arms not in the $m$ best ones. 
As $s=(p-q)^2/(p+q)$, the sampling regret achieved by the screening algorithm of Step 2 is at least $(p+q)/\log(k)$ times smaller. 
We can explain this gain by several reasons.
The  $p+q$ improvement comes from the fact that we explicitly take into account the fact that the rewards have a Bernoulli distribution. 
The $1/\log(k)$ improvement is  obtained by a careful design of the algorithm to keep the regret low, at the price of possibly $O(1/s)$ identification errors. 

Specified to the simplified version of the problem in Step 2 depicted above, the AL-Q-FK algorithm would return $k=\sqrt{2T}$ nodes out of the $m=M/2=4\sqrt{2T}$ nodes of community 1 with a sampling regret 
\begin{equation*}
O\pa{\frac{\sqrt{T}}{(p+q)s}\ \log(\sqrt{T})}.
\end{equation*}
This sampling regret is larger than the $O(\sqrt{T}/s)$ regret needed for our Step 2, so the AL-Q-FK cannot be used as a black-box for Step 2.

We emphasize also that the expansion of the communities in Step 2 is somewhat more complex than the simplified version described above: at Step 1, up to $O(1/s)$ nodes are misclassified, we only have access to the empirical connectivity $\hat \tau$, an arm can only be pulled $N_{1}$ times and the number of best arms is random.

We also emphasize that we cannot use the algorithm of \citep{YP14a} as a black-box to identify $\sqrt{2T}$ nodes of community 1 within $\mathcal{A}_{0}$ with at most $1/s$ errors and with a sampling regret $O(\sqrt{T}/s)$.
Indeed, if we take $\Theta(\sqrt{T})$ nodes and apply the procedure of \citep{YP14a} to classify them   with a sampling regret 
at most $O(\sqrt{T}/s)$, then a \emph{fixed} proportion of the nodes are misclassified  and pairing them together  at Step 3 would generate a final regret of order $\Theta(T)$. 
In addition, from the lower bounds in \citep{YP14a}, we observe that the above phenomenon occurs, whatever the algorithm, if we try to classify \emph{all} the nodes in $\mathcal{A}_{0}$. 
To overcome this issue, the algorithm of Step 2  recovers the class for  a \emph{fraction} only of the nodes in $\mathcal{A}_{0}$ with a sampling regret  at most $O(\sqrt{T}/s)$ and at most $1/s$ errors. 
When recovering  the class of $\sqrt{2T}$ nodes within $\mathcal{A}_{0}$, we do not sample pairs at random, but we carefully select them in order to avoid as much as possible the sampling of bad pairs.

%
%

\section{Constrained Optimal Pair-Matching}
\label{sec:contraint}
\subsection{Main Results}
Let us now consider the general problem, where sparse sampling {\bf (SpS)} is enforced. 
The algorithm described in Section \ref{sec:algo} for unconstrained pairs-matching uses extensively the opportunity to make ``localized" queries: At time $T$, a small number of $\Theta(\sqrt{T})$ nodes has been queried a large number of $\Theta(\sqrt{T})$ times, while other nodes have been queried less than $O(\log(s\sqrt{T})^2/s)$ times.
So, the strategy has to be adapted to fulfils {\bf (SpS)}.

For a sparsity bound $B_{T}$, denote by  $\Psi_{B_{T},T}$ the set of strategies $\psi$ fulfilling the Non-redundancy {\bf (NR)}, Invariance to labelling {\bf (IL)} and Sparse sampling {\bf (SpS)} properties at time $T$.

\begin{thm}
\label{thm:contraint}
Let $T$ and $n$ be positive integers with $T \leq |\cE^{\text{good}}|=2\binom{n/2}{2}$. Let $p,q \in [0,1/2]$ be two  parameters fulfilling (\ref{eq:pq}) and such that the parameter $s$, defined in \eqref{eq:def:ScalingParam}, fulfils
\[
s \leq \frac{1}{32(1+\rho^*)}.
\]
Then, for any $\mu \in \text{cSBM}(n/2,n/2,p,q)$,

\begin{equation*}
\inf_{\psi \in \Psi_{B_{T},T}}  \E_\mu \cro{N^{bad}(\psi,T)} \geq \frac{1}{32} \cro{\frac{\sqrt{T} \vee (T/B_{T})}{32(1+ \rho^*) s}\wedge T}.
\end{equation*}
Conversely, there exist two numerical constants $c_1,c_2>0$ such that, for any time horizon $T$ and constraint $B_T$ satisfying $1 \leq T \leq c_1 n(B_{T}\wedge n)$, there exists a strategy $\psi \in \Psi_{B_{T},T}$ corresponding to a polynomial-time algorithm, described in Section~\ref{sec:algo_contraint}, 
such that 
\begin{equation}\label{eq_upperbound_constrained}
\E_\mu \cro{N^{bad}(\psi,T)} \leq c_2\cro{\frac{\sqrt{T}\vee (T/B_{T})}{s}\wedge T}.
\end{equation}
\end{thm}

We refer to the appendix for a proof of this theorem.
The lower bound is proved in Section~\ref{sec:lower} and the upper bound in Section~\ref{section::proof:constrainedTHM}.

Compared with Theorem~\ref{thm:non-contraint}, Theorem~\ref{thm:contraint} shows that the sparse sampling constraint {\bf (SpS)} amounts to replace $\sqrt{T}$ by $\sqrt{T}\vee(T/B_{T})$ in the optimal sampling-regret. 
In particular, the sparse sampling constraint downgrades optimal rates only when $B_{T}$ is smaller than $\sqrt{T}$.
Actually, a close look at the unconstrained algorithm page \pageref{alg:unconstrained} reveals that,  by construction, it satisfies assumption \textbf{(SpS)} with $B_T = 17\sqrt{T}$. So, in the regime where $B_T \geq 17 \sqrt{T}$, 
the lower bound
cannot be worse than the upper-bound of the unconstrained setting of Theorem~\ref{thm:non-contraint}.

When $B_{T} \lesssim \sqrt{T}$, the optimal sampling-regret is of order $(T/(B_{T}s))\wedge T$. 
This rate can be understood as follows. 
If $B_{T}\leq 1/s$, there is not enough observations per node to infer their community better than at random, which induces an unavoidable linear regret. 
When $B_{T}\gg 1/s$, to proceed as in Step 3 of the constrained case, one needs to identify a sufficiently large set of nodes of the same community, among which one can sample up to $T$ pairs without adding regret.
As each node can now be paired with at most $B_{T}$ others, this set should be of size $\Theta(T/B_{T})$ instead of $\Theta(\sqrt{T})$ in the unconstrained case. 
As the identification of the community of a node requires at least $\Theta(1/s)$ queries, the sampling-regret expected to identify this large set of nodes is $\Theta(T/(B_{T}s))$.

The previous informal discussion suggests to extend the algorithm described in Section~\ref{sec:algo} for the unconstrained case.
This extension, fully described and commented in Section \ref{sec:algo_contraint}, still proceeds in $3$ steps and goes as follows.
The first step of the constrained algorithm is essentially the same as the first step of the unconstrained algorithm, with $\sqrt{T}$ replaced by 
$B=(B_{T}\wedge \sqrt{T})/2$. 
In this first step, all pairs are sampled among a set of $B/\log(sB)\leq B_T$ nodes, so the constraint cannot be violated.
Then,  to keep the sampling-regret under control while not violating the {\bf (SpS)} contraint, the trick is to apply recursively a variant of the screening algorithm in Step~2 and repeat these screenings until a total number of $\Theta(T/(B_{T}\wedge \sqrt{T}))$ nodes are correctly classified, with a small proportion of error.
Finally, one can sample at most $B_T\wedge \sqrt{T}$ pairs for each of these nodes in Step 3 with a controlled regret. 
The resulting algorithm extends the unconstrained one of Section~\ref{sec:algo} where $B_T\wedge \sqrt{T}=\sqrt{T}$  and where  the screening step is only applied once.
This extension is fully described in Section~\ref{sec:algo_contraint}.

To illustrate the theorem, one can discuss the results with the constraint $B_{T}=T^\gamma$, where $0<\gamma\leq 1/2$. As mentioned in the introduction, this situation arises with $\gamma=1-1/\alpha$ when  $T=n^{\alpha}$, and when, for fairness reasons, the algorithm is required to sample at most $B_{T}=cT/n=cT^{1-1/\alpha}$ times each node.
In this case, the optimal sampling-regret is of order $T\wedge (T^{1-\gamma}/s)$ which becomes, in the example discussed in introduction the mentioned rate $T\wedge T^{1/\alpha}/s$. 
It follows that any pair-matching algorithm that is $T^\gamma$-sparse up to time $T$ (besides satisfying {\bf (NR)} and {\bf (IL)}) has linear sampling-regret up to time $s^{-1/\gamma}$.
On the other hand, there exist strategies with optimal sampling-regret of order $T^{1-\gamma}/s$ after time $s^{-1/\gamma}$. 

Notice that the sparse sampling property $N_{a}(\psi,T)\leq B_{T}$ only constrains the algorithm at the time horizon $T$.
This time horizon has therefore to be specified beforehand for this constraint to be defined. 
In many practical situations, this specification is not reasonable and a more realistic constraint takes the form: $N_{a}(\psi,t)\leq B_{t}$ at any time $t\in \ac{1,\ldots,T}$. 
In the case where $B_{t}=\Theta(t^{\gamma}/(\log t)^{\tau})$, the constraint can be enforced using a doubling trick, without enlarging the regret by more than a multiplicative numerical constant.
This doubling trick is discussed in detail in Section~\ref{sec:doubling-contraint}.

\subsection{Algorithm with Sparse Sampling}\label{sec:algo_contraint}
The algorithm described page \pageref{alg:unconstrained} achieves optimal regret in the unconstrained case.
It identifies first a set of $\Theta(\sqrt{T})$ nodes from one community with $O(1/s)$ misclassified nodes and a regret of order $O(\sqrt{T}/s)$ in Steps 1 and 2.
Then, it pairs these nodes together in Step 3 with a $O(\sqrt{T}/s)$ regret (due to the misclassified nodes).

The algorithm described in this section follows essentially the same steps.
It identifies first a  set of nodes from a single community (with small error) and then samples pairs among them.
It has to be adapted to fulfil the  {\bf (SpS)} constraint.
As the unconstrained algorithm fulfils the {\bf (SpS)} constraint for any $B_{T}\geq 17\sqrt{T}$, it is assumed in the remaining of this section that $B_{T}=O( \sqrt{T})$. 
Moreover, as the result holds for $T \leq c_1 n(B_{T}\wedge n)$, this assumption is granted in the remaining of the section.

To respect the constraint {\bf (SpS)}, no node may be sampled in more than $B_{T}$ pairs. 
Hence, to perform the last step, the algorithm has to identify $\Theta(T/B_{T})$ nodes from one community.
It should achieve this identification with a sampling-regret smaller than $O(T/(sB_{T}))$ while respecting the {\bf (SpS)} constraint. 
To respect the {\bf (SpS)} constraint in the first step of the algorithm, a core-set $\Ncal_{init}$ of cardinality smaller than $B_{T}$ is chosen. 
Formally, in points 1. and 2. of Step 1 in the algorithm page \pageref{alg:unconstrained}, $\sqrt{T}$ is replaced by $(B_{T}\wedge \sqrt{T})/2$. 
Then, as in the unconstrained case, Step 2 expands the communities in order to identify, with high probability and up to a small error,  $\Theta(T/B_{T})$ nodes from one community. 
The main difference with the unconstrained case is that this expansion cannot be achieved in a single step of screening. 
Actually,
\begin{itemize}
\item[(i)] $\Theta(N/s)$ pairs are required to identify the community of $\Theta(N)$ new nodes. 
 \item [(ii)] Any node from the core-set $\Ncal_{init}$ cannot be sampled more than $B_{T}$ times.
 \end{itemize}
By (ii), one cannot sample more than  $O(|\Ncal_{init}|B_{T})$ pairs and by (i), it follows that at most $O(|\Ncal_{init}|B_{T}s)=O(B_{T}^2s)$ nodes can be classified with a single screening step based on $\Ncal_{init}$. 
The main idea of the new algorithm is to iterate the screening step, expanding progressively the communities. 
Along these iterations, to satisfy the {\bf (SpS)} constraint, the screening has to be conducted with more care than in step 2 of the unconstrained algorithm page \pageref{alg:unconstrained}.
The trick is to apply the  \texttt{SCREENING} function described page \pageref{alg:screening}, which compartmentalizes the nodes in order to enforce the condition {\bf (SpS)}. 
This iterative process outputs a set of $\Theta(T/B_{T})$ nodes from a single community (with  a small proportion of error with high probability).
The algorithm finally pairs nodes among this subset while respecting the {\bf (SpS)} constraint in Step 3 of the algorithm.
\newpage

\medskip

%
%
%

\noindent { \def\arraystretch{1.3}
\begin{tabular}{|l|}
\hline
 \textbf{Constrained Algorithm}\label{alg:constrained} \\
\hline
\begin{minipage}{0.99\textwidth} \centering
\begin{minipage}{0.97\textwidth}

\medskip
\noindent \textbf{Inputs:} $s$ scaling parameter, $T$ time horizon, $V_{init}$ the set of the $n$ nodes of the whole graph, $B_T$ constraint.

\medskip
\noindent \textbf{Internal constants:} set $c_{\Ocal_0} = 8 \vee (1 / c_1^\text{GC})$ and $B = (B_T \wedge \sqrt{T}) / 2$. 

\medskip
\noindent \textbf{Step 1: finding communities in a core-set}
\begin{enumerate}[topsep=0pt]
\item Sample uniformly at random an initial set $\Ncal_{init} \subset V_{init}$ of $N_{init} = \left\lceil \frac{B}{\log(sB)} \right\rceil$ nodes. 

\item Sample each pair of $\Ncal_{init}$ with probability $c_{\Ocal_0} \frac{B}{s}/\binom{N_{init}}{2}$, call $\Ocal_0 \subset \mathcal{E}$ the output.

\item Estimate mean connectivity $\tau=\frac{p+q}{2}$ by $\hat{\tau}=\frac{1}{|\Ocal_0|} \sum_{(x,x') \in \Ocal_0} A_{x,x'}.$

\item Run \texttt{GOODCLUST} on the graph $(\Ncal_{init}, \Ocal_{0})$ and output, for any $x \in \Ncal_{init}$, $\widehat{Z}_{x}$ the estimated community of $x$ (with the convention that the largest estimated community is labelled by 1).
\end{enumerate}

\medskip
\noindent \textbf{Step 2: iteratively expanding the communities}

\medskip
\noindent \textbf{Internal constants:} set $N^{(0)} = \left\lceil N_{init}/2 \right\rceil$,
\begin{equation}
t_f = \left\lceil \frac{ \log (\lceil 2 T / B \rceil / N^{(0)}) }{ \log \lfloor \log (sB) \rfloor } \right\rceil
\end{equation}
and for all $t \in \{0, \dots, t_f\}$,
\begin{equation}
N^{(t)} =   N^{(0)} \lfloor \log (sB) \rfloor^t  \wedge \left\lceil \frac{2 T}{B} \right\rceil.
\end{equation}

\begin{enumerate}[topsep=0pt, resume]
\item Let $\Ncal^{(0)}$ be a set of $N^{(0)}$ nodes in $\Ncal_{init} \cap \{\widehat{Z} = 1\}$ sampled uniformly at random,  and let $V^{(0)} = V_{init} \setminus \Ncal_{init}.$

\item \label{point_screening_algcontraint} \textbf{For} $t = 1, \ldots, t_f $, {\bf set}

\begin{equation}
(\Ncal^{(t)}, V^{(t)}) = \texttt{SCREENING}\pa{
	\Ncal^{(t-1)},
	N^{(t)},
	B,
	\hat{\tau},
	V^{(t-1)}
}.
\end{equation}

\end{enumerate}

\medskip
\noindent \textbf{Step 3: sampling pairs within estimated communities}

\begin{enumerate}[topsep=0pt, resume]
\item Sample pairs within the set $\Ncal^{(t_f)}$ while respecting the constraint \textbf{(SpS)} with $B_T$, until $T$ pairs have been sampled overall (the sampling method does not matter).
\end{enumerate}

\medskip

\end{minipage}%
\end{minipage} \\
\hline
\end{tabular} }

\noindent { \def\arraystretch{1.3}
\begin{tabular}{|l|}
\hline
 \textbf{Function} \texttt{SCREENING}$(\Ncal, N', B, \nu, V) = (\Ncal', V')$ \label{alg:screening} \\
\hline
\begin{minipage}{0.99\textwidth} \centering
\begin{minipage}{0.97\textwidth}

\medskip
\textbf{Inputs:} a reference core-set $\Ncal$ of cardinality $N$, a target number of nodes $N'$, a constraint $B \in \R_+$, a threshold $\nu \in [0,1]$, a set of ``new" nodes $V$.

\medskip
\textbf{Output:} a set of nodes $\Ncal'\subset V$ of cardinality at most $N'$ and the set of nodes $V'\subset V$ that are still ``new" after running \texttt{SCREENING}.
(Most of the nodes of $\Ncal'$ will belong to the most represented community in $\Ncal$.)

\medskip
\textbf{Internal constants:} a number of pairs per step $k = \lceil \frac{C_k}{s} \rceil$ and a number of steps $I = \lceil C_I \log(sB) \rceil$, with $C_k =2500$ and $C_I = 1026$.

\medskip
\begin{enumerate}[topsep=0pt]
\item Sample uniformly at random a set $\Acal_0$ of $|\Acal_0| = 4N'$ nodes in $V$.

\item \label{SCREEN_step2} Let $m = \lfloor N / (kI) \rfloor$. Take a uniform partition of $\Ncal$ into $m$ sets $(\Vcal_j)_{1 \leq j \leq m}$ of cardinality $kI$ and one set of cardinality smaller than $kI$.

Likewise, take a uniform partition of $\Acal_0$ into $m$ sets $(\Acal_0^{(j)})_{1 \leq j \leq m}$ with cardinality in $\{\lfloor 4N'/m \rfloor, \lceil 4N'/m \rceil \}$.


\item \label{step_4_contraint} 
\textbf{For} $j = 1, \dots, m$ \textbf{and} $i = 1, \dots, I$, \textbf{do}
\begin{itemize}
\item[] \textbf{For each} $x \in \Acal^{(j)}_{i-1}$, \textbf{do}
	\begin{itemize}[topsep=0pt]
	\item[i.] Sample $k$ nodes $(y_{k(i-1)+a}^x)_{a=1,\ldots,k}$ uniformly at random in $\Vcal_j \setminus \{y_{a}^x\}_{a=1,\ldots,k(i-1)} $.
	\item[ii.] Sample  pairs $(\{x,y_{k(i-1)+a}^x\})_{a=1,\ldots,k}$  and compute
	\begin{equation}
	\hat p_{x,i}=\frac{1}{ki} \sum_{a=1}^{ki} A_{xy_{a}^x}.
	\end{equation}
	\end{itemize}
\item[iii.] Select $\Acal^{(j)}_{i} = \ac{x \in \Acal^{(j)}_{i-1}: \hat p_{x,i} \geq \nu}$. 
\end{itemize}

\item \label{step_5_contraint} Set $\Ncal'$ a set of $N' $ nodes sampled uniformly at random from $\displaystyle \bigcup _{1 \leq j \leq m}\Acal^{(j)}_I$.\\* 
{\bf In case}\footnote{with high probability, this undesirable case does not happen} $|\bigcup _{1 \leq j \leq m}\Acal^{(j)}_I|< N'$, {\bf then} sample at random $N'$ nodes in $\Acal_0$.
\item Set $V' = V \setminus \Acal_0.$
\end{enumerate}

\textbf{Return} $(\Ncal', V')$.

\medskip
\end{minipage}%
\end{minipage} \\
\hline
\end{tabular} }

\subsection{Screening versus \texorpdfstring{$k$}{k} out of \texorpdfstring{$m$}{m} Best Arms Identification}
Similarly as in Section \ref{sec:discuss:step2}, let us compare the screening step to a $k$ out of $m$ best arms identification problem. 
The main additional feature compared to the situation discussed in Section \ref{sec:discuss:step2}, is that an arm $a$ cannot be sampled more than $B$ times. 
Hence, a simplified version of the screening problem amounts to identify $k$ out of $m$ best arms with tolerance $\epsilon=(p-q)/2$, with the constraint that each arm cannot be sampled more than $B$ times. 
In these simplified setting, the screening function achieves the following performance. 
Assume that $M\geq ck$ and $k,B\geq c'/s$.
With probability $1-c(sk)^{-1}$ , with a budget of $O(ks^{-1}\log(s(B\wedge k)))$ queries, and with a sampling regret at most $O(k/s)$, the screening function identifies at least $k$ arms of community 1 with at most $O((k(sB)^{-1})\vee s^{-1})$ errors.

The situation handled by the screening function is actually somewhat more complex than the stylized bandit problem depicted above. Actually, among the initial set of $N$ classified nodes, we have up to  $cN/(sB)$  misclassified nodes. 
At the same time, we cannot query more than $B$ times any of these classified nodes. 
Hence, we need a careful querying policy in order to avoid the misclassified nodes to generate errors, while keeping the {\bf (SpS)} condition enforced. 
Fulfilling together these two conditions is the main hurdle in the design and analysis of the screening function.

\subsection{Pathwise Sparse Sampling Algorithm}\label{sec:doubling-contraint}
The algorithm presented above fulfils the sparse sampling condition {\bf (SpS)} at time horizon $T$. 
In many practical situations, it is more natural to consider Condition {\bf (SpS)} at all times $t=1,2,\ldots$ rather than only at a predefined time horizon $t=T$. 
Formally, Condition {\bf (SpS)} would be replaced by $N_{a}(\psi,t)\leq B_{t}$, for all $t=1,2,\ldots$. 
It is possible to modify the previous algorithm to build a strategy $\psi$ such that, when $B_{t}=\Theta(t^{\gamma}\log^{-\tau}(t))$, the sampling regret $\E_\mu \cro{N^{bad}(\psi,t)}$ fulfils 
\begin{equation*}
\E_\mu \cro{N^{bad}(\psi,t)}= O\pa{\frac{\sqrt{t}\vee (t/B_{t})}{s}\wedge t},\quad \textrm{for}\ t=1,2,\ldots.
\end{equation*}
Assume that there exist $\gamma\in (0,1/2]$ and $\tau \in [0,+\infty)$ such that $B_{t}=t^{\gamma}/(\log t)^{\tau}$, so $\sqrt{t}\vee(t/B_{t})=t^{1-\gamma}\log^{\tau}(t)$. 
In this case, a pathwise sampling condition can be enforced using the simple doubling trick. 
For any positive integer $l$, let $t_l = 2^l$. 
At each time $t_l$, the new algorithm discards all nodes and pairs previously sampled and starts the algorithm of Section \ref{sec:algo_contraint} with the remaining nodes, time horizon $T=t_{l+1}-t_{l}$ and terminal sparse sampling constraint $N_{a}(\psi,t_{l+1}-t_{l})\leq \min_{t_{l}\leq t\leq t_{l+1}}B_{t}$. 
The resulting strategy does not depend on any time horizon and it fulfils the condition $N_{a}(\psi,t)\leq B_{t}$, for all $t=1,2,\ldots$. 

Moreover, for any $l$ such that $t_{l}\geq e^{\tau/\gamma}$, $\min_{t_{l}\leq t\leq t_{l+1}}B_{t}=B_{t_{l}}$. 
Hence, for any $l$ such that $t_{l}<c_1 n(B_{t}\wedge n)$ and for any $t$ such that $t_{l-1}\leq t \leq t_{l}<c_1 n(B_{t}\wedge n)$,
\begin{align*}
    \mathbb{E}\cro{N^{bad}(\psi,t)} &=O\pa{1+\sum_{k=1}^l\frac{\pa{t_k-t_{k-1}}^{1-\gamma}\log^{\tau}(t_{k}-t_{k-1})}{s}\wedge (t_{k}-t_{k-1})}\\
    &=O\pa{\pa{\frac{1}{s} \sum_{r=0}^{l-1}2^{r(1-\gamma)} (r\log(2))^{\tau}}\wedge t_{l}}\\
    &= O \pa{\frac{t_{l}^{1-\gamma}\log^\tau(t_{l})}{s}\wedge t_{l}}=O\pa{\frac{t^{1-\gamma}\log^\tau(t)}{s} \wedge t}.
\end{align*}
According to Theorem~\ref{thm:contraint}, the sampling-regret of the algorithm derived from the doubling trick is then rate optimal.

\section{Discussion}\label{Sec:KclassesSBM}
The present paper provides the optimal sampling-regret for pair-matching in
 the case where $G=(E,V)$ is a conditional SBM with a number of groups $K=2$, where the groups have $n/K$ elements, with intra class probability of connection $p$ and inter-class $q$. 
The algorithm depicted p.\pageref{alg:unconstrained} in Section~\ref{sec:algo} runs in polynomial time and has optimal sampling-regret given in Theorem~\ref{thm:non-contraint}, up to a multiplicative constant. Let us discuss the two following questions: How can we estimate the scaling parameter $s$?
 How does the rates depend on the number $K$ of groups?

\subsection{A Heuristic to Estimate the Scaling Parameter \texorpdfstring{$s$}{s}}
\label{sec:SNR}
The algorithms described p.\pageref{alg:unconstrained} and p.\pageref{alg:constrained} in Sections \ref{sec:algo} and \ref{sec:algo_contraint} take the scaling parameter $s$ as input.
This parameter is typically unknown in practice and an estimated value $\widehat s$ has to be plugged in the algorithm.
To  guarantee a sampling-regret smaller than $O(T\wedge (\sqrt{T}/s))$, the estimator $\widehat s$ should use at most $O(1/s^2)$ edges and satisfy $ \widehat s \asymp s$ with high probability. 
The following heuristic builds a possible estimator $\widehat s$.

Pick uniformly at random $N$ nodes in $V$ and sample all $N(N-1)/2$ pairs between these $N$ nodes. When $Ns>2$, $p=a/N$ and $q=b/N$, the results in \citep{MNS15} ensures that, as $N\to\infty$, $a$ and $b$ can be consistently estimated. 
Therefore, $Ns=(a-b)^2/(a+b)$ can also be consistently estimated from these $T=N(N-1)/2=O(1/s^2)$ observations. 
Yet, this estimator requires the knowledge $Ns$ larger than $2$ and cannot therefore be used directly when $s$ is unknown. 

However, when $p=a/N$ and $q=b/N$ and $N\to\infty$, 
it is theoretically possible to detect whether  $Ns=(a-b)^2/(a+b)$ is smaller or larger than 2.
To proceed, denote by $\cB$ the non-backtracking matrix associated to the graph, see \citep{BLM18} for a definition of the non-backtracking matrix. 
Let $\lambda_{1}, \lambda_{2},\ldots$ be the eigenvalues of $\cB$ ranked in decreasing order of their moduli. 
The main result of \citep{BLM18} shows that, when $p=a/N$ and $q=b/N$, with $a,b>0$ fixed, except on an event of vanishing probability as $N\to \infty$, 
\begin{gather*}
 |\lambda_{2}|^2<\lambda_{1}\quad \text{when}\quad Ns<2\enspace,\\
|\lambda_{2}|^2>\lambda_{1} \quad \text{when}\quad Ns>2\enspace.
\end{gather*}
Hence, in this asymptotic setting where $s=\Theta(1/N)$, it is possible to detect if $Ns>2$ by looking at the ratio $|\lambda_{2}|^2>\lambda_{1}$. In addition, when $Ns>2$, the ratio $2|\lambda_{2}|^2/\lambda_{1}$ consistently estimates $(a-b)^2/(a+b)$.

This result suggests the following recursive algorithm to estimate $s$: fix some $\epsilon>0$ and start with  a set $V_1$ of $2$ nodes $i$ and $j$ picked uniformly at random in $V$. Query the pair $\{i,j\}$ and let $E_1$ denote the set of edges in $E\cap \{i,j\}$.
At each step $k\geqslant 2$, pick at random a set $V_k$ of $2^k$ nodes in $V\setminus \cup_{\ell\leqslant k-1}V_{\ell}$.
Sample all pairs in $V_k$, and  denote by $E_k$ the set of edges among these pairs. 
Build the non-backtracking matrix $\cB_k$ of the graph $(V_k,E_k)$ and compute $\lambda^{(k)}_1$ and $\lambda^{(k)}_2$ the eigenvalues of this matrix with largest moduli.
If $|\lambda^{(k)}_{2}|^2<(1+\epsilon)\lambda^{(k)}_{1}$ iterate. If $|\lambda^{(k)}_{2}|^2>(1+\epsilon)\lambda^{(k)}_{1}$ stop, denote by $\widehat{k}$ the stopping iteration time and $\widehat{N}=2^{\widehat{k}}$ the number of nodes 
sampled in the last graph $(V_{\widehat{k}},E_{\widehat{k}})$.
Output $\widehat s=2|\lambda^{(\widehat{k})}_{2}|^2/(\widehat{N}\lambda^{(\widehat{k})}_{1})$.
 

Assume that $p=a/N$ and $q=b/N$ with $a,b\in \R^+$ fulfilling  $(a-b)^2/(a+b)>2$.  Let $\Omega_{N}$ denote the event where simultaneously $2\leq \widehat N s \leq 8(1+\epsilon)$ and $s/2\leq \widehat s \leq 2 s$.
Then the results of \citep{BLM18} suggest that the event $\Omega_{N}$ holds with probability tending to $1$ as $N\to \infty$. 
In addition, the total number of sampled edges is  $\sum_{k=1}^{\widehat{k}} \binom{2^k}2=O(\widehat N^2)=O(1/s^2)$ on this event. 
While the results of  \citep{BLM18} suggest that the procedure should work for vanishingly small $s$ (large $N$ limit), we emphasize  that they only hold in a setting where $p=a/N$, $q=b/N$, with $a,b$ fixed and $N\to \infty$, and we cannot turn them into a theoretical guarantee that $\Omega_{N}$ holds with probability close to $1$. 
We evaluate the performance of this heuristic numerically in Section~\ref{sec_simu_estim_s}.

\subsection{Case with \texorpdfstring{$K>2$}{more than two} Groups} \label{sec:Kgroups}
Let us discuss the case where the number of groups $K$ is larger than $2$, still assuming that all the groups have $n/K$ elements, with intra class probability of connection $p$ and inter-class $q$. 
Contrary to $K=2$, we expect in this case an information-computation gap and conjecture the following optimal rates for pair-matching.

\begin{conj}
Define $\Psi_{\infty}^{poly}$ as the intersection of $\Psi_{\infty}$ defined page \pageref{sec:Unconstrained}, with polynomial-time algorithms. 
Let
\begin{equation}
\label{eq:sk}
s_{K}= \frac{(p-q)^2}{q+(p-q)/K}.
\end{equation}
Under Assumption (\ref{eq:pq}) and $s_{K}\leq 1$, without computational constraint:  
\begin{equation}\label{eq:conj:expo}
\inf_{\psi\in \Psi_{\infty}} \E\cro{N^{bad}(\psi,T)} \asymp \pa{\pa{\frac{K\log(K)}{s_{K}}}^2\vee \frac{K\sqrt{T}}{s_{K}}}\wedge T.
\end{equation}
With polynomial time constraint: 
\begin{equation}\label{eq:conj:poly}
\inf_{\psi\in \Psi_{\infty}^{poly}} \E\cro{N^{bad}(\psi,T)} \asymp  \pa{\pa{\frac{K^2}{s_{K}}}^2\vee \frac{K\sqrt{T}}{s_{K}}}\wedge T.
\end{equation}
\end{conj}

Let us explain the heuristics leading to these rates.\smallskip

For $K=2$, a central tool to design the rate-optimal polynomial-time algorithm p.\pageref{alg:unconstrained} is the existence of polynomial-time algorithms (called {\tt GOODCLUST} p.\pageref{alg:unconstrained}) achieving non trivial classification for a cSBM$(N/2,N/2,p,q)$ when $Ns$ is larger than some constant. 
When $K>2$ and the number of nodes $N\to \infty$,  for $p$, $q$ scaling as $1/N$, the papers \citep{BLM18,AS15c,LM18} provide polynomial-time algorithms $\text{{\tt GOODCLUST}}^{\text{poly}}_K$
achieving a non trivial classification for 
\begin{equation*}
Ns_{K}>K^2=:\lambda_{K}^{poly}.
\end{equation*}
Furthermore, it is conjectured in \citep{Decelle2011} that there does not exist any polynomial-time algorithm achieving non-trivial classification when $Ns_{K}<K^2$. 
The threshold $\lambda^{poly}_{K}$ is known as the Kesten-Stigum (KS) threshold.
{While the conjecture of \citep{Decelle2011} is relative to the case where $K$ is fixed, $p=a/N$, $q=b/N$, and $N$ goes to infinity, this conjecture has been recently supported  non-asymptotically by a  Low-Degree polynomial lower bound  \citep{LowDegree2022,luo2023computational} whenever $K^2\leq N$. This supports the conjecture that, when $s_{K}\leq 1$, non-trivial clustering is possible only when $N= \Theta(K^2/s_{K})$.}

The information theoretic threshold $\lambda_{K}^{inf}$ for non-trivial classification is below $\lambda_{K}^{poly}$ for $K\geq 5$. 
Actually, the paper \citep{BanksMoore2016} proved that $\lambda_{K}^{inf}\asymp K\log(K)$ and $\lambda_{K}^{inf}<\lambda_{K}^{poly}$ for $K\geq 5$, so, if the conjecture of \citep{Decelle2011} holds, there is an information-computation gap for $K\geq 5$.
A consequence of the result of \citep{BanksMoore2016} is that there exist algorithms $\text{{\tt GOODCLUST}}^{\text{inf}}_K$, with exponential complexity,  achieving non-trivial classification for $Ns_{K}=O(K\log(K))$. 

Theorem~\ref{thm:non-contraint} requires that $\text{{\tt GOODCLUST}}$ has more than non-trivial classification, it should have vanishing classification error. 
Several papers have established, under Assumption (\ref{eq:pq}), the existence of algorithms $\text{{\tt GOODCLUST}}^{\text{poly}}_K$ and $\text{{\tt GOODCLUST}}^{\text{inf}}_K$ with misclassification proportion  smaller than $\exp(-c Ns_{K}/K)$,  for some positive constant $c$.  
This result is obtained for $Ns_{K}\geq c' \lambda_{K}^{poly}$ for $\text{{\tt GOODCLUST}}^{\text{poly}}_K$, see for example \citep{CRV15,GMZZ17,FC17,GV18} and for $Ns_{K}\gg \lambda_{K}^{inf}$ for $\text{{\tt GOODCLUST}}^{\text{inf}}_K$, see \citep{ZZ16}.

As a consequence, without computational constraint, a linear sampling regret is expected for any algorithm as long as the time horizon satisfies $\sqrt{2T} s_{K}< \lambda_{K}^{inf}$, or equivalently 
\begin{equation*}
T< 0.5 (\lambda^{inf}_{K}/s_{K})^2=0.5(K\log K /s_{K})^2.
\end{equation*}
On the other hand, when $T\gg (K(\log K)^2/s_{K})^2$, one can choose $N$ fulfilling $\lambda^{inf}_{K}/s_{K} \ll N\leq (K\sqrt{T}/s_{K})^{1/2} \ll \sqrt{T}$. 
Selecting $N$ nodes uniformly at random and 
 observing all pairs of these $N$ nodes,  {\tt GOODCLUST}$^{inf}_{K}$ classifies correctly the $N$ nodes, but a proportion at most $\exp(-cN s_{K}/K)$ of them. 
The sampling-regret for this step does not exceed the number $O(N^2)=O(K\sqrt{T}/s_{K})$ of pairs sampled. 
Since $N s_{K}/K \gg \log(K)$, the proportion of misclassified nodes among these $N$ nodes is small and a screening procedure  as in Step~2 of the algorithm p.\pageref{alg:unconstrained} can be applied  in order to classify correctly $\sqrt{T}$ nodes. As an average of $K/s_{{K}}$ queries is necessary to classify one new node, this step will have a regret scaling as $K\sqrt{T}/s_{K}$. Then, we can pair all nodes of the same group until the budget of $T$ queries is spent.  Hence, in the regime where $T\gg (K(\log K)^2/s_{K})^2$,
  the final regret should  be proportional to $N^2+K\sqrt{T}/s_{{K}}\asymp K\sqrt{T}/s_{{K}}$. 
To sum-up the discussion, without computational constraints, one can expect a sampling-regret of order
\begin{equation*}
\left((K\log(K)/s_{{K}})^2\vee K\sqrt{T}/s_{{K}}\right)\wedge T\,,
\end{equation*}
which is the conjectured rate (\ref{eq:conj:expo}).

Using polynomial time algorithms for clustering, the information-theoretic threshold $\lambda^{inf}_{K}$ should be replaced by the KS-threshold $\lambda^{poly}_{K}$. 
Following the same reasoning as before, linear regret is expected as long as
\begin{equation*}
T< 0.5 (\lambda^{poly}_{K}/s_{K})^2=0.5(K^2 /s_{K})^2.
\end{equation*}
On the other hand, when $\sqrt{T}\gg K^3/s_{K}$, 
one can pick $N$ nodes at random with $N$ fulfilling $\lambda^{poly}_{K}/s_{K} \ll N\leq (K\sqrt{T}/s_{K})^{1/2} \ll \sqrt{T}$.
A polynomial time algorithm $\text{{\tt GOODCLUST}}^{\text{poly}}_K$ run with all pairs based on these nodes classifies correctly these $N$  nodes, except for a proportion at most $\exp(-cN s_{K}/K)$ of them. 
The sampling-regret associated to this classification step is smaller than $N^2\leq K\sqrt{T}/s_K$.
The screening step classifies correctly $\sqrt{T}$ nodes with a regret $K\sqrt{T}/s_{K}$.
The remaining budget until sampling $T$ pairs is spent by pairing together nodes in a same estimated group. 
Ultimately, taking into account the computational constraint, one can expect a sampling-regret of order $((K^2/s_{K})^2\vee K\sqrt{T}/s_{K})\wedge T$, which is the conjectured rate  (\ref{eq:conj:poly}).

\subsection{Unbalanced partitions and pairwise dependent probabilities}
Although we use a simple random graph model for the ease of exposition, our analysis can be extended to a more general set-up which relaxes the following assumptions: (i) the graph has only two groups; (ii) the two groups have the same number  $n/2$ of nodes; (iii) the probabilities $p$ and $q$ of (intra and inter-group) connection are constants.  For the relaxation of  (i), we refer the reader to the Section~\ref{sec:Kgroups}, where we discuss conjectures  in a SBM with $K$ groups, for any integer $K\geq 2$.

	We can relax assumption (ii) by assuming that $|G_1| =c n$ and $|G_2|=(1-c)n$ with $c$ a numerical constant in  $(0,1)$. Among the ten points in Lemma~\ref{th_step1} and~\ref{th_step2}, one can readily check that only the point 5 about the estimation of $(p+q)/2$ does not hold anymore. Indeed, for $c\neq 1/2$, the global connectivity of the graph is not equal to $(p+q)/2$, and thus our estimator $\hat{\tau}$ of the global connectivity is not a consistent estimator  of $(p+q)/2$. A solution is  to replace $\hat{\tau}$ by   $\hat{\tau}'= (\hat{p}+\hat{q})/2$  where the estimators $\hat{p}$ and $\hat{q}$ of $p$ and $q$ are obtained from the following two-step procedure. One run \texttt{GOODCLUST} on a first set of sampled  pairs in the core-set;  then conditionally to the estimated group labels $\widehat{Z}_x$, one estimate $p$ by sampling new intra-group pairs (i.e.  pairs $(x,y)$ satisfying $\widehat{Z}_x = \widehat{Z}_y$) and one estimate $q$ by sampling new inter-group pairs ($\widehat{Z}_x \neq \widehat{Z}_y$).  


Instead of the model assumption (iii), let us consider the situation where the probability $p_{ij}$ of connection between the nodes $i$ and $j$ belongs to  $[p_0,p]$  if $i$ and $j$ are in the same group, while $p_{ij} \in [q,q_0]$ otherwise, with $q_0< p_0$. Turnkey clustering algorithms work in such situations $-$ see  \citep{ZZ16} for example. Then, still with the idea of having two distinct groups of nodes, assume that the intra group variations  $\delta:= (p-p_0) \vee (q_0-q)$ are smaller than the inter group separation $\Delta:= p_0-q_0$. Under this assumption, one can readily check that our estimator $\hat{\tau}$ of the global connectivity satisfies $|\hat{\tau} -(p_0+q_0)/2| \leq \Delta /4$ (i.e. the point 5 in Lemma~\ref{th_step1}), and that  all the other points in Lemma~\ref{th_step1} and~\ref{th_step2} still hold. In this situation of non-constant probabilities of connection,  note that the link  between the regret and the sampling regret (seen in Section \ref{Sec:Obj}) does not hold anymore, and we only control the sampling regret.

%
%
%

\section{Numerical experiments}
\label{sec_simulations}

\subsection{Unconstrained setting}
\label{sec_sims_unconstrained}

In this section, we empirically assess the sharpness of our unconstrained algorithm from Section~\ref{sec:algo} on the following parameters:
\begin{itemize}
\item the budget $T$ is taken in $\{50, 100,200,500, 1\,000,2\,000,5\,000, 10\,000,20\,000, 50\,000,$

$100\,000,200\,000,500\,000, 1\,000\,000,2\,000\,000,5\,000\,000\}$,

\item the connection probabilities are taken as $(p,q) = (0.6,0.4), (0.7,0.3), (0.55,0.45)$ or $(0.4,0.2)$, which corresponds to $s = 0.04, 0.16, 0.01,$ or $0.0666...$ respectively,

\item the groups are either balanced, or one group contains $20\%$ of the individuals while the other contains $80\%$,

\item the pool of available individuals is assumed infinite (\emph{i.e.} $n$ infinite), which means that one may freely add new individuals and decide their class independently of the classes of previously sampled individuals,

\item for each choice of parameter, 10 experiments are performed to obtain an averaged regret.
\end{itemize}

A few adaptations are required for the implementation to work efficiently and for all values of $T$. For the unconstrained algorithm,
\begin{itemize}
\item if $\log(s \sqrt{T}) < 1$, it is replaced by 1 in the formulas of the size $N$ of the core-set and of the number $I$ of steps in Step 2,
\item if the size of the core-set $N$ is smaller than the size $2kI$ required to have at least one estimated class large enough to perform step 2, $N$ is replaced by $2kI$ (more generally $KkI$ when there are $K$ groups),
\item \verb?GOODCLUST? classifies the individuals by applying the Lloyd algorithm to the first 2 eigenvectors of the trimmed-adjacency matrix of the graph (more generally to the first $K$ eigenvectors), see \cite{CRV15},
\item the values of the constants are fixed to $c_0 = 2$, $C_I = 10$, $C_k = 0.1$, and the number of individuals in the core-set is taken as $N = \lceil c_\Ncal \sqrt{T} / \log(s\sqrt{T}) \rceil$ with $c_\Ncal = 2$,
\item instead of sampling a set $\Acal_0$ in step 2 and pruning the individuals with low connectivity, we instead proceed individual by individual, discarding them if their connectivity dips below the threshold at one point and adding them to $\Acal_I$ if not.
\end{itemize}
These modifications are intended to make the algorithm work more efficiently and for smaller values of $T$ than the ones presented in the theorems above, without changing its theoretical properties.

\begin{figure}
\centering
\includegraphics[scale=.55]{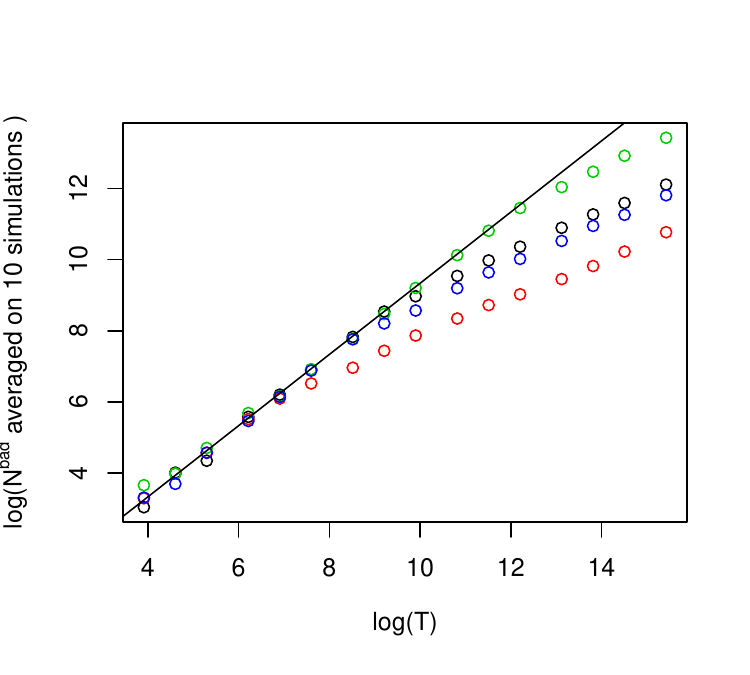}
\includegraphics[scale=.55]{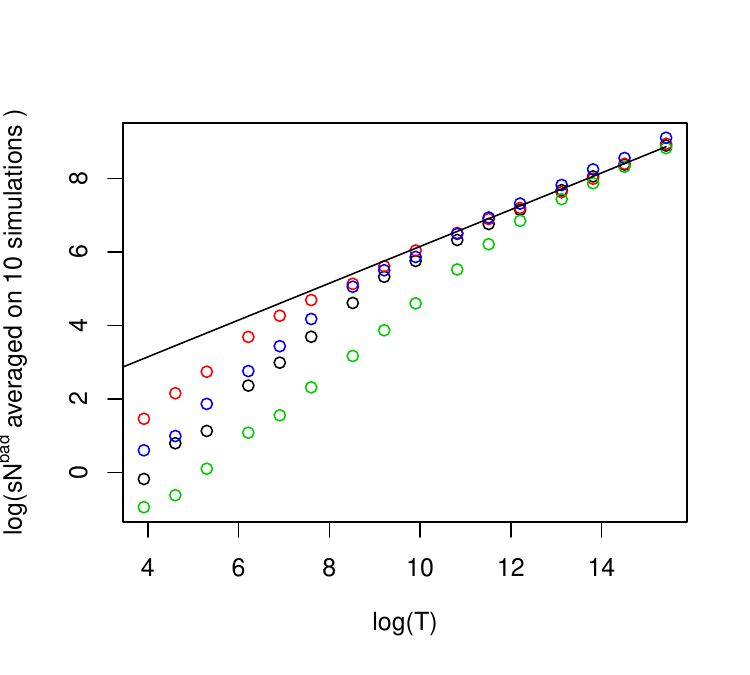}
\caption{Average number of errors $\bar{N}^\text{bad}$ over 10 simulations in the unconstrained case with balanced communities. The graphs show $\log(\bar{N}^\text{bad})$ (left) and $\log(s \bar{N}^\text{bad})$ (right) as a function of $\log(T)$, confirming the two regimes (linear in $T$ and proportional to $\sqrt{T}/s$). The lines have slope 1 and $1/2$ respectively. Green: $s=0.01$, black: $s=0.04$, blue: $s=0.06666...$, red: $s=0.16$.}
\label{fig_unconstrained}
\end{figure}

\begin{figure}
\centering
\includegraphics[scale=.55]{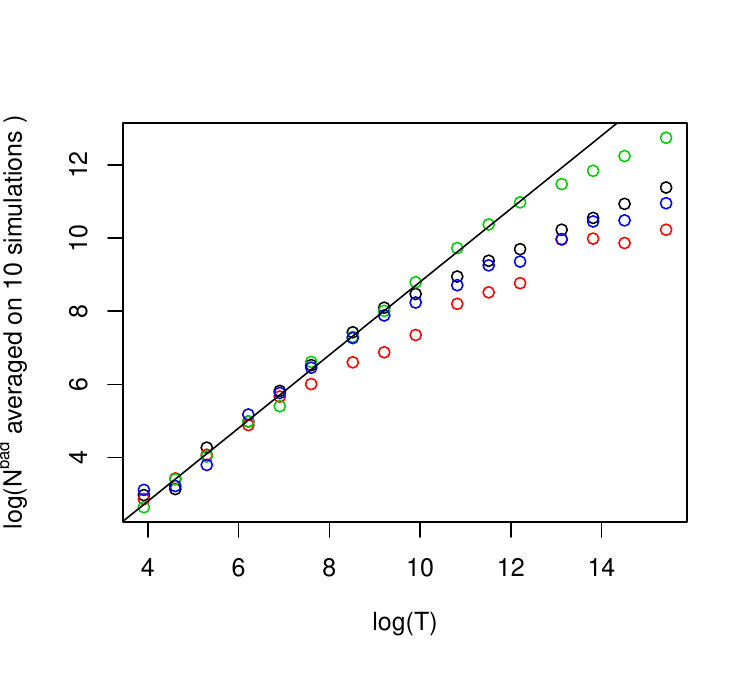}
\includegraphics[scale=.55]{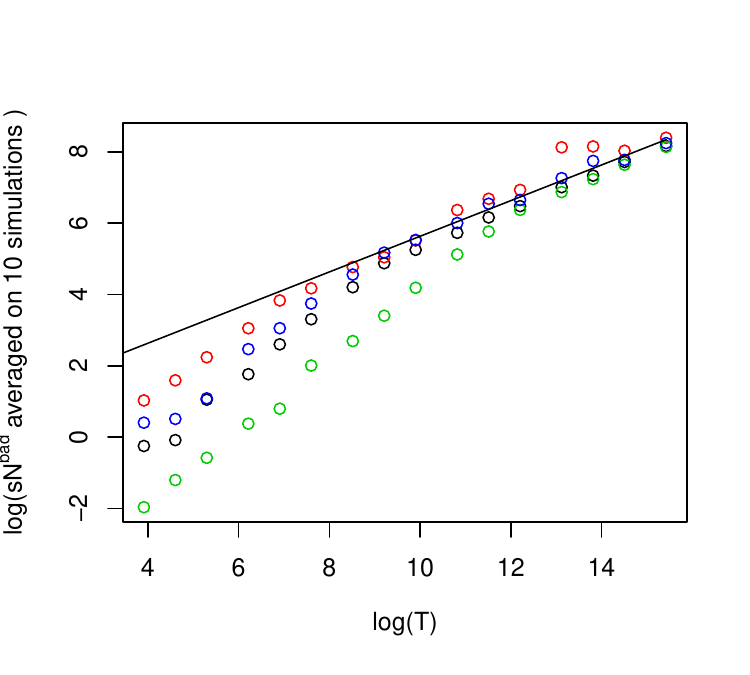}
\caption{Average number of errors $\bar{N}^\text{bad}$ over 10 simulations in the unconstrained case with a 80\%-20\% split between communities. The graphs show $\log(\bar{N}^\text{bad})$ (left) and $\log(s \bar{N}^\text{bad})$ (right) as a function of $\log(T)$. The lines have slope 1 and $1/2$ respectively. Green: $s=0.01$, black: $s=0.04$, blue: $s=0.06666...$, red: $s=0.16$.}
\label{fig_unconstrained_unbalanced}
\end{figure}

The results for the unconstrained algorithm are displayed in Figure~\ref{fig_unconstrained} (balanced communities) and Figure~\ref{fig_unconstrained_unbalanced} (unbalanced communities). In both cases, as predicted by Theorem~\ref{thm:non-contraint},  we observe an initial linear regime for the regret for $T=O(1/s^2)$, followed by a $\sqrt{T}/s$ regime for $T$ large compared to $1/s^2$.

\subsection{Constrained setting}

\begin{figure}
\centering
\includegraphics[scale=.55]{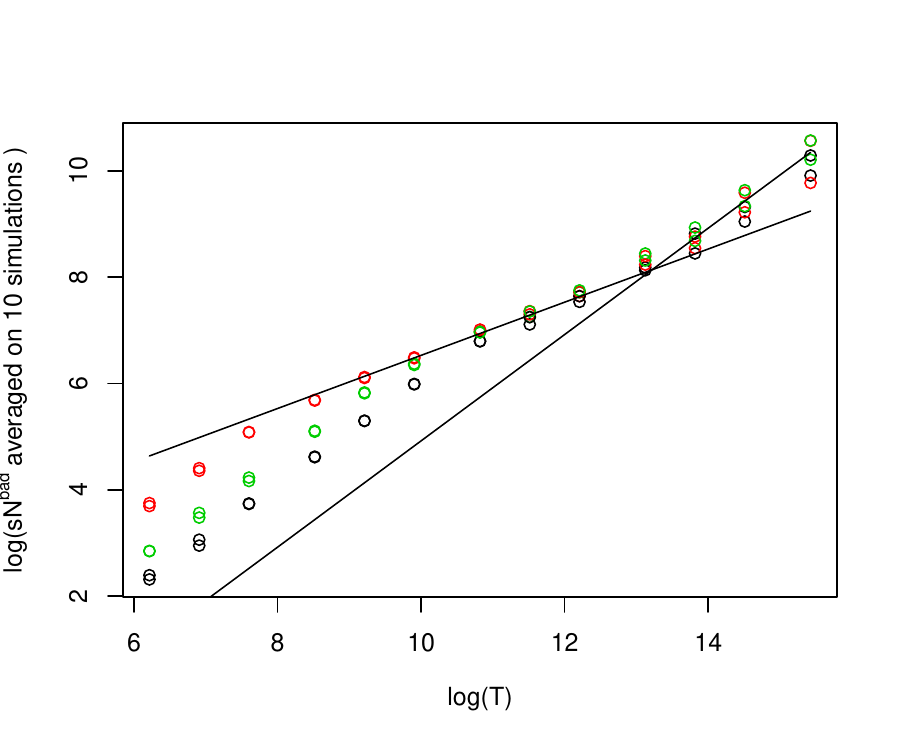}
\caption{Average number of errors $\bar{N}^\text{bad}$ over 10 simulations in the constrained case with balanced communities. The graphs show $\log(s \bar{N}^\text{bad})$ as a function of $\log(T)$. The lines have slope $1/2$ and $1$ respectively. This confirms the three regimes of Theorem~\ref{thm:contraint}: linear in $T$ for small and large $T$ and proportional to $\sqrt{T}/s$ in between. Black: $s=0.04$, green: $s=0.06666...$, red: $s=0.16$.}
\label{fig_constrained_balanced}
\end{figure}

In this section, the constrained algorithm from Section~\ref{sec:algo_contraint} is assessed, the results are displayed in Figure~\ref{fig_constrained_balanced}. The parameters are the same than in the previous section, with the difference that we only consider balanced communities and budgets $T$ greater or equal to $500$. The constraint $B_T$ is taken in $\{500, 1\,000\}$.

In addition to the adaptations from the previous section, we modified the algorithm as follows:
\begin{itemize}
\item if $\log(s \sqrt{T}) < 1$, since this corresponds to the linear case, the edges are sampled at random,
\item the values of the constants are fixed to $c_0 = 8$, $C_I = 10$, $C_k = 0.1$, and the number of individuals in the core-set is taken as $N = \lceil c_\Ncal \sqrt{T} / \log(s\sqrt{T}) \rceil$ with $c_\Ncal = 4$,
\item 
if the size of the core-set is smaller than $5/2 \times c_0/s = 20/s$, which ensures that the constraint is satisfied with high probability, $N$ is replaced by $20/s$.
\end{itemize}
Again, we observe empirically the three regimes from Theorem~\ref{thm:contraint}: the sampling regret is first linear in $T$, then proportional to $\sqrt{T}/s$ and then linear again as $T/(Bs)$.

\subsection{Estimation of the Scaling Parameter \texorpdfstring{$s$}{s}}
\label{sec_simu_estim_s}

In this section, we investigate the empirical performance of the heuristic procedure from Section~\ref{sec:SNR} for estimating $s$.
As $s\to 0$, the two key properties that we are looking for, are
$$2< \hat N s =O(1) \quad\textrm{and}\quad \hat s =\Theta(s).$$
We focus on the empirical evaluation of these two properties.

For non-vanishing values of $s$, random fluctuations of the eigenvalues of the non-backtracking matrix  can blur the asymptotic results of  \citep{BLM18}. We evaluate the noise level with the modulus of the third eigenvalue $|\lambda_{3}^{(k)}|$, whose asymptotic value remains smaller than $\sqrt{\lambda_{1}^{(k)}}$ when $s$ is vanishingly small. 
We changed the stopping criterion accordingly: the algorithm stops when $|\lambda_3^{(k)}|^2 < (1+\epsilon) \lambda_1^{(k)}$ and $|\lambda_2^{(k)}|^2 \geq (1+\epsilon) \lambda_1^{(k)}$, with $\epsilon = 0.1$. The additional condition $|\lambda_3^{(k)}|^2 < (1+\epsilon) \lambda_1^{(k)}$ is meant to ensure that the stopping condition $|\lambda_2^{(k)}|^2 \geq (1+\epsilon) \lambda_1^{(k)}$ is not due to pure random fluctuations.
This modification does not change the asymptotic properties of the estimator or the asymptotic of the number of edges sampled, that is $O(1/s^2)$ when $s \rightarrow 0$.

\begin{figure}
\centering
\includegraphics[scale=.49]{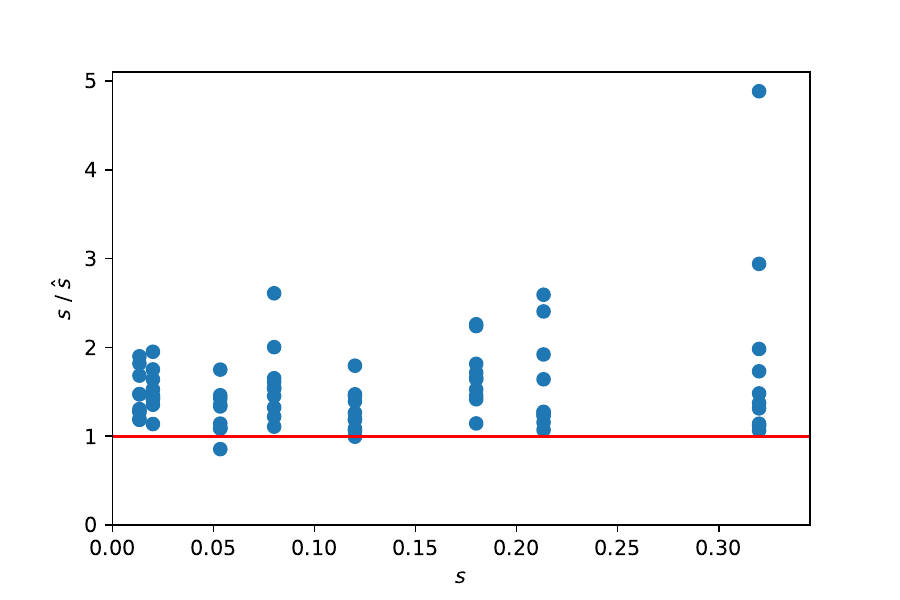}
\includegraphics[scale=.49]{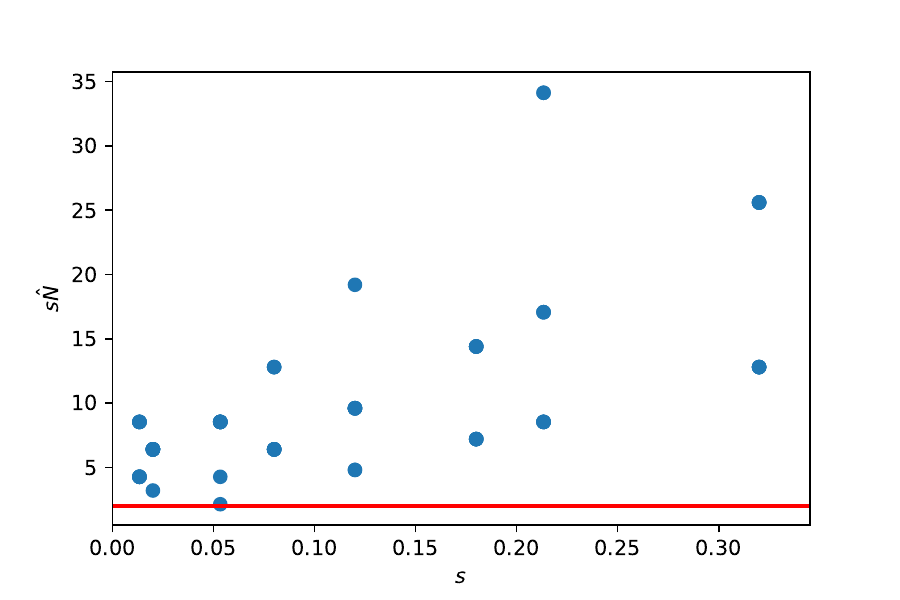}
\caption{On the left, ratio of the true scaling parameter $s$ to the estimated scaling parameter $\hat{s}$ as defined in Section~\ref{sec:SNR}. The red line corresponds to $s/\hat{s} = 1$. On the right, product $s\hat{N}$ where $\hat{N}$ is the number of nodes at the stopping time. The red line corresponds to $s\hat{N} = 2$.}
\label{fig_estimation_s}
\end{figure}

The parameters taken were balanced groups with 
$(p,q) = (0.45,0.05)$, $(0.4,0.1)$, $(0.35,0.15)$, $(0.3,0.2)$, $(0.3,0.1/3)$, $(0.8/3,0.2/3)$, $(0.7/3,0.1)$, or $(0.2,0.4/3)$. For each value of $(p,q)$, the ratio $s/ \hat s$ and the product $\hat N s$ are displayed in Figure~\ref{fig_estimation_s} for 10 simulations.

We used the SuNBEaM package from~\cite{TorresSE19}, available at~\cite{TorresSuNBEaM}, to compute the eigenvalues of the non-backtracking matrix. The results are displayed in Figure~\ref{fig_estimation_s}. The left-hand side figure shows that the estimator $\hat{s}$ is always comparable to $s$, and almost always smaller than $s$, which makes it a good choice to replace $s$ by $\hat{s}$ in the pair-matching algorithm.
Finally, the right-hand side figure shows that the condition $s\hat{N} > 2$ is always satisfied, thus ensuring that $\hat{s}$ consistently estimates $s$.

\subsection{Resilience to model misspecification}
\label{sec_misspe}

{The above simulations use data generated by a cSBM model. We now investigate how the algorithm's performance is affected when the data does not exactly follow a cSBM distribution.}

{We consider the following alternative model. Let $0 < p,q < 1$ be connectivity parameters as above, as well as a misspecification parameter $\sigma \in [0,1/2)$. When sampling a new node $i$, it is affected the label $Z_i = 0$ or 1 with probability 1/2. In addition to its label, it is assigned a hidden state $S_i \sim \text{Unif}([Z_i - \sigma, Z_i + \sigma])$. The probability that two nodes $i$ and $j$ are connected is given by}
\begin{equation*}
\mathbb{P}(A_{ij} = 1 | Z_i, Z_j, S_i, S_j) = p + (q-p) \left( |S_i - S_j| - \frac{2 \sigma}{3} \right).
\end{equation*}
{The case $\sigma = 0$ is the usual, well specified, case. In general, for any $\sigma$, $\mathbb{P}(A_{ij} = 1 | Z_i = Z_j) = p$ and $\mathbb{P}(A_{ij} = 1 | Z_i \neq Z_j) = q$, that is, the expected connectivity between members of the same class (resp. between members of different classes) is the same as in the well specified setting. Where the distribution differs is in the joint distribution of more than two nodes: if two nodes are connected, chances are good their hidden states are closer than average, thus making it more likely that if a new node connects to one of them, then it will connect to the other.}

\begin{figure}
\centering
\includegraphics[scale=.7]{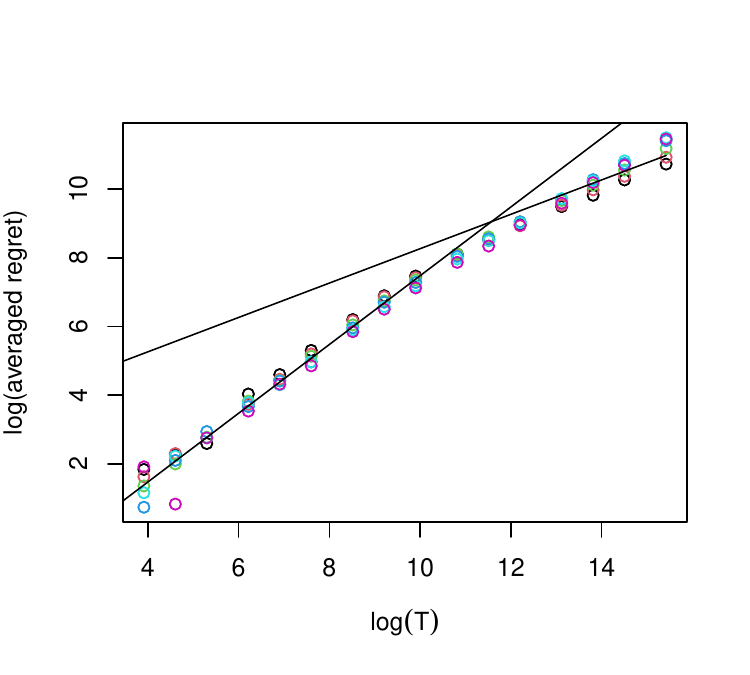}
\caption{{Logarithm of the average regret $\frac{1}{10} \sum_{k=1}^{10} \left( pT - \sum_{t=1}^T A_{\widehat{e}_t}^{(k)} \right)$ in the misspecified case, as a function of $\log(T)$. The lines have slope $1/2$ and $1$ respectively. Black: $\sigma=0$, red: $\sigma=0.1$, green: $\sigma=0.2$, blue: $\sigma=0.3$, cyan: $\sigma=0.4$, violet: $\sigma=0.5$. The two regimes from the well specified case (linear then proportional to $\sqrt{T}$) are still visible.}}
\label{fig_misspe_regret}
\end{figure}

{We take the same possible values of $T$ as in Section~\ref{sec_sims_unconstrained}, with balanced communities, $(p,q) = (0.6,0.4)$, and an infinite pool of individuals and misspecification values $\sigma \in \{0, 0.1, 0.2,$ $0.3, 0.4\}$. 10 experiments are performed for each choice of parameters, and the average regrets computed over them, are displayed in Figure~\ref{fig_misspe_regret}. The two regimes from the well specified case (linear then proportional to $\sqrt{T}$) are still visible.}

\appendix

\section{Proof of the Lower Bounds}
\label{sec:lower}
\subsection{Distributional Properties under Assumption {\bf (IL)}}
Recall that $\cE$ denotes the set of all pairs in $\ac{1,\ldots,n}$.
The invariance to labelling property enforces some invariances on the distribution of the $(N_{e}(\psi,T):e\in \cE)$, with $N_{e}(\psi,T)$ defined by (\ref{def:Ne}) and on the distribution of the $(N_{a}(\psi,T):a=1,\ldots,n)$ with $N_{a}(\psi,T)$ defined by (\ref{def:Na}). 

Let $\mu$ be a distribution in cSBM$(n/2,n/2,p,q)$ associated to a partition $G=\ac{G_{1},G_{2}}$ of $\ac{1,\ldots,n}$. 
Consider a permutation $\sigma$ which leaves the partition $G$ invariant, that is such that, either $\sigma(G_{1})=G_{1}$ and hence $\sigma(G_{2})=G_{2}$, or $\sigma(G_{1})=G_{2}$ and thus $\sigma(G_{2})=G_{1}$. 
Then, the distribution $\mu^{\sigma}$ defined page~\pageref{def:mu_sigma} is equal to the distribution $\mu$. Hence the invariance to labelling property ensures that for any  permutation $\sigma$ leaving $G$ invariant, the vectors $(N_{e}(\psi,t):e\in \cE;t=1,\ldots,\binom{n}{2})$ and $(N_{\sigma(e)}(\psi,T):e\in \cE;t=1,\ldots,\binom{n}{2})$ have the same distribution. As a consequence, the following properties holds.

\begin{lem}\label{lem:invariance}
When the strategy $\psi$ fulfils the invariance to labelling property, then  the random variables $(N_{e}(\psi,T):e\in \cE^{good})$ are pair-wise exchangeable.  The same property holds for $(N_{e}(\psi,T):e\in \cE^{bad})$ and $(N_{a}(\psi,T):a=1,\ldots,n)$. 
\end{lem}

\noindent{\bf Proof.}
Let $\ac{a,b},\ac{a',b'}$ denote two pairs in $\cE^{good}$ and let $\sigma$  be  a $G$-invariant permutation such that  $\sigma(\ac{a,b})=\ac{a',b'}$, and $\sigma(\ac{a',b'})=\ac{a,b}$. Since $\mu=\mu^{\sigma}$ and $\psi$ is invariant to labelling, the random variables $(N_{\ac{a,b}},N_{\ac{a',b'}})$ and $(N_{\ac{a',b'}},N_{\ac{a,b}})$ have the same distribution. The same reasoning applies for pairs in $\cE^{bad}$.

Consider now two nodes $a,b\in\ac{1,\ldots,n}$. 
Let  $\sigma$ be   a $G$-invariant permutation on $\ac{1,\ldots,n}$ such that $\sigma(a)=b$ and $\sigma(b)=a$. 
Since $\mu=\mu^{\sigma}$ and $\psi$ is invariant to labelling, the random variables  $(N_{a}(\psi,T),N_{b}(\psi,T))$ and $(N_{b}(\psi,T),N_{a}(\psi,T))$ have the same distribution. \hfill $\square$

\subsection{Proof of the Lower Bound in Theorems~\ref{thm:contraint} and~\ref{thm:non-contraint}}
This section contains the proof of the first part of Theorem~\ref{thm:contraint}. The first part of Theorem~\ref{thm:non-contraint} follows by taking $B_{T}=T$.

{Let $kl(p,q)=p\log(p/q)+(1-p)\log((1-p)/(1-q))$ be the Kullback-Leibler divergence between two Bernoulli distributions with parameters $p$ and $q$.}
We actually prove the following stronger lower bound: when $\tilde s=kl(p,q)\vee kl(q,p)$ satisfies $\tilde s\leq 1/16$, for any 
$\mu \in \text{cSBM}(n/2,n/2,p,q)$,
\begin{equation}
\label{eq:strong_lower_bound}
\inf_{\psi \in \Psi_{B_{T},T}}  \E_{\mu}\cro{N^{bad}(\psi,T)} \geq \frac{1}{32} \cro{\frac{\sqrt{T} \vee (T/B_{T})}{16 \tilde s}\wedge T}.
\end{equation}
The first part of Theorem \ref{thm:contraint} follows from this bound and from
Lemma \ref{lem:kullback} which ensures that 
$s \leq \tilde s\leq 2(1+\rho^*) s $ when (\ref{eq:pq}) holds.

Recall that $N_{a}(\psi,T)$ denotes the number of pairs involving the node $a$ sampled by the strategy $\psi$  up to time $T$.
Let $N_{a}^{bad}(\psi,T)$ be the number of pairs $\{a,b\}$ with $b$ not in the community of $a$ sampled up to time $T$. 
Hereafter in the proof, the strategy $\psi$ is fixed and, to simplify notations, the dependency of $N_a$ and $N_a^{\text{bad}}$ on $\psi$ is dropped out: $N_{a}^{bad}(\psi,T)$ is denoted $N_{a}(T)$ and $N_{a}^{bad}(\psi,T)$ is denoted $N_{a}^{bad}(T)$.
Let also $N_{a}^{good}(T)=N_{a}(T)-N_{a}^{bad}(T)$.
The number of between-group sampled pairs is
\begin{equation*}
N^{bad}(T)=\frac{1}{2}\sum_{a=1}^n {N_{a}^{bad}(T)}.
\end{equation*}
Let us also recall that $N_{\ac{a,b}}(\psi,T)\in \ac{0,1}$ (denoted $N_{\ac{a,b}}(T)$), is the number of times the pair $\{a,b\}$ has been sampled before time $T$. Likewise, let $N_{aB}(T) = \sum_{b \in B} N_{\ac{a,b}}(T)$ be the number of times a pair between node $a$ and the set of nodes $B$ has been sampled before time $T$.
For $t\geq 0$, let $\cF_{t}$ be the $\sigma$-algebra gathering information available up to time $t$: $\cF_{t}$ is the $\sigma$-algebra generated by $(\widehat\cE_{t}, (A_{e})_{e\in \widehat\cE_{t}},U_{0},\ldots,U_{t})$.

The main tools for proving Equation~\eqref{eq:strong_lower_bound} are the next two lemmas.
The first lemma is directly adapted from \citep{GMS18,KCG16}. 
{It is a derivative of the data processing inequality, and it merely states that the Kullback-Leibler divergence between two distributions of observation stopped at some stopping time is larger than the Kullback-Leibler divergence of processed versions of these distributions.}
\begin{lem}\label{lem:fonda}
Let $\tilde T$ be a stopping time with respect to the filtration $(\cF_{t})_{t\geq 0}$. Let $\mu, \mu' \in cSBM(n/2,n/2,p,q)$  and let $\nu=(\nu_{ab})_{a<b}$ and $\nu'=(\nu'_{ab})_{a<b}$ denote their connection probabilities, that is $\nu_{ab} = \mu(\{a,b\} \in E)$ and $\nu_{ab} = \mu'(\{a,b\} \in E)$ for all $a,b \in V$.
If $\tilde T \leq T$  a.s., then for any $\cF_{\tilde T}$-measurable random variable $\Zcal$ taking values in $[0,1]$,
\begin{equation}\label{eq:fonda}
\sum_{a<b}\E_{\mu}[N_{\ac{a,b}}(\tilde T)]kl(\nu_{ab},\nu'_{ab}) \geq kl(\E_{\mu}[\Zcal],\E_{\mu'}[\Zcal]),
\end{equation}
where $kl(p,q)=p\log(p/q)+(1-p)\log((1-p)/(1-q))$ is the Kullback-Leibler divergence between two Bernoulli distributions with parameters $p$ and $q$.
\end{lem}

\noindent {\bf Proof.} The lemma follows directly from Lemma 1 in \citep{KCG16} and Lemma 1 in  \citep{GMS18}. As discussed in Section \ref{sec:bandits}, the pair-matching problem can be seen as a bandit problem with restrictions on the set of admissible strategies. Since Lemma 1 in \citep{KCG16} and Lemma 1 in \citep{GMS18} hold for any strategy, Inequality (\ref{eq:fonda}) holds in particular for any strategy $\psi$ satisfying the constraints $\psi_{t}(\widehat{\Ecal}_{t}, \dots) \notin \widehat\cE_{t}$ and $N_{a}(t)\leq B_{T}$.  \hfill $\square$

While the previous lemma is only based on the bandit nature of the problem, the next lemma is based on the constraint that arms can only be sampled once. 

\begin{lem}\label{lem:sum-M}
Let $M$ be a positive real number and consider $T\geq 1$. 
Then
\begin{equation*}
\sum_{a=1}^n (N_{a}(T) \wedge M) \geq \pa{(M\sqrt{T})\vee \frac{MT}{B_{T}}} \wedge \frac{T}{2}.
\end{equation*}
\end{lem}

\noindent{\bf Proof of Lemma \ref{lem:sum-M}.}
Let $S_{1}=\ac{a: N_{a}(T)\leq M}$ and $S_{2}=\ac{a: N_{a}(T)> M}$.

\noindent
If $\sum_{a\in S_{1}} N_{a}(T)\geq {T/2}$ then $\sum_{a=1}^n (N_{a}(T) \wedge M) \geq \sum_{a\in S_{1}} N_{a}(T)\geq {T/2}$.\medskip

\noindent
Assume now that $\sum_{a\in S_{1}} N_{a}(T)<T/2$. Since $2T=\sum_{a=1}^n N_{a}(T)$,\begin{align*}
2T \leq T/2+ \sum_{a\in S_{2}} N_{a}(T)&= T/2+ \sum_{a\in S_{2}} N_{a S_{1}}(T)+ \sum_{a\in S_{2}} N_{a S_{2}}(T) \\
&=T/2+ \sum_{a\in S_{1}} N_{a S_{2}}(T)+ \sum_{a\in S_{2}} N_{a S_{2}}(T)\\
&\leq T + |S_{2}|(B_{T}\wedge |S_{2}|).
\end{align*}
Hence, $|S_{2}|\geq \sqrt{T}\vee (T/B_{T})$ and 
\begin{equation*}
\sum_{a=1}^n (N_{a}(T) \wedge M) \geq |S_{2}| M \geq (M\sqrt{T})\vee (MT/B_{T}).
\end{equation*}
The proof is complete. \hfill $\square$
\bigskip

With these two lemmas, the core inequality of the proof can be established.
This inequality shows that if $N_{a}(t)=O(1/\tilde s)$, then $N_{a}^{bad}(t)$ is of the same order of magnitude than $N_{a}(t)$. 

Let $G = (G_1,G_2)$ be a partition of $\ac{1,\ldots,n}$ with $G_{1}=\ac{1,\ldots,n/2}$ and $G_{2}=\ac{n/2+1,\ldots,n}$. Let $\mu \in cSBM(n/2,n/2,p,q)$ be the distribution of a conditional SBM with classes $G_1$ and $G_2$, within-group connection probability $p$ and between-group connection probability $q$. Unless specified, $\E=\E_{\mu}$ in the following. 
\begin{lem}\label{lem:inter}
Let $M$ be a positive integer such that $16M\tilde s\leq 1$ and define the stopping time $\tilde T=T\wedge \inf\ac{t:\max(N_{1}(t),N_{n}(t))\geq M}$.
Setting $N_{1+n}(T)=N_{1}(T)+N_{n}(T)$ and $N_{1+n}^{bad}(T)=N_{1G_{2}}(T)+N_{nG_{1}}(T)$,
\begin{equation}\label{ineq:main}
\E\cro{N_{1+n}^{bad}(\tilde T)} \geq \frac{1}{4} \E\cro{N_{1+n}(\tilde T)}\geq \frac{1}{4} \E\cro{N_{1}(T)\wedge M}.
\end{equation}
\end{lem}

\noindent{\bf Proof of Lemma \ref{lem:inter}.} {Some arguments of this proof are inspired by \citep{GMS18}.}
The last inequality in (\ref{ineq:main}) follows directly from
\begin{equation*}
{N_{1+n}(\tilde T)} \, \geq \, {N_{1}(T)}{\bf 1}_{ \tilde T =  T} + M {\bf 1}_{ \tilde T < T} \, \geq \, {N_{1}(T)\wedge M}.
\end{equation*}

It remains to show the first inequality. 
Consider the transposition $\sigma=(1,n)$ of $1$ and $n$ which switches the labels $1$ and $n$ while keeping other nodes unchanged.
Let $\mu^\sigma$ be the distribution of $(A_{\sigma(a),\sigma(b)})_{ab}$. The partition $G^\sigma=\ac{G^\sigma_1, G^\sigma_2}$ associated to $\mu^\sigma$, corresponds to $G$ with $1$ and $n$ switched, that is $G^\sigma_{1}=\ac{n,2,\ldots,n/2}$ and $G^\sigma_{2}=\ac{n/2+1,\ldots,n-1,1}$.

Let $M$ be a positive integer and set
\begin{equation*}
\Zcal=\frac{N_{1G_{2}}(\tilde T)+N_{nG_{1}}(\tilde T)}{2(M\wedge B_{T})} \in [0,1].
\end{equation*}
By invariance to labelling, 
\begin{align*}\E_{\mu^\sigma}\cro{N_{1G_{2}}(\tilde T)+N_{nG_{1}}(\tilde T)}&=\E_{\mu^\sigma}\cro{N_{1G^\sigma_{2}}(\tilde T)+N_{nG^\sigma_{1}}(\tilde T)+2N_{\ac{1,n}}(\tilde T)} \\
&=\E_{\mu}\cro{N_{1G_{1}}(\tilde T)+N_{nG_{2}}(\tilde T)+2N_{\ac{1,n}}(\tilde T)}.
\end{align*}
Hence, setting $\tilde M=M\wedge B_{T}$, Lemma \ref{lem:fonda} ensures that,
\begin{align*}
(&kl(p,q)\vee  kl(q,p)) \E_{\mu}\cro{N_{1}(\tilde T)+N_{n}(\tilde T)} \\
&\geq kl\pa{\E_{\mu}\cro{N_{1G_{2}}(\tilde T)+N_{nG_{1}}(\tilde T)}/(2\tilde M),\E_{\mu^\sigma}\cro{N_{1G_{2}}(\tilde T)+N_{nG_{1}}(\tilde T)}/(2\tilde M) }\\
&= kl\pa{\E_{\mu}\cro{N_{1G_{2}}(\tilde T)+N_{nG_{1}}(\tilde T)}/(2\tilde M),\E_{\mu}\cro{N_{1G_{1}}(\tilde T)+N_{nG_{2}}(\tilde T)+2N_{\ac{1,n}}(\tilde T)}/(2\tilde M) }\\
&\geq \frac{1}{2(M\wedge B_{T})} \frac{\pa{\E_{\mu}\cro{N_{1G_{2}}(\tilde T)+N_{nG_{1}}(\tilde T)}-\E_{\mu}\cro{N_{1G_{1}}(\tilde T)+N_{nG_{2}}(\tilde T)+2N_{\ac{1,n}}(\tilde T)}}^2}{\E_{\mu}\cro{N_{1G_{2}}(\tilde T)+N_{nG_{1}}(\tilde T)}\vee \E_{\mu}\cro{N_{1G_{1}}(\tilde T)+N_{nG_{2}}(\tilde T)+2N_{\ac{1,n}}(\tilde T)}},
\end{align*}
where the last line follows from Lemma \ref{lem:kullback}. 
Setting  $N_{1+n}^{good}(T)=N_{1G_{1}}(T)+N_{nG_{2}}(T)$, the last inequality can be written as
\begin{multline}
\label{ineq:main2}
2(M\wedge B_{T})\tilde s \E\cro{N_{1+n}(\tilde T)}\pa{\E\cro{N_{1+n}^{good}(\tilde T)+2N_{\ac{1,n}}(\tilde T)}\vee \E\cro{N_{1+n}^{bad}(\tilde T)}} \\
	\geq \pa{\E\cro{N_{1+n}^{good}(\tilde T)+2N_{\ac{1,n}}(\tilde T)}-\E\cro{N_{1+n}^{bad}(\tilde T)}}^2.
\end{multline}
If $\E\cro{N_{1+n}^{good}(\tilde T)}\leq \E\cro{N_{1+n}^{bad}(\tilde T)}$, then
\begin{equation*}
2\E\cro{N_{1+n}^{bad}(\tilde T)}\geq \E\cro{N_{1+n}^{good}(\tilde T)}+ \E\cro{N_{1+n}^{bad}(\tilde T)}=\E\cro{N_{1+n}(\tilde T)}
\end{equation*}
and Lemma \ref{lem:inter} follows.

Assume therefore that $\E\cro{N_{1+n}^{good}(\tilde T)}\geq \E\cro{N_{1+n}^{bad}(\tilde T)}$. 
It follows that 
\begin{equation*}
2\E\cro{N_{1+n}^{good}(\tilde T)}\geq \E\cro{N_{1+n}^{good}(\tilde T)}+ \E\cro{N_{1+n}^{bad}(\tilde T)}=\E\cro{N_{1+n}(\tilde T)},
\end{equation*}
so Inequality (\ref{ineq:main2}) implies
\begin{equation*}
4(M\wedge B_{T})\tilde s \E\cro{N_{1+n}^{good}(\tilde T)+2N_{\ac{1,n}}(\tilde T)}^2 \geq \pa{\E\cro{N_{1+n}^{good}(\tilde T)+2N_{\ac{1,n}}(\tilde T)}-\E\cro{N_{1+n}^{bad}(\tilde T)}}^2.
\end{equation*}
Rearranging the expression gives
\begin{align*}
\E\cro{N_{1+n}^{bad}(\tilde T)} &\geq \E\cro{N_{1+n}^{good}(\tilde T)+2N_{\ac{1,n}}(\tilde T)} \pa{1- \sqrt{4(M\wedge B_{T})\tilde s}}\\
&\geq \frac{1}{2} \E\cro{N_{1+n}^{good}(\tilde T)} \geq \frac{1}{4} \E\cro{N_{1+n}(\tilde T)},
\end{align*}
since $M\wedge B_{T}\leq 1/(16\tilde s)$ by assumption.
The proof is complete. \hfill $\square$
\bigskip

The lower bound in Theorem \ref{thm:contraint} can now be proved.
Recall that for any strategy $\psi \in \Psi_{B_{T},T}$, Assumption {\bf (IL)} implies that the sampling-regret $\E_{\mu}\cro{N^{bad}(\psi,T)}$ does not depend on
$\mu \in \text{cSBM}(n/2,n/2,p,q)$, see the remark page~\pageref{page:inv-sampling}.
Therefore, it is sufficient to prove (\ref{eq:strong_lower_bound}) for any strategy $\psi$ invariant by labelling, with the distribution $\mu$ defined above Lemma~\ref{lem:inter}. 

Let $M$ be a positive integer such that
\begin{equation*}
1\leq M\wedge B_{T} \leq \frac{1}{16\tilde s}.
\end{equation*}
First, Lemma~\ref{lem:invariance} ensures that, for any pair $\ac{a,b}\in\cE^{bad}$, 
$\E\cro{N_{\ac{a,b}}(T)} = \E\cro{N_{\ac{1,n}}(T)}$
 and hence 
\begin{equation*}
\E\cro{N^{bad}(T)} = \frac{n^2}{4} \E\cro{N_{\ac{1,n}}(T)} = \frac{n}{4} \E\cro{N_{1+n}^{bad}(T)}.
\end{equation*}
Lemma~\ref{lem:invariance} also ensures that $\E\cro{N_{a}(T)\wedge M}=\E\cro{N_{1}(T)\wedge M}$ for all $a\in\ac{1,\ldots,n}$. 
By Lemma~\ref{lem:inter}, it follows that
\begin{align*}
16\E\cro{N^{bad}(T)} = 4 n \E\cro{N_{1+n}^{bad}(T)}
 &\geq  4 n \E\cro{N_{1+n}^{bad}(\tilde T)}\\
 &\geq n \E\cro{N_{1}(T)\wedge M}= \sum_{a=1}^n \E\cro{N_{a}(T)\wedge M}.
\end{align*}
Hence, by Lemma~\ref{lem:sum-M}
\begin{align*}
 16\E\cro{N^{bad}(T)}&\geq  \pa{(M\sqrt{T})\vee \frac{MT}{B_{T}}} \wedge \frac{T}{2}.
  \end{align*}
 For $\tilde s\leq 1/16$, taking $M$ equal to the integer part of $1/(16\tilde s)$ gives
\begin{equation*}
16\E\cro{N^{bad}(T)} \geq \pa{\frac{\sqrt{T}}{32\tilde s}\vee \frac{T}{32 \tilde s B_{T}}} \wedge \frac{T}{2}\, .
\end{equation*}
Since the sampling-regret does not depend on the choice of $\mu$, the proof is complete.

\section{Proof of the Unconstrained Upper Bound}
\label{sec:upper}
This section proves the following result, from which follows the upper bound of Theorem \ref{thm:non-contraint}, as explained below Theorem~\ref{thm:upper-appendix}.

\begin{thm}
\label{thm:upper-appendix}
There exist numerical constants $c_{1},c_{2}>0$, such that, for any $T\leq c_{2} n^2$, with probability at least $1-13/T$, the algorithm described in Section \ref{sec:algo} fulfils
\begin{equation*}
N^{bad}(\psi,T)\leq c_{1} \pa{T\wedge \frac{\sqrt{T}}{s}}.
\end{equation*}
\end{thm}

Let us explain how the upper bound of  Theorem \ref{thm:non-contraint} follows from Theorem~\ref{thm:upper-appendix}. 
First, let us note that the upper bound of Theorem~\ref{thm:upper-appendix} also holds in expectation.
Indeed, since $N^{bad}(\psi,T)\leq T$,  the algorithm described in Section \ref{sec:algo} fulfils
\begin{equation}\label{eq:bound:T:fixed}
\E\cro{N^{bad}(\psi,T)}\leq c_{1} \pa{T\wedge \frac{\sqrt{T}}{s}}+13\leq c'_{1} \pa{T\wedge \frac{\sqrt{T}}{s}}.
\end{equation}
Second, we can get an horizon free algorithm by applying a doubling trick.
For any integer $l$, let $t_l = 2^l$. 
At each time $t_l$, discard all nodes and pairs involved in the previous iterations of the algorithm and restart the algorithm described in Section \ref{sec:algo} with time horizon $t_{l+1}-t_{l}$. The resulting strategy does not depend on any time horizon.
Let us prove that this horizon-free algorithm  also has a $O\pa{T\wedge (\sqrt{T}/{s})}$ sampling regret. The argument for this proof is classical:
according to the upper bound (\ref{eq:bound:T:fixed}),  for any $t_{l-1}\leq T \leq t_{l}<c_2n^2$,
\begin{align*}
    \mathbb{E}\cro{N^{bad}(\psi,T)} &\leq c_{1}    \pa{             \frac{\sqrt{t_0}}{s}\wedge t_{0}+\frac{\sqrt{t_1-t_0}}{s}\wedge (t_{1}-t_{0})+...+\frac{\sqrt{t_l-t_{l-1}}}{s}\wedge (t_{l}-t_{l-1})}\\
    &\leq c_{1} \pa{\frac{1}{s}+\frac{1}{s} \sum_{r=0}^{l-1}2^{r/2}}\wedge t_{l}\\
    &\leq c_{1} \pa{\frac{\sqrt{t_{l}}}{(\sqrt{2} - 1) s}\wedge t_{l}}
    \leq 4c_{1} \pa{\frac{\sqrt{T}}{s} \wedge T}.
\end{align*}
Hence, we have proved that the upper bound of Theorem \ref{thm:non-contraint} is a consequence of Theorem~\ref{thm:upper-appendix}.
\smallskip

The proof of Theorem~\ref{thm:upper-appendix} is quite lengthy.
To help the reader to understand the organization of this demonstration, the section starts with a sketch of proof.

\subsection{Outline of the Proof of Theorem \ref{thm:upper-appendix}}
As any strategy has at most linear regret, it is sufficient to prove that there exist two positive numerical constants $c_{\thresh}$ and $c_{1}$ such that, for any  $T\geq c_{\thresh}/s^2$, the number $N^{bad}(\psi,T)$ of pairs sampled among $\cE^{\text{bad}}$ by the strategy $\psi$ described in the algorithm p.\pageref{alg:unconstrained} in Section~\ref{sec:algo}  is smaller than $c_{1} \sqrt{T}/s$ with probability at least $1-13/T$. 
As a consequence, in the proof, without loss of generality, it is assumed that $T\geq c_{\thresh}/s^2$, for a sufficiently large constant $c_{\thresh}$.
To prove the theorem, it is sufficient to show that neither Steps 1., 2. nor 3. of the algorithm sample more than $O(\sqrt{T}/s)$ ``bad" pairs, where a bad pair involves one node from community $1$ and one from community $2$.

\paragraph{Step 1.}
In the first step, the algorithm samples at random a core-set $\Ncal$ of $N=\lceil \sqrt{T}/\log(s\sqrt{T})\rceil$ nodes (point 1. in the algorithm). 
In this core-set, with large probability, at least $\lceil N/4 \rceil $ nodes from each community are sampled.
This result follows from Hoeffding's concentration inequality for hypergeometric random variables, it is rigorously established in point 2 of Lemma~\ref{th_step1}.

Each pair of the core-set is sampled with probability proportional to $\sqrt{T}/\big(s\binom{N}{2}\big)$ (point 2. of the algorithm).
With high probability, the set of sampled pairs $\Ocal_{0}$ has cardinality $|\Ocal_{0}|\asymp \sqrt{T}/s$, see point 3. of Lemma~\ref{th_step1}. 
At this point, the observed graph follows a cSBM with connection probabilities $\widetilde{p} \asymp p \sqrt{T}/\big(s\binom{N}{2}\big)$ and $\widetilde{q} \asymp q \sqrt{T}/\big(s\binom{N}{2}\big)$ . 
By \eqref{eq:GoodClust}, setting $\widetilde s= (\widetilde{p}-\widetilde{q})^2/(\widetilde{p}+\widetilde q)$ the proportion of misclassified nodes by {\tt GOODCLUST} is upper bounded by
\begin{equation*}
\exp(-c^\text{GC}_{1} N \widetilde s) = \exp\pa{- c \frac{\sqrt{T}/s}{N}s} = \exp(-\log(s\sqrt{T}))\leq \frac{1}{Ns},
\end{equation*}
with probability at least $1 - c^\text{GC}_{2}/N^3$. 
In particular, at most $1/s$ nodes of the core-set are misclassified.
A rigorous proof of this last statement is provided in point 4 of Lemma~\ref{th_step1}.

Let us comment briefly the choice of the cardinalities $N$ of the core-set and $|\Ocal_{0}|$ of the sampled pairs in this first step of the algorithm. These are chosen to guarantee the following properties.
\begin{description}
 \item[(1.i)] $N$ is sufficiently large to make the probability $c^\text{GC}/N^3$ small and, on the other hand, $N$ is sufficiently small so that one can classify a large proportion of $\Ncal$ with less than $|\Ocal_0|=O(\sqrt{T}/s)$ observed pairs.
 \item[(1.ii)] $|\Ocal_{0}|$ is large enough to ensure that the proportion of misclassified nodes in $\Ncal$ satisfies $\exp\big(- c^\text{GC}_{1} \E[|\Ocal_{0}|]s/N\big) \leq 1/ (Ns)$.
 \item[(1.iii)] On the other hand, $|\Ocal_{0}|$ is small enough, namely $|\Ocal_0|=O(\sqrt{T}/s)$, to ensure a regret $O(\sqrt{T}/s)$ in this exploratory phase of the algorithm.
\end{description}
 Before moving to the screening step 2 of the algorithm, the estimator 
 \[
 \hat{\tau} = \frac{1}{|\Ocal_{0}|} \sum_{\{x,x'\} \in \Ocal_{0}} A_{\{x,x'\}}
 \]
 of $\tau=(p+q)/2$  is shown to satisfy, with large probability,
\begin{equation*}
|\hat \tau-p|\wedge |\hat \tau-q| \geq |\hat \tau -\tau|.
\end{equation*}
This property is obtained by a careful application of Bernstein inequality for hypergeometric random variables in point 5 of Lemma~\ref{th_step1}.
This estimation of $\tau$ is sufficient for the screening step.
\medskip

\paragraph{Step 2.} The second step of the algorithm samples uniformly at random a set $\Acal_{0}$ of $\lceil 8\sqrt{2T} \rceil$ nodes. 
These nodes are screened with the following objectives.
\begin{description}
 \item[(2.i)] A set of at least $\lceil \sqrt{2T} \rceil$ nodes among $\Acal_{0}$ are selected containing at most $1/s$ members of community 2.
 \item[(2.ii)] A set of at most $O(\sqrt{T}/s)$ bad pairs is sampled during this screening.
\end{description}
Claims (2.i) and (2.ii) are formally established in Lemma~\ref{th_step2}, Claim (2.i) in points 8 and 10 and Claim (2.ii) at point 9.

The main tool for proving these two properties is Lemma~\ref{lem:bandit}.
It ensures that the probability that a node from community $2$ is not removed after $i$ steps of screening  decreases exponentially fast with $i$. 
Therefore, after $I\asymp \log(s\sqrt{T})$ screening steps, each node from community 2 remains with probability at most $e^{-c\log(s\sqrt{T})}$. 
Since there are $O(\sqrt{T})$ nodes in  $\Acal_{0}$, the expected number of remaining nodes from community $2$ is upper bounded, when $T\gtrsim 1/s^2$, by
\begin{equation*}
O\left( \sqrt{T} e^{-c\log(s\sqrt{T})} \right) \lesssim \frac{1}{s}.
\end{equation*}
The same bound holds with high probability. 
Similar arguments are used to obtain that, with large probability, less than $\lceil 8\sqrt{2T} \rceil - \lceil \sqrt{2T} \rceil$ nodes are removed during the screening step, which shows property (2.i). 

The proof of Property (2.ii) is more involved. 
At step 7(b) of the algorithm, a bad pair is sampled when it involves either 
\begin{description}
 \item[(2.ii.a)] a node of community 2 and a well classified node of the core-set,
 \item[(2.ii.b)] a node of community 1 and a misclassified node of the core-set.
\end{description}

The number of pairs in the case (2.ii.a) is simply bounded from above by $|\Acal_{0}|=O(\sqrt{T})$ multiplied by the number of misclassified nodes in the core-set. We have checked in step 1, that  the number of misclassified nodes in the core-set is bounded from above by $O(1/s)$.
So, on this event, the number of such bad pairs is at most $O(\sqrt{T}\times 1/s)$.

The number of pairs  in the case (2.ii.b) is bounded from above as follows. 
During each screening step (point 7.), a node is queried $k=O(1/s)$ times. 
Thus, the number of queries of a node from community 2 during this screening step is $k$ times the number of screening steps before it is removed. 
Recall that, from Lemma~\ref{lem:bandit}, the probability that a node of community 2 remains after $i$ screening steps decreases exponentially fast with $i$. 
Hence, the expected number of queries of a node from community $2$ is bounded from above by
\begin{equation*}
k \times \sum_{i\geq 1} e^{-ci}=O( k)=O(1/s)\,.
\end{equation*}
The number of sampled pairs in case (2.ii.b) is smaller than the total number of queries on nodes from community $2$ in $\Acal_0$, which is smaller than $O(|\Acal_0|k)=O(\sqrt{T}/s)$.
This bound also holds with high probability, which proves property (2.ii.b).

\medskip

\paragraph{Step 3.} During Step 3. of the algorithm, pairs within $\Acal_{I}$ are sampled until $T$ pairs have been sampled overall.
On the event where $|\Acal_{I}|$ is larger than $\sqrt{2T}$, this sampling is possible.
 In addition, on the event where the number of nodes from community 2 in $\Acal_{I}$ is upper bounded by $1/s$, the number of bad pairs in $\Acal_{I}$ is smaller than 
 \[
 O(|\Acal_I|/s)=O(|\Acal_0|/s)=O(\sqrt{T}/s).
 \]

\subsection{Proof of Theorem \ref{thm:upper-appendix}}
All we need is to prove that there exists a numerical constant $c_{\thresh}\geq 1$, such that, for any  $T\geq c_{\thresh}/s^2$, the upper bound $N^{bad}(\psi,T)\leq c_{1} \sqrt{T}/s$ holds with probability at least $1-13/T$. We  focus then on the case where $T\geq c_{\thresh}/s^2$.

Denote by $\hat G=\{\hat G_{1},\hat G_{2}\}$ the partition of $\Ncal$ output by the {\tt GOODCLUST} algorithm 
and by $S \Delta S'$ the symmetric difference between two sets $S,S'$. Define the community labelling  vectors $Z$ and $\hat Z$ by $Z_{x}=j$ for all $x\in G_{j}$ and $\hat Z_{x}=j$ for all $x\in \hat G_{j}$.
The following lemma controls the first step of the algorithm.

\begin{lem}
\label{th_step1}
There exists numerical constants  $c_\thresh \geq e$ and $T_0 \geq 1$ such that, if $T_0 \leq T \leq {n^2}/{16}$ and $s\sqrt{T} \geq c_\thresh$, then with probability at least $1 - 9/T$:
\begin{enumerate}
\item \label{0smallKernel} only a small part of the nodes has been sampled: $N \leq \frac{n}{4}$;
\item \label{5classRep} the two communities of the sampled nodes are approximately balanced, that is $N_1 \wedge N_2 \geq N / 4$, where $N_j := | \{Z = j\} \cap \Ncal |$ is the number of nodes from community $j$ in $\Ncal$;
\item \label{2limitEdge} the cardinality of the sample pairs fulfils $\frac{ c_{\Ocal_0}\sqrt{T} }{ 2s}\leq |\Ocal_0|\leq \frac{3\, c_{\Ocal_0}\sqrt{T} }{ 2s}$;
\item \label{3classif} the fraction of misclassified nodes is upper bounded by
\begin{equation*}
\varepsilon_N = \inf_{\pi\,\textrm{permutation on}\,\ac{1,2}} \frac{1}{2N}\sum_{k=1}^2 |\ac{Z=k}\Delta \{\hat Z=\pi(k)\}|\leq \frac{1}{s N};
\end{equation*}
\item \label{6estimaP} $|\hat{\tau} - \frac{p+q}{2}| \leq \frac{p-q}{4}$.
\end{enumerate}
\end{lem}
We refer to Section \ref{1proofEtape} for a proof of this lemma.

At the end of the first step, $|\Ocal_{0}| = O(\frac{\sqrt{T}}{s})$ pairs have been sampled according to point~\ref{2limitEdge} of Lemma~\ref{th_step1}, thus resulting in a number of sampled bad pairs $O(\frac{\sqrt{T}}{s})$.
Let us now turn to the second step of the algorithm. 

Assume without loss of generality that the community labelling $\hat Z$ of the nodes in $\Ncal$ is mostly in agreement with $Z$, i.e. the infimum in the definition of $\varepsilon_N$ is achieved for the identity permutation:
\begin{equation*}
\varepsilon_{N}=\frac{1}{2N}\sum_{k=1}^2 |\ac{Z=k}\Delta \{\hat Z=k\}|.
\end{equation*}
If it is not the case, the remaining of the proof still holds but with $\ac{Z=1}$ replaced by $\ac{Z=2}$.

For each $x\in\Acal_{0}$, the (distinct) nodes $\ac{y_{1}^x,\ldots,y_{kI}^x}$ are sampled uniformly at random in $\Ncal\cap\ac{\hat Z=1}$. Let $\Vcal_{x,0} = \emptyset$ for all $x \in \Acal_0$ and $\Vcal_{x,i}=\ac{y_{1}^x,\ldots,y_{ki}^x}$ for $i=1,\ldots,I$. Note that $| \Vcal_{x,j} | = kj$ for all $x \in \Acal_{0}$.
By induction, construct the sequences of sets $(\Acal_i)_{0 \leq i \leq I}$, which contain the ``active" nodes remaining at each iteration, and $(\Ocal_i)_{0 \leq i \leq I}$, which contain the sampled pairs.

More formally, 
for  $i \geq 1$ and  all $x \in \Acal_{i-1}$, the pairs $\{ \{x, y^x_{(i-1)k+a}\}, 1 \leq a \leq k \}$ are observed at iteration $i$, so that
\begin{equation*}
\Ocal_i = \Ocal_{i-1} \cup \bigcup_{x \in \Acal_{i-1}} \{ \{x, y^x_{(i-1)k+a}\}, 1 \leq a \leq k \}.
\end{equation*}

We remind the reader that we estimate the connectivity between $x$ and community 1 by
\begin{equation*}
\hat{p}_{x,i} = \frac{1}{ki} \sum_{y \in \Vcal_{x,i}} A_{x,y}
\end{equation*}
and only keep the nodes whose estimated connectivity is large enough in the active set:
\begin{equation}
\Acal_i = \left\{ x \in \Acal_{i-1} : \hat{p}_{x,i} \geq \hat{\tau} \right\}.
\end{equation}

After $I$ iterations, the total number of sampled pairs is
\begin{equation*}
| \Ocal_I | = |\Ocal_0| + k \sum_{i=0}^{I-1} | \Acal_i |
\end{equation*}
and the number of sampled bad pairs from this step is upper bounded by
\begin{equation}
\label{eq_regret_step2}
k \sum_{i=0}^{I-1} | \Acal_i \cap \{ Z \neq 1 \} | + |\Acal_0 \cap \{Z=1\}| \times |\Ncal \cap \{\hat{Z} \neq Z\}|
\end{equation}
where the first term comes from the pairs connecting community 2 to the core-set and the second term comes from the pairs connecting community 1 to a misclassified vertex of the core-set.

The following lemma controls this screening step.
\begin{lem}
\label{th_step2}
There exists numerical constants $T_0'$, $c_\thresh'$  larger than 1 such that if $T_0' \leq T \leq (\frac{3n}{64\sqrt{2}})^2$
and $s \sqrt{T} \geq c_\thresh'$, then with probability at least $1 - 13/T$, Lemma~\ref{th_step1} holds and
\begin{enumerate}
\setcounter{enumi}{5}
\item \label{7limitedRessource} the algorithm does not run out of connections with the core-set of the first step: $k I \leq N$;
\item \label{6enoughNewVertices} it is possible to take $|\Acal_0|$ new vertices: $|\Acal_0| \leq \frac{3n}{4} \leq n - N$;
\item \label{8item_plus_de_mauvais_a_la_fin} few vertices from the wrong community remain: $| \Acal_I \cap \{ Z \neq 1 \} | \leq \frac{1}{s}$;
\item \label{10item_regret_constant} the number of sampled bad pairs from nodes in the wrong community is controlled:

$k\sum_{i=0}^{I-1} | \Acal_i \cap \{ Z \neq 1 \} | \leq C_\text{fail} \frac{\sqrt{T}}{s}$ for a numerical constant $C_\text{fail}$;
\item \label{9item_il_reste_des_bons} enough vertices from community 1 remain for the next step: $| \Acal_I \cap \{ Z = 1 \} | \geq  \sqrt{2T} $.
\end{enumerate}
\end{lem}
We refer to  Section \ref{3proof} for a proof of this lemma.

Equation~\eqref{eq_regret_step2} together with point~\ref{10item_regret_constant} of Lemma~\ref{th_step2} and point~\ref{3classif} of Lemma~\ref{th_step1} entail that the number of sampled bad pairs during the screening step is again $O(\sqrt{T}/s)$.
 
Finally, during the last step, the algorithm uses the remaining budget to observe pairs uniformly at random between vertices of $\Acal_I$. Point~\ref{9item_il_reste_des_bons} of Theorem~\ref{th_step2} ensures that the number of possible pairs is larger than $T-\sqrt{T/2}$, which allows to spend the whole budget (since at least $\lceil \sqrt{T/2} \rceil$ pairs have been observed in the previous steps), and point \ref{8item_plus_de_mauvais_a_la_fin} ensures that the number of sampled bad pairs of this step is again $O(\frac{\sqrt{T}}{s})$.

Hence, the total number of bad pairs sampled during the whole process is $O(\sqrt{T}/s)$.

\subsection{Proofs of the Technical Lemmas}

\subsubsection{Proof of Lemma~\ref{th_step1}}
\label{1proofEtape}

The proof of point~\ref{0smallKernel} is straightforward: since $\sqrt{T} \leq n/4$ by assumption, the condition $N \leq n/4$ holds as soon as $N \leq \sqrt{T},$ that is $\lceil \frac{\sqrt{T}}{\log(s\sqrt{T})} \rceil \leq \sqrt{T}$ by definition of $N$. Therefore point~\ref{0smallKernel} holds true as soon as $s\sqrt{T} \geq c_{thresh}$ for some numerical constant $c_{thresh}$.

\paragraph{Proof of point~\ref{5classRep}.}

There are only two communities, so it is enough to consider the first one. Since the communities are balanced, the number $N_1$ of nodes from community 1 in the core-set follows an hypergeometric distribution with parameters $(N, 1/2, n)$. Therefore,
\begin{align*}
\P\left(\left|N_1 - \frac{N}{2}\right| \geq \sqrt{2 N \log N} \right)
	&\leq \frac{2}{N^4}
\end{align*}
using Equation~\eqref{eq_Hoeffding_hypergeom}. Since $N = \left\lceil \frac{\sqrt{T}}{\log (s\sqrt{T})} \right\rceil$,
\begin{align*}
\frac{2}{N^4}
	\leq \frac{2 \log(s\sqrt{T})^4}{  T^2}
	\leq \frac{(\log T)^4}{8 T^2}
\end{align*}
using $s \leq 1$, which is upper bounded by $1/T$ for all $T\geq 1$. Assuming $s\sqrt{T} \geq c_{thresh}$ for some numerical constant $c_{thresh}\geq e$, one has $\frac{\sqrt{T}}{\log \sqrt{T}} \leq N \leq  \sqrt{T}$, so that
\begin{align*}
\frac{\sqrt{2 N \log N}}{N/4} 
	&\leq 4\sqrt{2} \sqrt{\frac{\log ( \sqrt{T})}{ \sqrt{T} / \log \sqrt{T}}} \\
	&\leq 4\sqrt{2} \sqrt{\frac{\log ( \sqrt{T})^2}{ \sqrt{T}}}.
\end{align*}
 Therefore, it is smaller than $1$ as soon as  $T \geq T_{0,2}$ for some numerical constant $T_{0,2}$, which entails
\begin{align*}
\P\left(\left|N_1 - \frac{N}{2}\right| \geq \frac{N}{4}\right)
	&\leq \frac{1}{T},
\end{align*}
and the same for $N_2$.

\paragraph{Proof of point~\ref{2limitEdge}.} The number $|\Ocal_0|$ of sampled pairs in the core-set $\Ncal$ follows a binomial distribution with parameters $\left(\binom{N}{2}, c_{\Ocal_{0}}  \sqrt{T}/s\binom{N}{2}\right)$. Therefore,

\begin{equation*}
\P \left( \left| |\Ocal_0|-  c_{\Ocal_{0}} \frac{\sqrt{T}}{s} \right| \geq \sqrt{2 c_{\Ocal_{0}} \frac{\sqrt{T}}{s} \log(2T)} + \log(2T) \right) \leq \frac{1}{T}
\end{equation*}
using Bernstein's inequality~\eqref{eq_Bernstein_sym}. This implies that \begin{equation}\label{proof_point3_arete_aleat}\frac{1}{2}  c_{\Ocal_{0}} \frac{\sqrt{T}}{s} \leq |\Ocal_0|\leq \frac{3}{2}  c_{\Ocal_{0}} \frac{\sqrt{T}}{s}
\end{equation}as soon as $T \geq T_{0,3}$ for some numerical constant $T_{0,3}$.

Let us check that the probability parameter of the binomial distribution is well defined, that is, the condition $c_{\Ocal_{0}}  \sqrt{T}/s\binom{N}{2} \in [0,1]$ is satisfied. One can show that $N \geq 8$ as soon as $T\geq T_{0,3}$ for some numerical constant $T_{0,3}$. Then
\begin{equation*}
\binom{N}{2} \geq \frac{N^2}{4}
\end{equation*}
so that the condition holds as soon as $c_{\Ocal_{0}}  \sqrt{T}/s \leq N^2/4,$ which is implied by
\begin{equation*}
c_{\Ocal_{0}}  \sqrt{T}/s \leq \frac{1}{4} \frac{T}{(\log(s\sqrt{T}))^2},
\end{equation*}
or equivalently
\begin{equation*}
\frac{s\sqrt{T}}{(\log(s\sqrt{T}))^2} \geq 4 c_{\Ocal_{0}}.
\end{equation*}
Hence, the condition $c_{\Ocal_{0}}  \sqrt{T}/s\binom{N}{2} \in [0,1]$ holds as soon as  $s\sqrt{T} \geq c_{thresh}$ for some numerical $c_{thresh}$.
This, together with \eqref{proof_point3_arete_aleat}, concludes the proof of point~\ref{2limitEdge}.

\paragraph{Proof of point \ref{3classif}.}
Since each pair of $\Ncal$ is sampled with probability $c_{\Ocal_{0}}  \sqrt{T}/s\binom{N}{2}$, the matrix $\widetilde{A}$ defined by $\tilde{A}_{x,x'} = A_{x,x'}$ if the pair $\{x,x'\}$ has been sampled and zero otherwise has the same distribution as the adjacency matrix of a fully observed SBM with connection probabilities $\widetilde{p} = p\,c_{\Ocal_{0}}  \sqrt{T}/s\binom{N}{2}$ and $\widetilde{q} = q\,c_{\Ocal_{0}}  \sqrt{T}/s\binom{N}{2}$. Therefore, the proportion $\varepsilon_N$ of misclassified nodes in $\Ncal$ by the {\tt GOODCLUST} algorithm is upper bounded by
\begin{equation}
\varepsilon_N \leq \exp \left( - c^\text{GC}_{1} N  \frac{(\widetilde{p}-\widetilde{q})^2}{\widetilde{p}+\widetilde q}\right)
\end{equation}
with probability at least $1 - c^\text{GC}/N^3$.
Hence with probability at least $1 - 1/T$ 
\begin{align*}
\varepsilon_N
	&\leq \exp \left( - 2\, c^\text{GC}_1 c_{\Ocal_0} \frac{\log(s \sqrt{T})}{2} \right) 
\end{align*}using $N := \lceil \frac{\sqrt{T}}{\log(s\sqrt{T})} \rceil \leq 2 \frac{\sqrt{T}}{\log(s\sqrt{T})}$ as soon as $T \geq T_{0,4}$ for some numerical constant $T_{0,4}$. 
Hence, by taking $c_{\Ocal_0} \geq 1/(c^\text{GC}_1)$, one has with probability at least $1 - 1/T$
\begin{align*}
\varepsilon_N
	\leq \exp \left( - \log(s \sqrt{T}) \right)
	= \frac{1}{s\sqrt{T}}
\end{align*}
so that
\begin{equation*}
\varepsilon_N \leq  \frac{1}{s N}
\end{equation*}
as soon as $N \leq \sqrt{T}$, which holds true when $s\sqrt{T} \geq c_{thresh}$ for some numerical constant $c_{thresh}.$

\paragraph{Proof of point \ref{6estimaP}.} Let $\Ocal_\within := \Ocal_0 \cap \Ecal^{good}$
be the subset of within-group  pairs, and $\Ocal_\out := \Ocal_0 \setminus \Ocal_\within$ the subset of pairs between two different communities. Then
\begin{equation*}
\hat{\tau} = \frac{|\Ocal_\within|}{|\Ocal_0|} \frac{1}{|\Ocal_\within|} \sum_{(x,x') \in \Ocal_\within} A_{x,x'}
		+ \frac{|\Ocal_\out|}{|\Ocal_0|} \frac{1}{|\Ocal_\out|} \sum_{(x,x') \in \Ocal_\out} A_{x,x'}.
\end{equation*}

Conditionally to the number of sampled pairs $|\Ocal_0|$ and the number of within-group pairs $|\Ocal_\within|$, the sum $\sum_{(x,x') \in \Ocal_\within} A_{x,x'}$ (resp. $\sum_{(x,x') \in \Ocal_\out} A_{x,x'}$) is independent of $\Ocal_0$, and is a sum of i.i.d. Bernoulli random variables with parameter $p$ (resp. $q$). Therefore, Bernstein's inequality (\ref{eq_Bernstein}) ensures that with probability at least $1-4/T$
\begin{align*}
\frac{|\Ocal_\within|}{|\Ocal_0|} \left| \frac{1}{|\Ocal_\within|} \sum_{(x,x') \in \Ocal_\within} A_{x,x'} - p \right|
	&\leq \sqrt{2p\frac{\log T}{|\Ocal_0|}} + \frac{\log T}{|\Ocal_0|} \\
\text{and} \qquad
\frac{|\Ocal_\out|}{|\Ocal_0|} \left| \frac{1}{|\Ocal_\out|} \sum_{(x,x') \in \Ocal_\out} A_{x,x'} - q \right|
	&\leq \sqrt{2q\frac{\log T}{|\Ocal_0|}} + \frac{\log T}{|\Ocal_0|}.
\end{align*}
Using point~\ref{2limitEdge}, one has $|\Ocal_0|\geq c_{\Ocal_0}\sqrt{T}/(2s)$ with probability at least $1 - 1/T$, so that
\begin{align*}
\left| \hat{\tau} - \left(\frac{|\Ocal_\within|}{|\Ocal_0|} p + \frac{|\Ocal_\out|}{|\Ocal_0|} q \right) \right|
	&\leq 2\sqrt{2p s \frac{\log T}{c_{\Ocal_0}   \sqrt{T}/2}} + 2 s \frac{\log T}{c_{\Ocal_0}   \sqrt{T}/2} \\
	&\leq 2(p-q)\sqrt{2 \frac{\log T}{c_{\Ocal_0}   \sqrt{T}/2}} + 2 (p-q) \frac{\log T}{c_{\Ocal_0}   \sqrt{T}/2}
\end{align*}
 with probability at least $1 - 5/T$, using $s = (p-q)^2/p \leq p-q.$ Finally, since $c_{\Ocal_0}/2 \geq 1$ and $|\Ocal_\within| = |\Ocal_0| - |\Ocal_\out|$,
\begin{align}
\nonumber \left| \hat{\tau} - \frac{p+q}{2} \right|
	&\leq \left|\frac{|\Ocal_\within|}{|\Ocal_0|} p + \frac{|\Ocal_\out|}{|\Ocal_0|} q -\frac{p+q}{2} \right|
		+ 2(p-q)\sqrt{2\frac{\log T}{\sqrt{T}}} + 2(p-q)\frac{\log T}{\sqrt{T}} \\
\label{eq_controle_erreur_p}
	&\leq \left|\left(2\frac{|\Ocal_\out|}{|\Ocal_0|}-1\right)\frac{p-q}{2} \right| + \frac{|p-q|}{16}
\end{align}
as soon as $T \geq T_{0,4}$ for some numerical constant $T_{0,4}$.

Conditionally to the number of pairs $|\Ocal_0|$ and the sizes $N_{1}$ and $N_{2}$ of the two communities sampled in $\Ncal$, the number $|\Ocal_\out|$  of between group pairs follows an hypergeometric distribution with parameters $\left(|\Ocal_0|, r, \binom{N}{2}\right)$ with $r = {N_1 N_2}/{\binom{N}{2}}$. Conditionally to $|\Ocal_{0}|$ and the event $ \frac{3}{8} \leq  r \leq \frac{5}{8}$, the random variable $|\Ocal_\out|$ dominates stochastically an  hypergeometric random variable with parameters $(|\Ocal_0|, \frac{3}{8}, \binom{N}{2})$ and it is stochastically dominated by an  hypergeometric random variable with parameters $(|\Ocal_0|, \frac{5}{8}, \binom{N}{2})$.  There exists a real $\gamma > 0$ such that $N_1 = \gamma N$ and $N_2 = (1-\gamma) N$ so that
\begin{equation*}
r = \frac{\gamma N (1-\gamma) N}{N (N-1)/2} =2 \gamma (1-\gamma)(1 + \frac{1}{N-1}) = 2 \gamma (1-\gamma)(1 + \frac{1}{\frac{\sqrt{T}}{\log(s\sqrt{T})} - 1})
\end{equation*}
Using point~\ref{5classRep}, one has with probability at least $1-1/T$ that $\frac{1}{4}\leq \gamma \leq \frac{3}{4}$ which entails $ \frac{3}{8} \leq  r \leq \frac{5}{8}$ as soon as $T \geq T_{0,5}$ for some numerical constant $T_{0,5}$. Therefore,
\begin{align*}
\P \left(
		|\Ocal_\out| \leq \frac{3 |\Ocal_0|}{8} - \sqrt{\frac{|\Ocal_0| \log T}{2}}
	\right)
&\leq \frac{1}{T} + \frac{1}{T}
\end{align*}
using Equation~\eqref{eq_Hoeffding_hypergeom}, and similarly
\begin{align*}
\P \left(
		|\Ocal_\out| \geq \frac{5|\Ocal_0|}{8} + \sqrt{\frac{|\Ocal_0| \log T}{2}}
	\right)
&\leq \frac{1}{T} + \frac{1}{T}.
\end{align*}Using point~\ref{2limitEdge}, one has with probability at least $1-1/T$ that $|\Ocal_0|\geq c_{\Ocal_0}\sqrt{T}/(2s)$ which entails $\sqrt{\frac{|\Ocal_0| \log T}{2}} \leq \frac{|\Ocal_0|}{16}$ as soon as $T \geq T_{0,6}$ for some numerical constant $T_{0,6}$. Hence
\begin{equation*}
\frac{5}{16} \leq \frac{|\Ocal_\out|}{|\Ocal_0|} \leq \frac{11}{16}
\end{equation*}
with probability $1 - \frac{5}{T}$.
This, together with Equation~\eqref{eq_controle_erreur_p}, concludes the proof of point~\ref{6estimaP} (which holds with probability $1 - \frac{8}{T}$).

\subsubsection{Proof of Lemma~\ref{th_step2}}
\label{3proof}
\paragraph{Proof of point~\ref{7limitedRessource} and point~\ref{6enoughNewVertices}.}

There exists a constant $c'_\thresh$ such that $4C_{I}C_{k}\leq  \frac{s\sqrt{T}}{(\log(s\sqrt{T}))^2}$ as soon as $s\sqrt{T} \geq c'_\thresh$. It follows that $kI \leq N$.

Point~\ref{6enoughNewVertices} follows from straightforward algebra.

\paragraph{Proof of point~\ref{8item_plus_de_mauvais_a_la_fin}.}
For all $x \in \Acal_0$, denote by $T_x = \max \{ i : x \in \Acal_i\}$ the index of the last iteration where the vertex $x$ was in the active set. Let us first show that if $x$ is not in the first community, then $T_x$ has sub-exponential tails.

\begin{lem}\label{lem:bandit}
Set $\rho'=1/2000$.
 If $C_k \geq (\log 3) / \rho'$ then
\begin{equation}
\label{eq_queueExpo}
\forall i \in \Nbb^* \qquad \P(T_x \geq i) \leq e^{- \rho' C_k i}.
\end{equation}
\end{lem}

We refer to Section~\ref{proof:lem:bandit} for a proof of this lemma.

Let us now prove point~\ref{8item_plus_de_mauvais_a_la_fin}.
Let $T^{(1)}=|\Ocal_0|$ and $V_x = \one_{T_x \geq I}$. Conditionally on $\Fcal_{T^{(1)}}$, the variables $(V_x)_{x \in \Acal_0 \cap \{Z \neq 1\}}$ are i.i.d. Bernoulli random variables with parameter $r \leq e^{-\rho' C_k I}$ by equation~\eqref{eq_queueExpo}. Therefore, for all $i \in \Nbb$,
\begin{equation*}
\P\left( \sum_{x \in \Acal_0 \cap \{Z \neq 1\}} V_x = i \right) \leq \frac{|\Acal_0|^i}{i!} r^i
\end{equation*}
so that
\begin{align*}
\P( |\Acal_I \cap \{Z \neq 1\}| \geq i )
	&\leq \sum_{j \geq i} \frac{|\Acal_0|^j}{j!} r^j \\
	&\leq \frac{(|\Acal_0|r)^i}{i!} \sum_{j \geq 0} \left(\frac{|\Acal_0|r}{i}\right)^j \\
	&\leq 2 \frac{(|\Acal_0|r)^i}{i!}
\end{align*}
as soon as $i \geq 2|\Acal_0|r$. 
For $i=\lceil 1/s \rceil $, this condition holds if $16\sqrt{2} \leq (s\sqrt{T})^{\rho' C_{k}C_{I}-1}$ which holds when
$C_I C_{k} \geq 4 /  \rho'$ and $s\sqrt{T}\geq c'_{th}$.

Taking $i = \lceil 1/s \rceil$ and using that $i! \geq (i/e)^i$ for all $i \geq 1$, it follows that
\begin{align*}
\P\left( |\Acal_I \cap \{Z \neq 1\}| \geq \frac{1}{s} \right)
	&\leq 2 \left(\frac{e|\Acal_0|r}{\lceil 1/s \rceil}\right)^{\lceil 1/s \rceil} \leq  2 \left(se|\Acal_0|r\right)^{ 1/s }
\end{align*}as soon as $se|\Acal_0|r \leq 1.$

We want to take $r$ small enough such that $2 (se|\Acal_0|r)^{1/s} \leq 1/T$, that is
\begin{align*}
\log (se|\Acal_0|) + s\log (2T) \leq (-\log r),
\end{align*}
which holds as soon as
\begin{align*}
\rho' C_k I \geq \log (s \sqrt{T}) + \log (32e\sqrt{2}) + s \log T.
\end{align*}using $|\Acal_0| \leq 16 \sqrt{2T}$.

Note that $\frac{s \log T}{\log (s\sqrt{T})} = 2 \frac{s \sqrt{T}}{\log (s \sqrt{T})} \frac{\log \sqrt{T}}{\sqrt{T}}\leq 2$, since $\log(x)/x$ is decreasing for $x>e$  and $s\sqrt{T} \leq \sqrt{T}$, so that there exists a numerical constant $c_\thresh'$ such that if $s\sqrt{T} \geq c_\thresh'$, then point~\ref{8item_plus_de_mauvais_a_la_fin} is implied by
\begin{align*}
\rho' C_k I \geq 4 \log (s \sqrt{T}),
\end{align*}
which holds when $C_I C_{k} \geq 4 /  \rho'$.

\paragraph{Proof of point~\ref{10item_regret_constant}.}

Note that
\begin{align*}
k \sum_{i=0}^{I-1} | \Acal_i \cap \{Z \neq 1\} |
	&= k \sum_{x \in \Acal_0 \cap \{Z \neq 1\}} T_x.
\end{align*}

Conditionally on $\Acal_{0}$, the random variables $(T_x)_{x \in \Acal_0 \cap \{Z \neq 1\}}$ are i.i.d. random variables which are stochastically dominated by random variables $Y_x \sim \Ecal(\rho' C_k)$ by Equation~\eqref{eq_queueExpo}. These exponential random variables satisfy
\begin{equation*}
\E \left(Y_x - \frac{1}{\rho' C_k} \right)^2 \leq \frac{1}{(\rho' C_k)^2}
\end{equation*}
and for all $a \in \Nbb$ such that $a \geq 3$
\begin{equation*}
\E \left(Y_x - \frac{1}{\rho' C_k} \right)_+^a \leq \frac{a!}{(\rho' C_k)^a},
\end{equation*}
so that Bernstein's inequality, see for instance Proposition 2.9 of \citep{Mas07}, entails that for all $t > 0$
\begin{equation*}
\P\left( \sum_{x \in \Acal_0} Y_x - \frac{|\Acal_0|}{\rho' C_k}
	\geq \frac{2 \sqrt{|\Acal_0| t}}{\rho' C_k}
		+ \frac{t}{\rho' C_k} \right) \leq e^{-t}
\end{equation*}
and therefore by taking $t = \log T$, with probability at least $1 - 1/T$:
\begin{align*}
\sum_{x \in \Acal_0 \cap \{Z \neq 1\}} T_x
	&\leq \frac{16 \sqrt{2T}}{\rho' C_k}
		+ \frac{2 \sqrt{16 \sqrt{2T} \log T}}{\rho' C_k}
		+ \frac{\log T}{\rho' C_k} \\
	&\leq \frac{32 \sqrt{T}}{\rho' C_k}
\end{align*}
as soon as $T \geq T_0'$ for some numerical constant $T_0'$. Hence, with probability at least $1 - 1/T$,
\begin{align*}
k \sum_{i=0}^{I-1} | \Acal_i \cap \{Z \neq 1\} |
	& \leq \frac{64 \sqrt{T}}{\rho' s}
\end{align*}using $k \leq 2 C_K/s.$

\paragraph{Proof of point~\ref{9item_il_reste_des_bons}.}

The same proof as the one of Equation~\eqref{eq_controle_pxi} shows that for all $x \in \Acal_0 \cap \{Z=1\}$, for all $i \geq 1$ and for all $t > 0$,
\begin{equation}
\P\left(\hat{p}_{x,i} <
		p
		- \frac{|p-q|}{8}
		- |p-q| \sqrt{\frac{t}{2ki}}
		- 2\sqrt{2p\frac{t}{ki}}
		- 2\frac{t}{ki}
	\right) \leq 3e^{-t},
\end{equation}
so that by union bound and the inequality $k\geq C_{k}$,
\begin{align*}
\P\left(\exists i \geq 1, \quad \hat{p}_{x,i} <
		\frac{7p+q}{8} - |p-q|\left[ 
		 \sqrt{\frac{\log(2 \pi^2 i^2)}{C_k i}} \left( \frac{1}{\sqrt{2}} + 2\sqrt{2} \right)
		+ 2 \frac{\log(2 \pi^2 i^2)}{C_k i} \right]
	\right) \leq \frac{1}{4}.
\end{align*}

Therefore, if $C_k$ is larger than a numerical constant,
\begin{align*}
\P\left(\exists i \geq 1, \quad \hat{p}_{x,i} <
		\frac{3p+q}{4}
	\right) \leq \frac{1}{4},
\end{align*}
which, combined with point~\ref{6estimaP} of Theorem~\ref{th_step1}, implies
\begin{align*}
\P\left(\exists i \geq 1, \quad \hat{p}_{x,i} <
		\hat{\tau}
	\right) \leq \frac{1}{4}.
\end{align*}

Let $V_x = \one_{x \in \Acal_I}$ for all $x \in \Acal_0 \cap \{Z=1\}$. The above inequality ensures that conditionally on $\Acal_{0}$, the $(V_x)_{x \in \Acal_0 \cap \{Z=1\}}$ are i.i.d. Bernoulli random variable with parameter $r \geq 3/4$. Therefore, Hoeffding's inequality entails
\begin{equation*}
\P\left( |\Acal_I \cap \{Z=1\}| \leq \frac{3 |\Acal_0 \cap \{Z=1\}|}{4} - \sqrt{|\Acal_0| \frac{\log T}{2}} \right) \leq \frac{1}{T}.
\end{equation*}

Let us assume for now that $|\Acal_0 \cap \{Z=1\}| \geq 2\sqrt{2T}$ with probability $1 - 1/T$. Then this ensures that for $T$ larger than some numerical constant, 
\begin{equation*}
\P\left( |\Acal_I \cap \{Z=1\}| \leq \sqrt{2T} \right) \leq \frac{1}{T} + \frac{1}{T}.
\end{equation*}

To conclude, note that conditionally on $\Ncal$, the random variable $|\Acal_0 \cap \{Z=1\}|$ is an hypergeometric random variable with parameters $(\lceil 8\sqrt{2T} \rceil, r', n-N)$ where 
\begin{align*}
r' = \frac{\frac{n}{2} - |\Ncal \cap \{Z=1\}|}{n-N}
	\geq \frac{\frac{n}{2} - \frac{3}{4}\frac{n}{4}}{n}
	\geq \frac{5}{16}
\end{align*}
by points~\ref{0smallKernel} and \ref{5classRep} of Theorem~\ref{th_step1}. Therefore, Equation~\eqref{eq_Hoeffding_hypergeom} implies that
\begin{equation*}
\P\left( |\Acal_0 \cap \{Z=1\}| \leq \frac{5}{16} 8\sqrt{2T} - \sqrt{\frac{16\sqrt{2T} \log T}{2}} \right) \leq \frac{1}{T},
\end{equation*}
so that for $T$ larger than a numerical constant
\begin{equation*}
\P\left( |\Acal_0 \cap \{Z=1\}| \leq  2\sqrt{2T} \right) \leq \frac{1}{T}.
\end{equation*}

\subsection{Proof of Lemma~\ref{lem:bandit}}\label{proof:lem:bandit}

Let $x \in \Acal_{0} \cap \{ Z \neq 1\}$ and assume that we are in the event of probability at least $1-9/T$ where Theorem~\ref{th_step1} holds. For all $i \in \Nbb^*$,
\begin{align*}
\P(T_x \geq i)
	&= \P\left(\forall j \in \{1, \dots, i\}, \hat{p}_{x,j} \geq \hat{\tau} \right) \\
	&\leq \P\left(\hat{p}_{x,i} \geq \hat{\tau} \right) \\
	&\leq \P\left(\hat{p}_{x,i} \geq \frac{p + 3 q}{4} \right)
\end{align*}
using point~\ref{6estimaP} of Lemma~\ref{th_step1}.

Following the same proof as in point~\ref{6estimaP} of Theorem~\ref{th_step1}, one can show that for all $x\in \Acal_{0}\cap \ac{Z\neq 1}$, all $i \geq 1$ and all $t > 0$,
\begin{align*}
\P\left(\hat{p}_{x,i} \geq
		q
		+ \frac{|\Vcal_{x,i}^-|}{|\Vcal_{x,i}|} |p-q|
		+ 2\sqrt{2p\frac{t}{ki}}
		+ 2\frac{t}{ki}
	\right) \leq 2e^{-t}
\end{align*}
where $\Vcal_{x,i}^- := \Vcal_{x,i} \cap \{Z \neq 1\}$.

For $s\sqrt{T} \geq c_\thresh'$, with  $c_\thresh'$ such that $\frac{\log (s\sqrt{T})}{s\sqrt{T}}\leq 1/64$, one has $\frac{1}{s} \leq N/64$.
Then, points~\ref{5classRep} and~\ref{3classif} of Lemma~\ref{th_step1} imply that $|\Ncal \cap \{\hat{Z}=1\} \cap \{Z \neq 1\}| \leq N/64$ and $\hat{N}_1 := |\Ncal \cap \{\hat{Z}=1\}| \geq N/8$. Therefore, the proportion of misclassified vertices in $\Ncal \cap \{\hat{Z}=1\}$ is at most $1/8$, so that
conditionally on $\hat N_{1}$ and the event of Lemma~\ref{th_step1}
 $|\Vcal_{x,i}^-|$ is stochastically dominated by an hypergeometric distribution with parameters $(ki,1/8,\hat{N}_1)$. Hence, Equation~\eqref{eq_Hoeffding_hypergeom} entails
\begin{align*}
\P\left(
		\frac{|\Vcal_{x,i}^-|}{|\Vcal_{x,i}|} \geq \frac{1}{8} + \sqrt{\frac{t}{2ki}}
	\right) \leq e^{-t},
\end{align*}
so that for all $i \geq 1$ and $t > 0$,
\begin{equation}
\label{eq_controle_pxi}
\P\left(\hat{p}_{x,i} \geq
		q
		+ \frac{|p-q|}{8} + |p-q| \sqrt{\frac{t}{2ki}}
		+ 2\sqrt{2p\frac{t}{ki}}
		+ 2\frac{t}{ki}
	\right) \leq 3e^{-t}.
\end{equation}

Note that
\begin{multline*}
\frac{p+3q}{4} - \left( q
		+ \frac{|p-q|}{8} + |p-q| \sqrt{\frac{t}{2ki}}
		+ 2\sqrt{2p\frac{t}{ki}}
		+ 2\frac{t}{ki} \right) \\
	\geq  \frac{|p-q|}{8}
			- \sqrt{\frac{t}{C_k i}} \left( \frac{|p-q|\sqrt{s}}{\sqrt{2}} + 2\sqrt{2p s} \right) 
			- 2\frac{ts}{C_k i}\\ 
	\geq  |p-q| \left( \frac{1}{8}
			- \sqrt{\frac{t}{C_k i}} \left( \frac{1}{\sqrt{2}} + 2\sqrt{2} \right) 
			- 2\frac{t}{C_k i}
		\right).
\end{multline*}
since $s = (p-q)^2/p\leq 1$.

Thus, there exists a numerical constant $\rho=10^{-3}$ such that by taking $t = \rho C_k i$,
\begin{align*}
\frac{p+3q}{4} \geq q
		+ \frac{|p-q|}{8} + |p-q| \sqrt{\frac{t}{2ki}}
		+ 2\sqrt{2p\frac{t}{ki}}
		+ 2\frac{t}{ki},
\end{align*}
so that
\begin{align*}
\P\left(\hat{p}_{x,i} \geq \frac{p+3q}{4}
	\right) \leq 3e^{-\rho C_k i}
\end{align*}
and finally by letting $\rho'=\rho/2$ and if $C_k \geq (\log 3) / \rho'$:
\begin{equation*}
\forall i \in \Nbb^* \qquad \P(T_x \geq i) \leq e^{- \rho' C_k i}.
\end{equation*}


\section{Proof of the Constrained Upper Bound}
\label{section::proof:constrainedTHM}
This section proves the upper bound in Theorem~\ref{thm:contraint}.
Recall that $B=(B_{T}\wedge \sqrt{T})/2$ in the Constrained Algorithm page \pageref{alg:constrained}.

It is enough to prove the upper bound in Theorem~\ref{thm:contraint} in the case where $sB \geq c_{thresh}$ for some numerical constant $c_{thresh} \geq 1$. 
Indeed, if $sB \leq c_{thresh}$, Equation~\eqref{eq_upperbound_constrained} automatically holds with $c_{2} \geq c_{thresh}$. 
Hereafter, it is then assumed that $sB \geq c_{thresh}$.

The first step of the  Constrained Algorithm page \pageref{alg:constrained} is almost identical to that of the Unconstrained Algorithm after replacing $\sqrt{T}$ by $B = (B_T \wedge \sqrt{T})/2$ in the cardinality of the core-set. 
The following lemma is a slight variant of Lemma~\ref{th_step1} in this setting. The proof is omitted.

\begin{lem}
\label{th_step1_constrain}
There exist numerical constants $c_{thresh} \geq e$ and $B_0 \geq 1$, such that, if $B \geq B_0$ and $sB \geq c_{thresh}$ and $T/B \leq n/136$, then with probability at least $1 - 9 / (sB)$: 
\begin{enumerate}
\item \label{item_constrained_C1_Ninit} the number $N_{init}$ of sampled nodes satisfies $N_{init} \leq \frac{n}{8}-4 \sum_{t=1}^{t_f-1} N^{(t)}$,
\item at least $N_{init}/4$ nodes of each community have been sampled, that is $| \{Z = j\} \cap \Ncal_{init} | \geq N_{init} / 4$ for each $j \in \{1,2\}$,
\item  the proportion $\varepsilon_{N_{init}}$ of misclassified nodes satisfies
\begin{equation}
\varepsilon_{N_{init}}=\inf_{\pi\,\textrm{permutation on}\,\ac{1,2}} \frac{1}{2 N_{init}}\sum_{k=1}^2 |\ac{Z=k}\Delta \{\hat Z=\pi(k)\}|\leq \frac{4}{512^2}\, \frac{1}{s B},
\end{equation}

\item $|\hat{\tau} - \frac{p+q}{2}| \leq \frac{p-q}{4}$.

\end{enumerate}
\end{lem}

At the end of the first step, $|\Ocal_0|$ pairs have been sampled and the sampling-regret therefore does not exceed $\mathbb{E}\left[|\Ocal_0|\right] = c_{\Ocal_0} B/s \leq c_{\Ocal_0} T/(sB)$ since, by definition of $B$, $T \geq B^2$. 

Let us proceed with the second step.
To show that the sampling regret in the second step does not exceed $O(T/(sB))$, it is sufficient to prove that there exist two numerical constants $c_{proba}$ and $c_{regret}$ such that for any $(T,B)$ satisfying $sB \geq c_{thresh}$ and $T/B \leq n/136$, the number of bad pairs sampled during the second step is bounded from above by $c_{regret} T/(s B)$ with probability at least $1-c_{proba}/(sB)$. 
Indeed, since the number of bad pairs sampled in the second step  $N^{bad}_{step 2}(\psi,T)$ cannot be larger than $T$,
  it directly follows that the sampling-regret during the second step is upper bounded by
  \begin{equation*}
    \mathbb{E}\left[N^{bad}_{step 2}(\psi,T)\right] \leq c_{regret} \frac{T}{sB_T} + T \frac{c_{proba}}{sB_T} \leq c' \frac{T}{sB_T}.
\end{equation*}

The following lemma provides such a control of the number of bad pairs accumulated in step 2, as well as an upper bound on the number of misclassified nodes.
It is a counterpart to Lemma~\ref{th_step2}  of the unconstrained case. 
\begin{lem}
\label{th_step2_constrained}
There exist two numerical constants $B'_0 \geq 1$ and $c'_{thresh}\geq e$ such that if $B \geq B'_0$, $sB \geq c'_{thresh}$ and $T/B \leq n/136$, then with probability at least $1 -63 / (sB)$, Lemma~\ref{th_step1_constrain} holds and for all iterations of \texttt{SCREENING} in point~\ref{point_screening_algcontraint} of the constrained algorithm,
\begin{enumerate}
\setcounter{enumi}{4}

\item \label{item_constrained_always_vertices} it is always possible to sample $|\Acal_0|$ new vertices: $|V^{(0)}| \geq \ldots \geq |V^{(t_f-1)}| \geq \frac{7n}{8}$;

\item \label{item_contrainte_allIterations} No node from $\Acal_0$ has more than $2B$ adjacent pairs sampled during the whole execution of the constrained algorithm.

\item \label{item_lem1.3_initial_kernel_large_enough} the algorithm does not run out of connections with the reference core-set: \\ $kI \leq N^{(0)} \leq \ldots \leq N^{(t_f-1)}$;

\end{enumerate}

\noindent
and there exists a numerical constant $\cfail$ such that for all $t \in \ac{1, \dots, t_f}$, during the call $\texttt{SCREENING}\pa{\Ncal^{(t-1)}, N^{(t)}, B, \hat{\tau}, V^{(t-1)}}$,
\begin{enumerate}[resume]
\item \label{item_regret_constrained} the number of bad pairs sampled during the $t^{th}$-call to \texttt{SCREENING} from nodes in $\Acal_0$ is controlled: \\
$
\sum_{x \in \Acal_0} \left| \left\{ y^x_a : (x,y^x_a) \text{ sampled and } Z_{y^x_a} \neq  Z_x \right\} \right|
	\leq \cfail \frac{N^{(t)}}{s}$;
	
\item \label{item_misclass_constrained} few vertices from the wrong community remain: $| \Ncal^{(t)} \cap \{ Z \neq 1 \} | \leq 8 N^{(t)}/(sB)$;
	
\item \label{item_enoughVertices_constrained} enough vertices from community 1 remain for the construction of the core-set $\Ncal^{(t)}$ of $N^{(t)}$ nodes: $\sum_{j=1}^m | \Acal_I^{(j)} \cap \{ Z = 1 \} | \geq N^{(t)}$;
\end{enumerate}
As a consequence, the total number of bad pairs sampled during the second step is upper bounded by
\begin{equation*}
\cfail \sum_{t=1}^{t_f} \frac{N^{(t)}}{s} \leq 2 \cfail \frac{N_{t_f}}{s}\leq 4 C_{fail} \frac{T}{sB},
\end{equation*}
with probability larger than $1 - 63 / (sB)$. 
\end{lem}

\noindent We refer to Section \ref{section::proof:secondLem} for a proof of Lemma~\ref{th_step2_constrained}.

Let us now conclude the proof of the upper bound of Theorem~\ref{thm:contraint}.
In the third step, the core-set $\Ncal^{(t_f)}$ has $\lceil T/B \rceil \leq 2T/B$ nodes and a proportion of misclassified nodes smaller than $8 / (sB)$ with probability larger than $1 - 63/(sB)$ by point~\ref{item_misclass_constrained} of Lemma~\ref{th_step2_constrained}. Since each node of $\Ncal^{(t_f)}$ is sampled at most $B$ times, the number of bad pairs sampled during the third step is smaller than $16T/(sB)$ with probability at least $1 - 63/ (sB)$, and smaller than $T$ otherwise.

Hence, using again that we always have $N^{bad}(\psi,T)\leq T$,  the total sampling-regret $\E\cro{N^{bad}(\psi,T)}$ during the whole process is $O(T/(sB))$. The  proof of the upper bound of Theorem~\ref{thm:contraint} is complete.

\subsection{Proof of Lemma~\ref{th_step2_constrained}}
\label{section::proof:secondLem}

Lemma~\ref{th_step2_constrained}  simultaneously controls all the iterations of \texttt{SCREENING}. To prove it, we use the following lemma which controls each iteration.
\begin{lem}
\label{lem_screening}
There exists a numerical constant $c'_{thresh}\geq e$ such that the following holds.

Let $\Ncal \subset V_{init}$, $N' \in \Nbb$, $B > 0$, $\nu \in [0,1]$ and $V \subset V_{init}$, and
\begin{equation}
(\Ncal',V') = \texttt{SCREENING}(\Ncal, N', B, \nu, V).
\end{equation}Write $N = |\Ncal|$.

Assume that $s B \geq c'_{thresh}$, that $B \leq 4 N' \leq 4 N \log(sB)$, that the proportion of misclassified nodes $|\Ncal \cap \{Z \neq 1\}|/|\Ncal|$ is upper bounded by $\cmisc/(sB)$ for some constant $\cmisc \in [8/512^2, 8]$, that $\nu \in [\frac{p+3q}{4}, \frac{3p+q}{4}]$, that $|V| \geq 7n/8$ and that no node in $V$ is adjacent to a pair sampled before this call to \texttt{SCREENING}. Then with probability at least $1 - 6/(sN')$,
\begin{enumerate}
\item \label{SCREEN_missclassifiedProportion} the proportion $|\Ncal' \cap \{Z \neq 1\}|/|\Ncal'|$ of misclassified nodes after \texttt{SCREENING} is upper bounded by $\cmiscpost/(sB)$ where $\cmiscpost = \cmisc \vee 8$ if $N' \geq B \log(sB)^{3/2}$ and $\cmiscpost = 512 \cmisc$ otherwise.

\item \label{SCREEN_samplingRegret} the number of sampled bad pairs is controlled: there exists a numerical constant $\cfail$ (for instance $\cfail = 26 C_k + 2 = 65 002$) 
such that
\begin{equation}
 \sum_{x \in \Acal_0} \left| \left\{ y^x_a : \text{ $(x,y^x_a)$ was sampled and } Z_{y^x_a} \neq  Z_x \right\} \right|
	\leq \cfail \frac{N'}{s}.
\end{equation}

\item \label{SCREEN_constraint} no node in $\Ncal'$ or $V$ has more than $B$ adjacent pairs sampled during this call to \texttt{SCREENING}.

\item \label{SCREEN_consommationNewNodes} $|V'| \geq |V| - 4N'$.

\item \label{SCREEN_enoughNewVertices} it is possible to construct the core-set $\Ncal'$ with $N'$ nodes after Step~\ref{step_4_contraint}: $\sum_{j=1}^m | \Acal_I^{(j)} \cap \{ Z = 1 \} | \geq N'$.

\item \label{SCREEN_nodesInVAreNew} no node in $V'$ is adjacent to a pair sampled before or during this call to \texttt{SCREENING}.
\end{enumerate}
\end{lem}

 Lemma~\ref{lem_screening} is proved in Section~\ref{section::proof:lemma11}.\medskip 
 
To prove Lemma~\ref{th_step2_constrained}, we control the $t_f$ screening calls at the second step of the constrained algorithm page \pageref{alg:constrained} as follows. For the first step, denote by $E_0$ the event of probability $1-9/(sB)$ where all the points of Lemma~\ref{th_step1_constrain} are true. For each  $t\in \{1,\ldots,t_f\}$, denote by $E_t$ the event where all the points of Lemma~\ref{lem_screening} are satisfied by the output of \texttt{SCREENING} at the $t^{th}$-call, which is $(\Ncal^{(t)}, V^{(t)}) = \texttt{SCREENING}\pa{
	\Ncal^{(t-1)},
	N^{(t)},
	B,
	\hat{\tau},
	V^{(t-1)}
}$. On the event $\bigcap_{0\leq t \leq t_f} E_t$, all the points of Lemma~\ref{th_step2_constrained} can be easily derived, see Section~\ref{proof:lemC2:pointBYpoint} for a detailed proof.

Therefore, Lemma~\ref{th_step2_constrained} holds with a probability at least $\mathbb{P}\left( \bigcap_{0\leq t \leq t_f} E_t\right).$ To prove that $\bigcap_{0\leq t \leq t_f} E_t$ holds with high probability, we proceed by induction. First, the event $E_0$ holds with probability at least $1-9/(sB)$ by Lemma~\ref{th_step1_constrain}. 
Next, for any $t \in \{1,\ldots,t_f\}$, we check in Section~\ref{proof:lemC3:assumption}, that, on the event $E_0\cap \ldots\cap E_{t-1}$, the assumptions of  Lemma~\ref{lem_screening} holds at the $t^{th}$-call of the \texttt{SCREENING} routine.
Hence, according to Lemma~\ref{lem_screening}, conditionally on the event $E_0\cap \ldots\cap E_{t-1}$, the event $E_{t}$ holds with probability at least $1-6/(sN^{(t)})$. By induction, we thus have
\begin{align*}\mathbb{P}\left( \bigcap_{0\leq t \leq t_f} E_t\right) &= \mathbb{P}\left(E_0\right)\, \mathbb{P}\left(E_1 | E_0\right) \ldots \mathbb{P}\left(E_{t_f} | E_{t_{f}-1},\ldots,E_0\right)\\ &\geq \left(1-\frac{9}{sB}\right) \prod_{t=0}^{t_f}\left(1-\frac{6}{sN^{(t)}}\right),\end{align*}which is larger than
\begin{align*}
1 -\frac{9}{sB}- \sum_{t=1}^{t_f} \frac{6}{sN^{(t)}}
	&= 1 - \frac{9}{sB} - \frac{6}{sN^{(0)}} \sum_{t=1}^{t_f-1} \lfloor \log(sB) \rfloor^{-t} - \frac{6}{s \lceil T / B \rceil} \\
	&\geq 1 - \frac{9}{sB}- \frac{12 \log(sB)}{s B}\times \frac{\lfloor \log(sB) \rfloor^{-1}}{{1 - \lfloor \log(sB) \rfloor^{-1}}} - \frac{6}{s (T / B)} \\
	&\geq 1 - \frac{9}{sB} - \frac{48}{s B} - \frac{6}{s B} = 1 - \frac{63}{s B},
\end{align*}using for the last inequality that $B \leq \sqrt{T}/2$ and
$sB \geq c'_{thresh}$ for some numerical constant $c'_{thresh}>0$. 

To conclude, Lemma~\ref{th_step2_constrained} holds with probability at least $1 - 63/(sB)$, provided that the conclusions of Lemma~\ref{th_step2_constrained} hold on the event $\bigcap_{0\leq t \leq t_f} E_t$, and that the assumptions of Lemma~\ref{lem_screening} are satisfied at each call of \texttt{SCREENING}. These two points are proved in the next two subsections.

\subsubsection{The Conclusions of Lemma~\ref{th_step2_constrained} holds on \texorpdfstring{$\bigcap_{0\leq t \leq t_f} E_t$}{the Intersection of the Events E\_t}}\label{proof:lemC2:pointBYpoint}
Assume that the event $\bigcap_{0\leq t \leq t_f} E_t$ holds, and let us show that all the points of Lemma~\ref{th_step2_constrained} are fulfilled. 
\smallskip

\noindent \textbf{Points \ref{item_lem1.3_initial_kernel_large_enough}, \ref{item_regret_constrained} and \ref{item_enoughVertices_constrained}.}
Points~\ref{item_regret_constrained} and \ref{item_enoughVertices_constrained} of Lemma~\ref{th_step2_constrained}  follow directly from Point \ref{SCREEN_samplingRegret} and  Point \ref{SCREEN_enoughNewVertices} of Lemma \ref{lem_screening}. As for
Point~\ref{item_lem1.3_initial_kernel_large_enough}, it is satisfied when
\begin{equation*}
4 C_k C_I \frac{\log(s B)}{s} \leq \frac{B}{2 \log(s B)},
\end{equation*}
which holds
as soon as $sB \geq c'_{thresh}$ for some numerical constant $c'_{thresh}$.

\smallskip

\noindent \textbf{Point~\ref{item_misclass_constrained}.} In the initial core-set, the proportion of misclassified nodes is upper bounded by Lemma~\ref{th_step1_constrain} as follows
\begin{equation*}
|\Ncal^{(0)} \cap \{Z \neq 1\}|/|\Ncal^{(0)}| \leq 2 \varepsilon_N \leq \frac{8}{512^2} \times\frac{1}{sB}.
\end{equation*}
For the next core-set $\Ncal^{(1)}$, it implies that  
\begin{equation*}
|\Ncal^{(1)} \cap \{Z \neq 1\}|/|\Ncal^{(1)}| \leq 
        512 \frac{8}{512^2} \times\frac{1}{sB} = \frac{8}{512} \times\frac{1}{sB}
\end{equation*}
using $c_{misclas} = 8/512^2$ in the point~\ref{SCREEN_missclassifiedProportion} of Lemma~\ref{lem_screening}. For the subsequent core-sets, the proportion of misclassified nodes is  upper bounded as above, updating the value of $c_{misclas}$ at each step. We thus have
\begin{equation*}
|\Ncal^{(2)} \cap \{Z \neq 1\}|/|\Ncal^{(2)}| \leq 
       512 \frac{8}{512} \times \frac{1}{sB} = \frac{8}{sB},
\end{equation*}
and for all $t\geq 3,$
\begin{equation*}
|\Ncal^{(t)} \cap \{Z \neq 1\}|/|\Ncal^{(t)}| \leq 
        \frac{8}{sB},
\end{equation*}
since $N^{(t)} \geq B \log(sB)^{3/2}$ as soon as $t \geq 3$ and $sB \geq c'_{thresh}$ for some numerical constant $c'_{thresh}$. 

\smallskip

\noindent \textbf{Point~\ref{item_constrained_always_vertices}.} At the $t^{th}$-call to \texttt{SCREENING}, the output of ``new" nodes $V^{(t)}$ satisfies the recursive inequality $|V^{(t)}|\geq |V^{(t-1)}| - 4 N^{(t)}$ by construction of the algorithm. The sequence of inequalities telescopes, leaving
\begin{align*}
|V^{(t)}|
	\geq |V^{(0)}| - \sum_{s=1}^{t} 4N^{(s)},
\end{align*}
which is larger than $7n/8$ since $|V^{(0)}| = n - N_{init}$ and $N_{init} \leq n/8 -\sum_{s=1}^{t_f-1} 4N^{(s)}$ by the point~\ref{item_constrained_C1_Ninit} of Lemma~\ref{th_step1_constrain}.

\smallskip

\noindent \textbf{Point~\ref{item_contrainte_allIterations}.} A node can fall into four categories: \\* 
1/ it is never used; \\*
2/ it is used in Step 1 and possibly in the first iteration of \texttt{SCREENING}. Then the number of adjacent sampled pairs is at most $N_{init} + B$ by construction of Step 1 and by point \ref{SCREEN_constraint} of Lemma~\ref{lem_screening}, which is smaller than $2B$ as soon as $sB \geq c_{thresh}$ for some numerical constant $c_{thresh}$; \\*
3/ it is used in (at most) two consecutive iterations of \texttt{SCREENING} (and nowhere else). Then the number of adjacent sampled pairs is at most $2B$ by Lemma~\ref{lem_screening}; \\*
4/ it is used in the last iteration of \texttt{SCREENING} and (possibly) in Step 3. Then the number of adjacent sampled pairs is at most $B + B$ by Lemma~\ref{lem_screening} and by construction of Step 3.

\subsubsection{Check of the Assumptions of Lemma~\ref{lem_screening}}\label{proof:lemC3:assumption}

Assume that the events $E_0, \ldots,E_{t-1}$ hold together, and let us check Lemma~\ref{lem_screening} assumptions. First, the condition $sB \geq c'_{thresh}$ comes from Lemma~\ref{th_step2_constrained}. Then, following the proof of Point~\ref{item_misclass_constrained}, we can check that $|\Ncal^{(t-1)} \cap \{Z \neq 1\}|/|\Ncal^{(t-1)}| \leq c_{misclas}/(sB)$ for $c_{misclas}\in [8/512^2, 8]$. For the threshold $\hat \tau$ taking value in $[\frac{p+3q}{4}, \frac{3p+q}{4}]$, it is stated in Lemma~\ref{th_step1_constrain}. The input of ``new'' nodes $V^{(t-1)}$ satisfies $|V^{(t-1)}| \geq 7n/8$, as seen above in the proof of Point~\ref{item_constrained_always_vertices}. Finally, the inequality $B \leq 4 N^{(t)} \leq 4 N^{(t-1)} \log(sB)$ is satisfied by construction of the algorithm, as soon as $sB\geq c'_{thresh}$ for some numerical constant $c'_{thresh}$.

\subsection{Proof of Lemma~\ref{lem_screening}: Control of \texttt{SCREENING}}
\label{section::proof:lemma11}

In this section, we work  conditionally to $\Fcal_{T_\text{start}}$ where $T_\text{start}$ is the number of pairs sampled before the current call to \texttt{SCREENING}.

Let us state the two main technical results that allow to prove Lemma~\ref{lem_screening}. Write $\Vcal(x) := \Vcal_j$ for each $j \in \{1, \dots, m\}$ and $x \in \Acal_0^{(j)}$. The first one controls the properties of the sets $(\Vcal(x))_{x\in \Acal_0}$. Given a subset of nodes $S$, denote by $\misclas(S)$ the set of misclassified nodes in $S$, that is the set of all $x \in S$ such that $Z_x \neq 1$ in \texttt{SCREENING}.

\begin{lem}
\label{lem_Vcal}The sets $(\Vcal(x))_{x\in \Acal_0}$ satisfy
\begin{enumerate}
\item \label{point_1_Vcal}
For all $y \in \Ncal$, $|\ac{ x \in \Acal_0: y \in \Vcal(x) }| \leq B$,
\item \label{point_2_Vcal}
$\displaystyle \Pbb\pa{
		\left| \ac{x \in \Acal_0 : | \misclas(\Vcal(x)) | \geq \frac{kI}{16} } \right| \geq \frac{\cmiscpost}{2} \frac{N'}{sB}
	} \leq \frac{2}{sN'}
$

where $\cmiscpost$ is defined as in Lemma~\ref{lem_screening},

\item \label{point_3_Vcal}
$\displaystyle
		\sum_{x \in \Acal_0} | \misclas(\Vcal(x)) | \leq \frac{N'}{s}.
$
\end{enumerate}
\end{lem}


The proof of the above lemma is postponed to Section~\ref{sec_proof_lem_Vcal}. The next lemma allows to control the effectiveness of Step~\ref{step_4_contraint} of \texttt{SCREENING}. Its proof follows the same lines as the proof of  Lemma~\ref{lem:bandit} (for proving \eqref{eq_Tx2}) and Point~\ref{9item_il_reste_des_bons}  of Lemma~\ref{th_step2} (for proving \eqref{eq_Tx1}), it is therefore omitted.

\begin{lem}\label{lem_Tx}
 Conditionally to the choice of the set $\Acal_0$ and $(\Vcal(x))_{x\in \Acal_0}$, the variables $(T_x)_{x\in \Acal_0}$ are independent and for all $x\in \Acal_0$ and all $i \in\{1,\ldots,I\},$
\begin{equation}\label{eq_Tx2}
 \Pbb\pa{
		T_x \geq i \, \Big{|} \, Z_x \neq 1 \text{ and } |\misclas(\Vcal(x))| \leq \seuil
	} \leq e^{-i}
\end{equation}
\begin{equation}\label{eq_Tx1}
 \Pbb\pa{
		T_x \geq I \, \Big{|} \, Z_x = 1 \text{ and } |\misclas(\Vcal(x))| \leq \seuil
	} \geq \frac{3}{4}.
\end{equation}
\end{lem}

Let us now prove Lemma~\ref{lem_screening}.
Note that Points~\ref{SCREEN_consommationNewNodes} and~\ref{SCREEN_nodesInVAreNew} follow from the construction of the algorithm and that Point~\ref{SCREEN_constraint} follows straighforwardly from point~\ref{point_1_Vcal} of Lemma~\ref{lem_Vcal} (for the nodes from $\Ncal$) and from the construction of the algorithm (for the nodes from $\Acal_0$).

\subsubsection{Proof of Point~\ref{SCREEN_missclassifiedProportion}}

\begin{align*}
|\Ncal' \cap \{Z \neq 1\}|
	&\leq \sum_{j=1}^m |\misclas(\Acal^{(j)}_I)| \\
	&= \sum_{j=1}^m \!\!\!\!\! \underset{|\misclas(\Vcal(x))| > \seuil}{\sum_{x \in \Acal_0 \text{ s.t.}}} \!\!\!\!\!\! \one_{x \in \Acal^{(j)}_I \text{ and } Z_x \neq 1}
	+  \sum_{j=1}^m \underset{|\misclas(\Vcal(x))| \leq \seuil}{\sum_{x \in \Acal_0 \text{ s.t.}}} \!\!\!\!\!\! \one_{x \in \Acal^{(j)}_I \text{ and } Z_x \neq 1} \\
	&\leq \sum_{x \in \Acal_0} \one_{|\misclas(\Vcal(x))| \geq \seuil}
	+  \!\!\!\!\! \underset{|\misclas(\Vcal(x))| \leq \seuil}{\sum_{x \in \Acal_0 \text{ s.t.}}} \!\!\!\!\!\! \one_{T_x \geq I \text{ and } Z_x \neq 1} \\
	&\leq \frac{\cmiscpost}{2} \frac{N'}{sB}
	+  \!\!\!\!\! \underset{|\misclas(\Vcal(x))| \leq \seuil}{\sum_{x \in \Acal_0 \text{ s.t.}}} \!\!\!\!\!\! \one_{T_x \geq I \text{ and } Z_x \neq 1}
\end{align*}
with probability at least $1 - 2/(sN')$ by point~\ref{point_2_Vcal} of Lemma~\ref{lem_Vcal}. 

The second term is dominated by a binomial random variable with parameters $(|\Acal_0|, e^{-I})$ by Lemma \ref{lem_Tx}, so it is dominated by a binomial random variable $X$ with parameters $(4N', 1/(sB)^{1026})$ since $I \geq 1026 \log(sB)$.
Equation~\eqref{eq_large_deviations} implies that for $sB \geq 512$ (which is implied by $c'_{thresh} \geq 512$),
\begin{align*}
\label{eq_deviation_nbMauvaisVcal}
\Pbb\pa{\frac{1}{4N'} X \geq \frac{1}{512 sB}} 
	&\leq \exp\pa{- \frac{1}{2} 4N' \frac{1}{512 sB} \log\pa{\frac{(sB)^{1025}}{512}}}
	\leq \exp\pa{- 4 N' \frac{\log(sB)}{sB} }.
\end{align*} 
We want this probability to be smaller than $1/(sN')$, that is
\begin{align*}
\frac{\log(sB)}{s B} \geq s \frac{\log(sN')}{4 sN'},
\end{align*}
which is true since $s \leq 1$, and the function $x \longmapsto \frac{\log x}{x}$ is nonincreasing for $x \geq e$, and $4 sN' \geq sB \geq e$ by assumption. 

Therefore,
\begin{equation*}
\Pbb\pa{ |\Ncal' \cap \{Z \neq 1\}| \geq \frac{\cmiscpost + (8/512)}{2} \frac{N'}{sB} } \leq \frac{3}{sN'}
\end{equation*}
which implies Point~\ref{SCREEN_missclassifiedProportion} (since $\cmiscpost \geq 8/512$ by definition).

\subsubsection{Proof of Point~\ref{SCREEN_samplingRegret}}

Given a subset $S$ of $\Acal_0$, denote by $\regret(S)$ the number of sampled bad pairs coming from nodes in $S$ during \texttt{SCREENING}, that is
\begin{equation*}
\regret(S) = \sum_{x \in S}
	| \{ y^x_i : i \leq k ((T_x+1) \wedge I) \textup{ and } Z_{y^x_i} \neq Z_x \} |.
\end{equation*}

The total number of bad pairs sampled during \texttt{SCREENING} can be decomposed into
\begin{align}
\regret\pa{\Acal_0}
	=& \,  \sum_{x \in \Acal_0} \regret(\{x\}) \one_{|\misclas(\Vcal(x))| > \seuil} \nonumber \\
		&+  \sum_{x \in \Acal_0 \cap \{Z = 1\}} \regret(\{x\}) \one_{|\misclas(\Vcal(x))| \leq \seuil} \nonumber \\
		&+  \sum_{x \in \Acal_0 \cap \{Z \neq 1\}} \regret(\{x\}) \one_{|\misclas(\Vcal(x))| \leq \seuil} \nonumber \\
		\label{eq_regret_bcp_mal_classes}
	\leq& \,  kI \sum_{x \in \Acal_0} \one_{|\misclas(\Vcal(x))| \geq \seuil} \\
		\label{eq_regret_regret_bien_classes}
		&+  \sum_{x \in \Acal_0 \cap \{Z = 1\}} |\misclas(\Vcal(x))| \\
		\label{eq_regret_regret_mal_classes}
		&+ \sum_{x \in \Acal_0 \cap \{Z \neq 1\}} k(T_x+1)  \one_{|\misclas(\Vcal(x))| \leq \seuil}
\end{align}

The first sum is controlled by Point~\ref{point_2_Vcal} of Lemma~\ref{lem_Vcal}:
\begin{align*}
\Pbb\pa{
		\eqref{eq_regret_bcp_mal_classes} \geq \frac{\cmiscpost}{2} \frac{N' kI}{sB}
	} \leq \frac{2}{sN'}.
\end{align*}

Thus, since $kI/B \leq 4 C_k C_I \log(sB)/(sB)$ by definition, there exists a constant $c'_{thresh}$ such that $kI/B \leq 2/\cmiscpost$ as soon as $sB \geq c'_{thresh},$ so that 
\begin{align*}
\Pbb\pa{
		\eqref{eq_regret_bcp_mal_classes} \geq \frac{N'}{s}
	} \leq \frac{2}{sN'}.
\end{align*}

Likewise, by Point~\ref{point_3_Vcal} of Lemma~\ref{lem_Vcal},
\begin{align*}
	\eqref{eq_regret_regret_bien_classes}
		\leq \frac{N'}{s}.
\end{align*}

By Lemma \ref{lem_Tx}, the variables $T_x$ in the third sum are stochastically dominated by i.i.d. exponential random variables with parameter 1. Therefore, using the inequality $k \leq 2C_k / s$, the term \eqref{eq_regret_regret_mal_classes} is stochastically dominated by
\begin{align*}
8 C_k \frac{N'}{s} + 2 \frac{C_k}{s} \sum_{i=1}^{4 N'} Y_i 
\end{align*}
where $(Y_i)_{i \in \Nbb^*}$ are i.i.d. exponential random variables with parameter 1.
These exponential random variables satisfy
\begin{equation*}
\E \left(Y_i - 1 \right)^2 \leq 1
\end{equation*}
and for all $a \in \Nbb$ such that $a \geq 3$
\begin{equation*}
\E \left(Y_i - 1 \right)_+^a \leq a!,
\end{equation*}
so that Bernstein's inequality, see for instance Proposition 2.9 of \citep{Mas07} entails for all $t > 0$
\begin{equation*}
\P\left( \sum_{i = 1}^{4 N'} Y_i - 4 N'
	\geq 4 \sqrt{N' t}
		+ t \right) \leq e^{-t}
\end{equation*}
and therefore by taking $t = N'$, with probability at least $1 - e^{-N'} \geq 1 - 1/N' \geq 1 - 1/(sN')$
\begin{align*}
\sum_{i = 1}^{4 N'} Y_i
	\leq 9 N'.
\end{align*}

Hence, with probability at least $1 - 3/(sN')$,
\begin{align*}
\regret\pa{\Acal_0}
	\leq (26 C_k + 2) \frac{N'}{s}.
\end{align*}

\subsubsection{Proof of Point~\ref{SCREEN_enoughNewVertices}.}

Write $\Acal_I = \bigcup_{j=1}^m \Acal^{(j)}_I$ and for each $x \in \Acal_0$, let $V_x = \one_{x \in \Acal_I}$ indicate whether $x$ has been kept until the end of Step~\ref{step_4_contraint} of \texttt{SCREENING}.
Lemma \ref{lem_Tx} ensures that the random variables $(V_x)_{x \in \Acal_0 \cap \{Z= 1\} \text{ s.t. } |\misclas(\Vcal(x))| \leq \seuil}$ dominate i.i.d. Bernoulli random variables with parameter $3/4$. Therefore, Hoeffding's inequality~\eqref{eq_Hoeffding_hypergeom} entails
\begin{multline*}
\P\left( \big{|}\Acal_I \cap \{Z=1\}\big{|} \leq \frac{3 \big{|}\Acal_0 \cap \{Z=1\} \cap \{x : |\misclas(\Vcal(x))| \leq \seuil \}\big{|}}{4} - \sqrt{|\Acal_0| \frac{\log(sN')}{2}} \right) \\
	\leq \frac{1}{sN'}.
\end{multline*}

Note that $|\{x \in \Acal_0 \text{ s.t. } |\misclas(\Vcal(x))| > \seuil \}| \leq \frac{\cmiscpost}{2} N'/(sB)$ with probability at least $1 - 2/(sN')$ by Lemma \ref{lem_Vcal}. Since $|\Acal_0| = 4N'$, the previous equation entails
\begin{equation*}
\P\left( |\Acal_I \cap \{Z=1\}| \leq \frac{3 |\Acal_0 \cap \{Z=1\}|}{4} - \frac{ 3\,\cmiscpost N'}{8 sB} - \sqrt{4N' \frac{\log(sN')}{2}} \right)
	\leq \frac{3}{sN'}.
\end{equation*}

Let us assume for now that $|\Acal_0 \cap \{Z=1\}| \geq \frac{11}{7} N'$ with probability at least $1 - 1/(sN')$. Then this ensures that for $N'$ and $sB$ larger than some numerical constants (which is guaranteed by $B\geq B_0$ and $sB \geq c'_{thresh}$), 
\begin{equation*}
\P\big{(} |\Acal_I \cap \{Z=1\}| \leq N' \big{)} \leq \frac{4}{sN'},
\end{equation*}
which gives point~\ref{SCREEN_enoughNewVertices}, provided that $|\Acal_0 \cap \{Z=1\}| \geq \frac{11}{7} N'$.

The random variable $|\Acal_0 \cap \{Z=1\}|$ is an hypergeometric random variable with number of draws $4N'$ and initial probability of a winning draw $r' \in [\frac{3}{7}, \frac{4}{7}]$ because the number of nodes that have not been sampled at the start of \texttt{SCREENING} is bigger than $7n/8$ by assumption and because the true communities are balanced.

Therefore, Hoeffding's inequality~\eqref{eq_Hoeffding_hypergeom} implies
\begin{equation*}
\P\left( |\Acal_0 \cap \{Z=1\}| \leq \frac{3}{7} 4N' - \sqrt{\frac{4N' \log(sN')}{2}} \right) \leq \frac{1}{sN'},
\end{equation*}
so that for $N'$ large enough (which is implied by $B \geq B_0$ for some numerical constant $B_0$).
\begin{equation*}
\P\left( |\Acal_0 \cap \{Z=1\}| \leq \frac{11}{7} N' \right) \leq \frac{1}{sN'}.
\end{equation*}

\subsection{Proof of Lemma~\ref{lem_Vcal}}
\label{sec_proof_lem_Vcal}

\subsubsection{Points 1 and 3}
To check Point~\ref{point_1_Vcal}, it suffices to check that $\lceil 4N'/m \rceil \leq B$. For $sB \geq c'_{thresh}$ with a numerical constant $c'_{thresh}$ large enough, one has $m =  \lfloor N / (kI) \rfloor \geq N / (2kI)$ and
\begin{align}
\label{eq_ineg_N_1_sur_m}
\left\lceil \frac{4N'}{m} \right\rceil
	\leq \frac{16 N' kI}{N}
	\leq 64 C_k C_I \frac{(\log(sB))^2}{s}
	\leq B.
\end{align}

For Point~\ref{point_3_Vcal}, note that 
\begin{align*}
\sum_{x \in \Acal_0} |\misclas(\Vcal(x))|
	&\leq \left\lceil \frac{4N'}{m} \right\rceil  \sum_{j=1}^m |\misclas(\Vcal_j)| \\
	&\leq 16kI \frac{N'}{N} |\misclas(\Ncal)| \\
	&\leq 64 C_k C_I \frac{\log(sB)}{s} \cmisc \frac{N'}{sB} \\
	&\leq 64 C_k C_I \frac{\log(sB)}{sB} \cmiscpost \frac{N'}{s} \\
	&\leq \frac{N'}{s}
\end{align*}
by assumption on the the number of misclassified nodes in $\Ncal$, and as soon as $sB \geq c'_{thresh}$ for some numerical constant $c'_{thresh}$.

\subsubsection{Point~\ref{point_2_Vcal}, Small Core-sets}

In this section, we assume $N' < B \log(sB)^{3/2}$.
By Equation~\eqref{eq_ineg_N_1_sur_m}.
\begin{align*}
\left| \ac{ x \in \Acal_0 : |\misclas(\Vcal(x))| \geq \seuil } \right|
	&\leq \left\lceil \frac{4N'}{m} \right\rceil \sum_{j=1}^m \one_{|\misclas(\Vcal_j)| \geq \seuil} \\
	&\leq 16 \frac{N'}{N} kI  \sum_{j=1}^m \one_{|\misclas(\Vcal_j)| \geq \seuil}.
\end{align*}

Note that
\begin{align*}
\frac{kI}{16} \sum_{j=1}^m  \one_{|\misclas(\Vcal_j)| \geq \seuil}
	&\leq \sum_{j=1}^m  |\misclas(\Vcal_j)| \\
	&= |\misclas(\Ncal)| \leq \frac{\cmisc N}{sB}
\end{align*}
by assumption, so that
\begin{align*}
\left| \ac{ x \in \Acal_0 : |\misclas(\Vcal(x))| \geq \seuil } \right|
	&\leq 16 \frac{N'}{N} kI \times \frac{16}{kI} \frac{\cmisc N}{sB} \\
	&= 256 \cmisc \frac{N'}{sB} \\
	&= \frac{\cmiscpost}{2} \frac{N'}{sB}.
\end{align*}

This bound is not random, it holds with probability 1.

\subsubsection{Point~\ref{point_2_Vcal}, Large Core-sets}

In this section, we assume $N' \geq B \log(sB)^{3/2}$.


The number of misclassified nodes in each $\Vcal_j$ can be controlled more easily by introducing a coupling with i.i.d. Bernoulli random variables. Note that this coupling is a theoretical tool and does not appear in the algorithm.

\begin{lem}
\label{lem_couplage}
Let $K$ be a random variable taking values in $\{0, \dots, N\}$. Let $(X_x)_{x \in \Ncal}$ be a vector of random variables taking values in $\{0,1\}$ such that
\begin{itemize}
\item $\sum_{x \in \Ncal} X_x = K$
\item the distribution of $(X_x)_{x \in \Ncal}$ is invariant under permutation of $\Ncal$
\end{itemize}

Note that these two points together with the distribution of $K$ characterize the distribution of $(X_x)_{x \in \Ncal}$. Then for all $u > 0$, there exists a coupling with i.i.d. Bernoulli random variables $(Y_x)_{x \in \Ncal}$ with parameter $u$ such that by writing $M = \sum_{x \in \Ncal} Y_x$, $M$ is independent of $(X_x)_{x \in \Ncal}$ and
\begin{equation}
\label{eq_ineg_couplage}
M \geq K \; \Longrightarrow \; \pa{ \forall x \in \Ncal, \; X_x \le Y_x }.
\end{equation}
\end{lem}

\paragraph{Proof of Lemma \ref{lem_couplage}.}
Let $M$ be a binomial random variable with parameters $(N, u)$ such that $M$ and $K$ are independent. Let $(\widetilde{X}_i)_{1 \leq i \leq N}$ and $(\widetilde{Y}_i)_{1 \leq i \leq N}$ be random variables such that conditionally to $M$ and $K$ and for all $1 \leq i \leq N$,
\begin{equation*}
\widetilde{X}_i = \begin{cases} 1 & \text{ if } i \leq K \\ 0 & \text{ otherwise} \end{cases}
\end{equation*}
\begin{equation*}
\widetilde{Y}_i = \begin{cases} 1 & \text{ if } i \leq M \\ 0 & \text{ otherwise} \end{cases}.
\end{equation*}

Let $\sigma$ be a uniform random variable in the set of bijections from $\{1, \dots, N\}$ to $\Ncal$ that is independent of $K$, $M$, $(\widetilde{X}_i)_i$ and $(\widetilde{Y}_i)_i$, and define $X'_x = \widetilde{X}_{\sigma^{-1}(x)}$ and $Y_x = \widetilde{Y}_{\sigma^{-1}(x)}$ for all $x \in \Ncal$.

Then the random vector $(X'_x)_{x \in \Ncal}$ has the same distribution as the random vector $(X_x)_{x \in \Ncal}$, the random variables $(Y_x)_{x \in \Ncal}$ are i.i.d. Bernoulli random variables with parameter $u$, and Equation~\eqref{eq_ineg_couplage} holds for these two vectors.
\hfill $\square$
\bigskip

Let $M$ and $(Y_x)_{x \in \Ncal}$ be the random variables given by Lemma~\ref{lem_couplage} applied to $(X_x)_{x \in \Ncal} = (\one_{\widehat{Z}_x \neq Z_x})_{x \in \Ncal}$, $K = |\misclas(\Ncal)|$ and $u = 2\cmiscpost / (sB) + 4\log(sB)^2 / B$. Note that the algorithm is invariant by permutation of the nodes of $\Ncal$, so that we may assume without loss of generality that the distribution of these $(X_x)_{x \in \Ncal}$ is invariant by permutation of $\Ncal$.

By Assumption of Lemma~\ref{lem_screening}, we have $K\leq \cmisc / (sB)$. 
Let us show that $M \geq \cmisc / (sB)$ with probability at least $1 - 1/(sN')$, which implies $M \geq K$ with probability at least $1 - 1/(sN')$.
Since $M$ is a binomial random variable with parameters $(N,u)$, Bernstein's inequality~\eqref{eq_Bernstein} entails
\begin{equation*}
\Pbb\pa{
		M \leq N u - \sqrt{2 N u t} - t
	} \leq e^{-t}.
\end{equation*}

Since $\sqrt{2 a b} \leq \frac{a}{2} + b$ for all $a,b > 0$, it holds with probability at least $1 - 1/(sN')$
\begin{align*}
M &\geq N u - \frac{N u}{2} - \log(sN') - \log(sN') \\
	&\geq \frac{\cmisc N}{sB} + 2 N \frac{\log(sB)^2}{B} - 2 \log(sN').
\end{align*}

Note that
\begin{align*}
\frac{2 \log(sN')}{2 N \frac{\log(sB)^2}{B}}
	\leq \frac{ \frac{\log(sN \log(sB))}{N} }{ \frac{\log(sB)^2}{B} } = \frac{ \frac{\log(sN \log(sB))}{sN \log(sB)} }{ \frac{\log(sB)}{sB} }
	\leq 1,
\end{align*}
as soon as $sB \geq e$ since the application $x \longmapsto (\log x)/x$ is nonincreasing for $x \geq e$ and $sN \log(sB) \geq sB \geq e$ (the second last inequality comes from the assumption $N'= N \lfloor \log(sB) \rfloor \geq B \log(sB)^{3/2}$ made at the beginning of  the current subsection). Therefore,
\begin{equation*}
\Pbb\pa{
		M \leq \frac{\cmisc N}{sB}
	} \leq \frac{1}{sN'}\,,
\end{equation*}
and finally, according to Lemma~\ref{lem_couplage},
\begin{equation}
\label{eq_couplage_vf}
\Pbb\pa{
	\forall x \in \Ncal, \quad \one_{\widehat{Z}_x \neq Z_x} \leq Y_x
} \geq 1 - \frac{1}{sN'}.
\end{equation}

{We can now proceed to the conclusion of the proof of Point~\ref{point_2_Vcal} when $N' \geq B \log(sB)^{3/2}$.}
We have
\begin{align*}
\left| \ac{ x \in \Acal_0 : |\misclas(\Vcal(x))| \geq \seuil } \right|
	&\leq \left\lceil \frac{4N'}{m} \right\rceil  \sum_{j=1}^m \one_{|\misclas(\Vcal_j)| \geq \seuil} \\
	&\leq 16 \frac{N'}{N} kI  \sum_{j=1}^m \one_{ \sum_{x \in \Vcal_j} Y_x \geq \seuil}
\end{align*}
with probability at least $1 - 1/(sN')$ by Equations~\eqref{eq_ineg_N_1_sur_m} and~\eqref{eq_couplage_vf}.

Note that the random variable $\sum_{j=1}^m \one_{ \sum_{x \in \Vcal_j} Y_x \geq \seuil}$ is a binomial random variable with parameters $(m, \Pbb(\sum_{x \in \Vcal_j} Y_x \geq \seuil))$, and that $\sum_{x \in \Vcal_j} Y_x$ is a binomial random variable with parameters $(kI, u)$ with $u = 2\cmiscpost / (sB) + 4 \log(sB)^2 / B$. Since $5 u \leq 1/16$ for $sB \geq c'_{thresh}$, we can apply Equation~\eqref{eq_large_deviations} to obtain
\begin{align}
\label{eq_temp_proba_malclasses_dans_un_Vcal}
\Pbb\pa{
		\sum_{x \in \Vcal_j} Y_x \geq \seuil
	} \leq \exp\pa{ - \frac{kI}{32} \log \frac{1}{16 u} }.
\end{align}

Note that
\begin{align*}
\log \frac{1}{16 u} &\geq \log \frac{sB}{256 (1 \vee ( s \log(sB)^2 ))} \\
	&= \log (sB) - \log 256 - 0 \vee \log (s \log (sB)^2) \\
	&\geq \frac{2}{3} \log (sB) - 0 \vee \log ((sB)^{1/3}) \\
	&\geq \frac{2}{3} \log (sB) - \log (s B) / 3 = \frac{1}{3} \log(sB),
\end{align*}
when $sB \geq c'_{thresh}$ for $c'_{thresh}$ large enough. Therefore, 
Equation~\eqref{eq_temp_proba_malclasses_dans_un_Vcal} implies
\begin{align*}
\Pbb\pa{
		\sum_{x \in \Vcal_j} Y_x \geq \seuil
	} \leq \exp\pa{ - \frac{kI}{96} \log(sB) }.
\end{align*}

It remains to control the probability that a binomial random variable with parameters $\pa{ m, \exp\pa{ - \frac{kI}{96} \log(sB) } }$ exceeds $\frac{\cmiscpost}{2} \frac{N' / (sB)}{16 kI N'/N}$. To apply Equation~\eqref{eq_temp_proba_malclasses_dans_un_Vcal}, check that
\begin{align*}
\frac{
\frac{\cmiscpost}{2} \frac{N' / (sB)}{16 kI N'/N}
}{
	m \exp\pa{ - \frac{kI}{96} \log(sB) }
}
	&= \frac{\cmiscpost}{2} \frac{ \frac{N'}{m} }{ 16 kI \frac{N'}{N} } \frac{1}{sB}
			\exp\pa{\frac{kI}{96} \log(sB) } \\
	&\geq \frac{\cmiscpost}{2} \frac{1}{16 sB}
			\exp\pa{\frac{kI}{96} \log(sB) } \quad \text{ since } m \leq \frac{N}{kI} \\
	&\geq \exp\pa{\frac{kI}{200} \log(sB) } \geq 5,
\end{align*}
for $sB \geq c'_{thresh}$. Thus, Equation~\eqref{eq_large_deviations} and $\cmiscpost = \cmisc \vee 8$ imply
\begin{align*}
\Pbb\pa{
	16 \frac{N'}{N} kI \sum_{j=1}^m \one_{ \sum_{x \in \Vcal_j} Y_x \geq \seuil} \geq \frac{\cmisc \vee 8}{2} \frac{N'}{sB}
	}
		&\leq \exp\pa{ - \frac{\cmisc \vee 8}{4} \frac{N' / (sB)}{16 kI N'/N} \frac{kI}{200} \log(sB) } \\
		&\leq \exp\pa{ - \frac{N \log(sB)}{1600 sB} } \\
		&\leq \exp\pa{ - \frac{N'}{1600  sB} }
	\quad \text{since } N \log(sB) \geq N'.
\end{align*}

We want this probability to be smaller than $1 / (sN')$, that is
\begin{align*}
\frac{N'}{1600  sB} \geq \log(sN')
\end{align*}
which holds as soon as $N' \geq \cro{2 \times 1600 sB \log(1600 s^2 B)},$ which is implied by the assumption $N' \geq B \log(sB)^{3/2}$ for $sB\geq c'_{\thresh}$. Thus,
\begin{align*}
\Pbb\pa{
	\left| \ac{ x \in \Acal_0 : |\misclas(\Vcal(x))| \geq \seuil } \right| \geq \frac{\cmiscpost}{2} \frac{N'}{sB}
	}
		&\leq \frac{2}{sN'}.
\end{align*}
The proof is complete.


\section{Probabilistic Inequalities}
We recall Bernstein and Hoeffding inequalities for binomial and hypergeometric distributions.

\begin{lem}
\label{lem_concentration_usuelle}
For $n \geq 1$, $p \in [0,1]$ and $N \geq n$, let $X$ be either a binomial random variable with parameters $(n,p)$ or a sum of $m$ i.i.d. hypergeometric random variables with parameters $(\frac{n}{m},p,N)$. 
Then, for all $t>0$,
\begin{equation}
\label{eq_Hoeffding_hypergeom}
\P\left( X - np \geq \sqrt{\frac{nt}{2}} \right) \leq e^{-t}
	\qquad \text{and} \qquad
\P\left( |X - np| \geq \sqrt{\frac{nt}{2}} \right) \leq 2 e^{-t}
\end{equation}
and
\begin{equation}
\label{eq_Bernstein}
\P \left(X-np \geq \sqrt{2 np {t}} + {t} \right) \leq e^{-t},
\end{equation}
\begin{equation}
\label{eq_Bernstein_sym}
\P \left( \left| X-np \right| \geq \sqrt{2 np{t}} + {t} \right) \leq 2e^{-t}.
\end{equation}
\end{lem}

The following lemma allows to control large deviations of binomial and hypergeometric random variables.
\begin{lem}
\label{lem_large_deviations}
Let $X$ be either a binomial random variable with parameters $(n,p)$ or a sum of $m$ i.i.d. hypergeometric random variables with parameters $(\frac{n}{m},p,N)$. Then for all $c \in [p, 1]$,
\begin{equation}
\label{eq_large_deviations_kl}
\Pbb\pa{ X \geq n c } \leq e^{- n \cdot kl(c, p)}
\end{equation}
where $kl(c,p) = c\log(c/p) + (1-c)\log((1-c)/(1-p))$.

In particular, if $c \geq 5 p$,
\begin{equation}
\label{eq_large_deviations}
\Pbb\pa{ X \geq n c } \leq e^{- \frac{1}{2} n c \log \frac{c}{p}}.
\end{equation}
\end{lem}

\noindent{\bf Proof of Lemma \ref{lem_large_deviations}.}
The large deviation Inequality~\eqref{eq_large_deviations_kl} is derived by the classical Cramèr-Chernoff's method, see for instance Chapter 2 in~\citep{Mas07}.

For Inequality~\eqref{eq_large_deviations}, note that for all $0<\alpha < 1/p$,
\begin{align*}
kl(\alpha p, p) 
	&= \frac{\alpha p}{2} \log \alpha
		+ \cro{
			\frac{\alpha p}{2} \log \alpha - (1 - \alpha p) \log \frac{1-p}{1-\alpha p}
		} \\
	&= \frac{\alpha p}{2} \log \alpha
		+ \cro{
			\frac{\alpha p}{2} \log \alpha - (1 - \alpha p) \log \pa{ 1 + p \frac{\alpha - 1}{1-\alpha p} }
		} \\
	&\geq \frac{\alpha p}{2} \log \alpha
		+ \cro{
			\frac{\alpha p}{2} \log \alpha - p (\alpha - 1)
		} \\
	&\geq \frac{\alpha p}{2} \log \alpha
		+ p \cro{
			\frac{\alpha \log \alpha}{2} + 1 - \alpha
		},
\end{align*}
and the term inside the square brackets is positive as soon as $\alpha \geq 5$.
\hfill $\square$\medskip

We also recall some classical controls on the Kullback-Leibler divergence between two Bernoulli distribution.
\begin{lem}\label{lem:kullback}
For any $p_{1},p_{2} \in [0,1]$,
\begin{equation*}
\frac{(p_{1}-p_{2})^2}{p_{1}\vee p_{2}} \leq kl(p_{1},p_{2}) \leq \frac{(p_{1}-p_{2})^2}{p_{1}(1-p_{1})\wedge p_{2}(1-p_{2})}.
\end{equation*}
In particular, for any $q \leq p \leq 1/2$,
\begin{equation*}
s = \frac{(p-q)^2}{p+q}\leq \frac{(p-q)^2}{p} \leq kl(p,q)\vee kl(q,p) \leq \frac{2(p-q)^2}{q}=2(1+p/q)s.
\end{equation*}
\end{lem}

\paragraph{Miscellaneous inequalities.}

The following inequality is used repeatedly in the proofs.
\begin{lem}
\label{lem_inversion_xlogx}
For all $x > 0$ and $y \geq 0$,
\begin{equation}
x \geq (2 y \log y) \vee e  \; \Longrightarrow \; \frac{x}{\log x} \geq y.
\end{equation}
\end{lem}

\begin{funding}
Christophe Giraud acknowledges partial support by grant ANR-19-CHIA-0021-01 (BiSCottE, ANR), and by grant ANR-21-CE23-0035 (ASCAI,  ANR and DFG). The work of Yann Issartel was supported by an AMX  scholarship, via \'Ecole Polytechnique.
\end{funding}

\bibliographystyle{alpha}
\bibliography{matchingSBM}

\end{document}